\pdfoutput=1
\documentclass{article}

\PassOptionsToPackage{numbers,compress}{natbib}
\usepackage[eandd, final]{neurips_2026}

\usepackage[utf8]{inputenc}
\usepackage[T1]{fontenc}
\usepackage[hidelinks,hypertexnames=false]{hyperref}
\hypersetup{
  pdftitle={TSB-SEG: A Systematic Time-Series Segmentation Benchmark},
  pdfauthor={Félix Chavelli, Arik Ermshaus, Fan Yang, Patrick Schäfer, John Paparrizos, Paul Boniol},
  pdfsubject={NeurIPS 2026, Track on Evaluations and Datasets},
  pdfkeywords={time-series segmentation, change point detection, state detection, benchmark},
}
\usepackage{url}
\usepackage{booktabs}
\usepackage{amsfonts}
\usepackage{amsmath}
\usepackage{amssymb}
\usepackage{amsthm}
\usepackage{nicefrac}
\usepackage{microtype}
\usepackage{xcolor}
\usepackage{colortbl}
\usepackage{graphicx}
\usepackage{float}
\usepackage{algorithm}
\usepackage{wrapfig}
\usepackage{algpseudocode}
\usepackage{tabularx}
\usepackage{multirow}
\usepackage{longtable}
\usepackage{subcaption}
\usepackage{placeins}

\usepackage{pifont}
\usepackage{array}
\usepackage{enumitem}
\usepackage[most]{tcolorbox}
\usepackage{mdframed}

\newtcolorbox{questionbox}{%
  enhanced, frame hidden, colback=white,
  borderline north={0.6pt}{0pt}{black!75},
  borderline south={0.6pt}{0pt}{black!75},
  left=4pt, right=4pt, top=2pt, bottom=4pt,
  before skip=6pt, after skip=6pt, halign=center}

\usepackage{tikz}
\definecolor{markyes}{HTML}{1E7D4F}
\definecolor{markno}{HTML}{B23A3A}
\newcommand{\cmark}{\tikz[baseline=0pt, x=1ex, y=1ex, line width=0.6pt, line cap=round, line join=round]\draw (0,0.75) -- (0.38,0.32) -- (1.05,1.35);}
\newcommand{\xmark}{\tikz[baseline=0pt, x=1ex, y=1ex, line width=0.5pt, line cap=round]{\draw (0.1,0.3) -- (0.95,1.15); \draw (0.1,1.15) -- (0.95,0.3);}}
\newcommand{\greencmark}{{\color{markyes}\cmark}}
\newcommand{\redxmark}{{\color{markno}\xmark}}

\graphicspath{{./figures/}}

\title{{\normalfont \texttt{TSB-SEG}}: A Systematic Time-Series Segmentation Benchmark}

\newcommand{\authorentry}[3]{\parbox[t]{0.45\textwidth}{\centering\textbf{#1}\\\mdseries #2\\\texttt{#3}}}

\author{%
    \authorentry{F\'elix Chavelli}{Inria, ENS, CNRS, PSL}{felix.chavelli@inria.fr} \\
    \And
    \authorentry{Arik Ermshaus}{Humboldt-Universit\"at zu Berlin}{ermshaua@informatik.hu-berlin.de} \\
    \And
    \authorentry{Fan Yang}{The Ohio State University}{yang.7007@osu.edu} \\
    \And
    \authorentry{Patrick Sch\"afer}{Humboldt-Universit\"at zu Berlin}{patrick.schaefer@hu-berlin.de} \\
    \And
    \authorentry{John Paparrizos}{The Ohio State University and Aristotle University of Thessaloniki}{john@paparrizos.org} \\
    \And
    \authorentry{Paul Boniol}{Inria, ENS, CNRS, PSL}{paul.boniol@inria.fr} \\
}

\begin{document}

\maketitle

\begin{abstract}
Time-series segmentation, studied as either \emph{Change Point Detection} (CPD) or \emph{State Detection} (SD), underpins a broad range of monitoring and diagnostic applications. However, the field suffers from a persistent divide between the statistical community (focusing on CPD) and the machine learning and data mining communities (focusing on SD). Consequently, the literature remains fragmented: evaluations typically cover narrow methodological families, datasets are often small or homogeneous, and CPD and SD are rarely evaluated together. We address these gaps with \texttt{TSB-SEG}, a unified benchmark comprising $592$ univariate and multivariate time series from $8$ heterogeneous datasets and surveying $26$ algorithms that span five decades of progress. To ensure a fair comparison, we introduce a two-pool design: a scalable \emph{main pool} of tractable detectors used for the aggregate ranking, and an \emph{extended-scope pool} for computationally intensive methods. All methods are unified under \texttt{tsseg}, an open-source library %
to ensure full reproducibility. Our systematic evaluation yields three insights: (a) no single detector dominates across domains, and per-dataset hyperparameter tuning remains decisive for performance, (b) CPD accuracy is a useful, though imperfect, proxy for SD accuracy, and (c) data characteristics should guide method selection. We provide actionable guidance based on data properties, error tolerances, and tuning budgets, identify impactful hyperparameters, and characterize algorithm robustness to these choices. %
\end{abstract}

\vspace{-0.1cm}
\section{Introduction} \label{sec:introduction}
\vspace{-0.2cm}
The widespread availability of sensing and data acquisition technologies has enabled the large-scale collection of real-valued observations over time, commonly referred to as \emph{time series}~\cite{Lara2013Survey,Kanawaday2017Machine,Woollam2022SeisBench,Ermshaus2023ClaSP,paparrizos2016screening,paparrizos2018fast,paparrizos2016detecting,liu2023amir,liu2021decomposed,liu2024adaedge,mckeown2016predicting,paparrizos2021vergedb,goel2016social,krishnan2019artificial,jiang2020pids,jiang2021good,dziedzic2019band,paparrizos2023accelerating,paparrizos2023querying,paparrizos2022fast,jeung2010effective,paparrizos2011advanced,hishida2025beyond,d2024beyond}. Among the many tasks they enable (classification, clustering, anomaly detection, similarity search, pattern mining and forecasting~\cite{boss2015,Sch_fer_2017,hivecote,middlehurst2024bake,paparrizos2015k,paparrizos2024bridging,paparrizos2019grail,boniol2024divetimeseriesanomalydetection,paparrizos2022tsb,liu2024elephant,senin2014grammarviz,paparrizos2022vus,10.14778/3407790.3407792,lin2003symbolic,shieh2008isax,SFA2012,paparrizos2024survey,dhondt2025structured,yuan2012visual,feremans2025motiplus,Yang2025SPARTAN,yang2025sail,Yuqietal-2023-PatchTST,ansari2024chronos,liu2024timemmd,liu-etal-2024-lstprompt,pmlr-v235-liu24ae,liu2025evaluating,zhao2023performative,zhao2025tackling,zhao2025timerecipe,boniol2021sand,boniol2021sandinaction,paparrizos2020debunking,paparrizos2017fast,liu2024time,sylligardos2023choose,sylligardos2025msad,boniol2024adecimo,boniol2025vus,paparrizos2025advances,paparrizos2025time,paparrizos2023odyssey,boniol2022theseus,boniol2023new,bariya2021k,liu2025easyad,liu2025mllm4ts,liu2025tsb,boniol2024interactive,giannoulidis2025burst,papadopoulos2026power,huang2026hydra,huang2026glassboxad,paparrizos2026comprehensive,li2026mufasa,pastrana2026read}), identifying changes in the underlying data-generating process is fundamental to a  broad range of monitoring and diagnostic applications, spanning human activity recognition~\cite{Yin2008Sensor,chamroukhiJointSegmentationMultivariate2013,Ermshaus2023Human}, industrial condition monitoring~\cite{Tran2022Machine}, and health assessment~\cite{Kemp2000Analysis}.

In these settings, the signal can be viewed as a sequence of \emph{segments}, each reflecting the characteristic behaviour of an underlying process \emph{state}~\cite{Ermshaus2024ClaSS}, separated by \emph{change points}. The literature studies this through two complementary formulations: \emph{change point detection} (CPD)~\cite{Truong2020Selective,Aminikhanghahi2017Survey,Ermshaus2023ClaSP}, which localizes boundaries, and \emph{state detection} (SD)~\cite{Wang2023Time2State,Lai2024E2Usd}, which additionally labels recurring states. The two diverge as soon as states recur: in a wearable recording of walking, running, then walking again, CPD only needs the two boundaries, whereas SD must also recognize the first and last segments as the same activity; an industrial test cycle revisits the same operating modes dozens of times, and a monitored patient alternates between a few physiological regimes. Despite decades of progress, existing benchmarks (Table~\ref{tab:benchmark_comparison}) still suffer from three major limitations%
: (L1) datasets are homogeneous or small-scale, while methodological coverage rarely extends beyond a single algorithmic family~\cite{Burg2020Evaluation}; (L2) SD lacks a comprehensive empirical comparison~\cite{Wang2023Time2State,Ermshaus2025CLaP}; (L3) CPD and SD are studied in isolation, leaving the relationship between boundary localization and state identification largely unexplored. Therefore, the following question remains unanswered:

\begin{questionbox}
\emph{What have five decades of advances in segmentation achieved,\\ when evaluated in a unified benchmark?}
\end{questionbox}

\begin{wraptable}{r}{7.5cm}
    \centering
    \small
    \setlength{\tabcolsep}{4pt}
    \caption{Comparison with existing TSS benchmarks.\label{tab:benchmark_comparison}}
    \begin{tabularx}{\linewidth}{l @{\extracolsep{\fill}} ccc cc c}
    \toprule
    \textbf{Benchmark} &
    \multicolumn{3}{c}{\textbf{Dataset}} &
    \multicolumn{2}{c}{\textbf{Task}} &
    \multicolumn{1}{c}{\textbf{Eval.}} \\
    \cmidrule(lr){2-4}\cmidrule(lr){5-6}\cmidrule(lr){7-7}
    & \# TS & Uni. & Multi. & CPD & SD & \# Alg.  \\
    \midrule
    UTSA~\cite{Gharghabi2018Domain} & 32  & \greencmark & \redxmark & \greencmark & \redxmark & 3  \\
    TCPD~\cite{Burg2020Evaluation} & 42 & \greencmark & \greencmark & \greencmark & \redxmark & 14  \\
    SKAB~\cite{Katser2020SKAB} & 34  & \redxmark & \greencmark & \greencmark & \redxmark & 10  \\
    TSSB~\cite{Ermshaus2023ClaSP} & 75  & \greencmark & \redxmark & \greencmark & \redxmark & 7  \\
    HAS~\cite{Ermshaus2023Human} & 250  & \redxmark & \greencmark & \greencmark & \redxmark & 13  \\
    \midrule
    \textbf{TSB-SEG} & \textbf{592} & \greencmark & \greencmark & \greencmark & \greencmark & \textbf{26}  \\
    \bottomrule
    \end{tabularx}
\end{wraptable}

In this paper, we address this question by introducing \textbf{\texttt{TSB-SEG}}\footnote{\url{https://github.com/fchavelli/TSB-SEG}, with the library \url{https://github.com/fchavelli/tsseg}.}, the most comprehensive benchmark for time-series segmentation to date. \texttt{TSB-SEG} curates $592$ univariate and multivariate series from $8$ datasets spanning heterogeneous domains, complemented by a long-series scalability case study. The benchmark surveys $26$ algorithms representing five decades of progress, organised by how they search the segmentation and what they compare, and integrated within the open-source library, \texttt{tsseg}~\cite{tsseg2026}, which provides a unified  \texttt{fit}/\texttt{predict} API. 
A scalable \emph{main pool} forms the basis of the aggregate ranking, while an \emph{extended-scope pool} gathers methods that cannot run on every series (Section~\ref{sec:benchmark_methods}).

\noindent \textit{Key findings.} Our evaluation reveals three transferable insights: (i)~no single detector consistently dominates across datasets, making per-dataset tuning essential; (ii)~CPD and SD scores are strongly correlated, making CPD accuracy a useful proxy for SD accuracy; (iii)~data characteristics should guide method selection.

\noindent \textit{Contributions.} (i) A taxonomy of CPD and SD methods by search procedure and homogeneity criterion (Section~\ref{sec:benchmark_methods}); (ii) \texttt{TSB-SEG}, the first unified benchmark to jointly evaluate CPD and SD across the two-pool design above (Section~\ref{sec:benchmark}); (iii) $8$ complementary CPD and SD metrics, including soft and symmetric variants of the standard CPD measures and the recent WARI and SMS~\cite{chavelli2025} for SD (Section~\ref{sec:metrics}); (iv) a large-scale evaluation with rigorous statistical testing and clear practical recommendations (Section~\ref{sec:experiments}); (v) \texttt{tsseg}, an open-source library packaging all algorithms and metrics, with the \texttt{TSB-SEG} pipelines and dataset loaders for full reproducibility.

\vspace{-0.1cm}
\section{Foundation and Related Work} \label{sec:fandRW}
\vspace{-0.2cm}

In this section, we review the most relevant related work, formalize time-series segmentation, and discuss the limitations of existing benchmarks. Key notations are summarised in Appendix~\ref{app:notation}.

\noindent\textbf{Notation:}
An \emph{abstract process}~\cite{Wang2024Unsupervised} is defined as $P=(\mathcal{S},R)$, where $\mathcal{S}=\{s_1,\dots,s_{n_S}\}$ denotes a finite set of discrete and distinct \emph{states}, and $R\subseteq \mathcal{S}\times\mathcal{S}$ denotes the set of admissible \emph{transitions} between states, with $(s_i,s_j)\in R$ only if $s_i\neq s_j$. Each state is assumed to generate a stationary and distinguishable signature in the observed data. For clarity, we exclude \emph{homonym} states, whose signals are indistinguishable, as well as states whose semantics drift over time. A \emph{time series} $T=(t_1,\dots,t_n)\in\mathbb{R}^{n\times d}$ is a sequence of $d$-dimensional observations generated by $P$ and recorded by $d$ sensors (univariate when $d=1$, multivariate otherwise). A \emph{subsequence} (or \emph{window}) $T_{s,e}$ denotes the contiguous slice $(t_s,\dots,t_e)$ of length $e-s+1$. The \emph{state sequence} $Q=(s_{t_1},\dots,s_{t_n})\in \mathcal{S}^n$ assigns a state to each timestamp. A \emph{change point} (CP) is an index $i$ such that $s_{t_{i-1}}\neq s_{t_i}$ and the contiguous interval between two CPs is a \emph{segment} of the time series, generated by a single state.

\subsection{Time Series Segmentation (TSS): \textit{A Multifaceted Problem}} \label{sec:tss}
\vspace{-0.1cm}

\begin{figure}[t]
    \centering
    \includegraphics[width=1.0\linewidth]{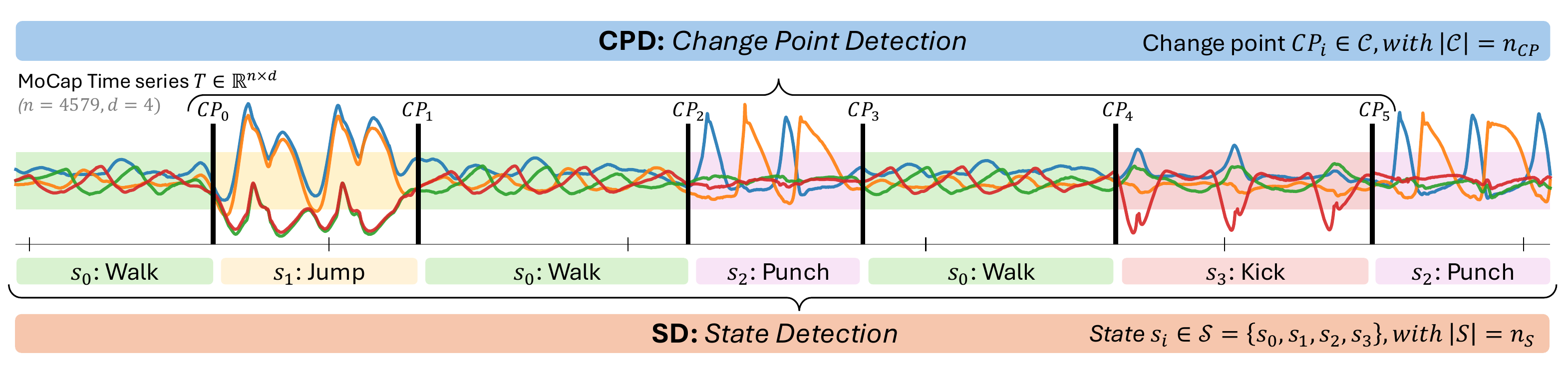}
    \caption{Time Series Segmentation: \textit{Change Point Detection} (CPD) and \textit{State Detection} (SD).}
    \label{fig:probdef}
\end{figure}

TSS partitions a time series into homogeneous segments and is studied through two complementary formulations, as illustrated in Figure~\ref{fig:probdef}.

\vspace{-0.2cm}
\paragraph{Change Point Detection (CPD).}
CPD~\cite{bassevilleDetectionAbruptChanges,Truong2020Selective,Aminikhanghahi2017Survey} aims to identify the temporal indices at which the underlying data-generating process changes, thereby partitioning $T$ into segments that are internally homogeneous while remaining distinct from adjacent ones. It is also a special case of Change Interval Detection~\cite{Cao2022Detecting}, where the change is localized to an instantaneous boundary.

\vspace{-0.2cm}
\paragraph{State Detection (SD).}
SD (a.k.a.\ \emph{semantic segmentation} or \emph{regime shift detection}) reconstructs the latent state sequence~\cite{Wang2024Unsupervised} by assigning an unsupervised label to every observation, such that temporally disjoint but structurally similar regions receive the same label. In this sense, SD extends CPD with a clustering objective: equal-state segments share a label, different-state segments do not. Since the labels are inferred without supervision, the optimal outcome is a prediction $\hat Q$ \emph{isomorphic} to the ground-truth $Q$. By construction, $\hat Q$ induces a segmentation (a change point wherever the label changes), so every SD method is also a CPD method, which lets us score SD methods on both tasks.

In most settings, the number of segments or states is assumed to be unknown a priori, reflecting a challenge analogous to determining the number of clusters in unsupervised clustering~\cite{Nguyen2015Survey}.

\subsection{Gaps in Existing Benchmarks} \label{sec:differences}

Existing TSS benchmarks listed in Table~\ref{tab:benchmark_comparison} suffer from three structural limitations:

\noindent\textbf{(L1) Limited dataset diversity.} Most existing benchmarks for TSS remain limited in either scale or scope. Many focus on a single application domain~\cite{Ermshaus2023Human}, contain a small number of time series~\cite{Gharghabi2018Domain,Katser2020SKAB,Burg2020Evaluation}, or restrict evaluation to either univariate~\cite{Gharghabi2018Domain,Ermshaus2023ClaSP} or multivariate settings~\cite{Katser2020SKAB,Ermshaus2023Human}. As a result, no current benchmark simultaneously captures diversity in domains, dimensionality, and segmentation complexity at scale, which limits the external validity of empirical findings.

\noindent\textbf{(L2) SD remains underexplored.} In contrast to CPD, which has been extensively studied~\cite{Gharghabi2018Domain,Katser2020SKAB,Burg2020Evaluation,Ermshaus2023ClaSP,Ermshaus2023Human}, SD has received comparatively little attention from an empirical evaluation perspective. To the best of our knowledge, SD methods have only been compared within individual method papers~\cite{Wang2023Time2State,Ermshaus2025CLaP}, each with its own datasets and protocol, and no dedicated SD benchmark exists.

\noindent\textbf{(L3) CPD and SD are studied in isolation.} Despite their conceptual proximity, CPD and SD have been evaluated separately, with no benchmark jointly evaluating both tasks on shared datasets or analyzing the relationship between boundary location accuracy and state identification quality. It thus remains unclear how far advances in CPD transfer to SD.

\section{Objectives and Research Questions} \label{sec:objandRQ}

The primary objective of this work is to establish a systematic and reproducible evaluation framework for TSS methods that addresses the limitations of existing benchmarks discussed in Section~\ref{sec:differences}. Guided by this objective, we investigate the following research questions:

\begin{enumerate}
    \item[\textbf{RQ1:}] \textbf{Do we have a ``one size fits all'' method?}  
    \textit{This question examines whether any existing CPD or SD method achieves consistently strong performance across varying domains.}

    \item[\textbf{RQ2:}] \textbf{Can CPD performance serve as a reliable proxy for SD accuracy?}  
    \textit{Since CPD labels are easier to obtain, establishing CPD accuracy as a proxy for SD would be impactful.}

    \item[\textbf{RQ3:}] \textbf{Which actionable guidelines  exist in real applications?}  
    \textit{
    We analyse runtime-accuracy trade-offs, scalability, and sensitivity to input properties, to provide practical guidelines.}
\end{enumerate}

\begin{figure}[t]
    \centering
    \includegraphics[width=\linewidth]{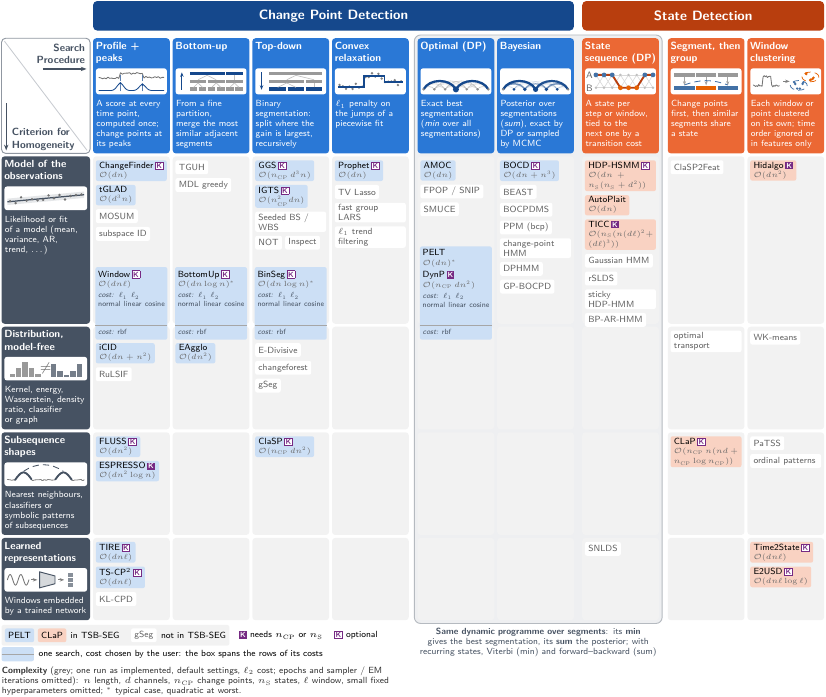}
    \caption{Time-series segmentation taxonomy: search procedure (columns; DP: dynamic programming) and homogeneity criterion (rows). Tinted: evaluated in \texttt{TSB-SEG}; white: representative methods not evaluated (Appendix~\ref{app:method_details}). K: needs (filled) or accepts (outlined) $n_{CP}$ or $n_S$ (Section~\ref{sec:exp_strategy}).}
    \label{fig:taxonomy}
\end{figure}

\vspace{-0.1cm}
\section{\texttt{TSB-SEG}: Our Proposed Benchmark} \label{sec:benchmark}
\vspace{-0.1cm}

We present \textbf{\texttt{TSB-SEG}}, a benchmark specifically designed to address the gaps outlined in Section~\ref{sec:differences}. For \textbf{(L1)}, we curate 8 heterogeneous datasets ($592$ labeled series, univariate and multivariate, from diverse domains) and PAMAP2 as a long-series scalability case study; for \textbf{(L2)}, we treat SD as a first-class task with 7 dedicated methods and state-level annotations on every dataset, enabling the first reproducible large-scale SD comparison; for \textbf{(L3)}, we also score the SD methods on CPD, through the change points their labels induce, enabling direct cross-task analysis. All algorithms and metrics are released through \texttt{tsseg}, and the evaluation pipelines through the \texttt{TSB-SEG} repository; 13 of the 26 algorithms were absent from existing toolkits (\texttt{aeon}~\cite{aeon}, \texttt{sktime}~\cite{sktime}, \texttt{ruptures}~\cite{Truong2020Selective}, and the R package \texttt{changepoint}~\cite{JSSv058i03}) and were re-implemented or adapted (univariate/multivariate extensions); see Appendix~\ref{app:mts_adapt}.

\subsection{Taxonomy: Search Procedure and Homogeneity Criterion} \label{sec:benchmark_methods}

We organise the 26 algorithms with two questions (Figure~\ref{fig:taxonomy}), which extend to SD the decomposition of a CPD method into a search method and a cost function~\cite{Truong2020Selective}: \emph{how is the segmentation searched?} (columns) and \emph{what must stay the same within a segment?} (rows). The search sets the runtime of a method and how it chooses the number of changes (a threshold, a penalty, a prior, or a given $K$); the criterion sets which changes it can detect at all (details in Appendix~\ref{app:method_details}).

\noindent\textbf{How CPD methods search.} The simplest search is local: a \emph{profile} method scores every time point, usually by contrasting the windows on its two sides, and reports the peaks of this score (Window~\cite{Truong2020Selective}, ChangeFinder~\cite{changefinder2006}, iCID~\cite{Cao2022Detecting}, FLUSS~\cite{Gharghabi2018Domain}, ESPRESSO~\cite{Deldari2020ESPRESSO}, TIRE~\cite{DeRyck2020Change}, TS-CP$^2$~\cite{Deldari2020Time}, tGLAD~\cite{Imani2023tGLAD}). The other searches compare whole segments. \emph{Bottom-up} search starts from a fine partition and repeatedly merges the two most similar neighbours (BottomUp~\cite{Keogh2001Online}, EAgglo~\cite{Matteson2013Nonparametric}); \emph{top-down} search (binary segmentation) does the reverse, placing the best single split and recursing on each part (BinSeg~\cite{Scott1974Cluster}, GGS~\cite{Hallac2016Greedy}, IGTS~\cite{Sadri2017Information}, ClaSP~\cite{Ermshaus2023ClaSP}). Both are greedy: apart from the local readjustments of GGS, they never revisit an earlier split or merge. \emph{Convex relaxation} fits a piecewise model with an $\ell_1$ penalty on its jumps, most of which become exactly zero (Prophet~\cite{Taylor2018Forecasting}). \emph{Optimal} search minimises a sum of segment costs exactly by dynamic programming, under a penalty per change or a fixed number of them (AMOC~\cite{Hinkley1970Inference}, PELT~\cite{Killick2012Optimal}, DynP~\cite{Truong2020Selective}), and \emph{Bayesian} search weighs every segmentation by its posterior probability (BOCD~\cite{Adams2007Bayesian}).

\noindent\textbf{How SD methods search.} SD must also decide which segments share a state. \emph{State-sequence} methods do both at once: as in a hidden Markov model, they give a state to every step or window and pay for each switch (AutoPlait~\cite{Matsubara2014AutoPlait}, HDP-HSMM~\cite{Johnson2013Bayesian}, TICC~\cite{Hallac2017Toeplitz}). \emph{Segment-then-group} methods cluster the segments of a CPD method (CLaP~\cite{Ermshaus2025CLaP}, on those of ClaSP). \emph{Window clustering} labels each window independently: time order is ignored (Hidalgo~\cite{Allegra2019Data}) or only enters the representation, learned contrastively from adjacent windows (Time2State~\cite{Wang2023Time2State}, E2USD~\cite{Lai2024E2Usd}). The optimal, Bayesian, and state-sequence searches share one dynamic programme over the position of the last change, which keeps the cheapest segmentation (optimal), sums over all of them (Bayesian), or also tracks the state of each segment (state sequence).

\noindent\textbf{What must stay the same.} Most methods compare the \emph{fit or likelihood of a model} of the observations (a mean, a covariance, an autoregression); others compare \emph{distributions} without a model (kernels, energy or Wasserstein distances, density ratios), the \emph{shapes of subsequences} (nearest neighbours, classifiers, symbolic patterns), or a \emph{representation learned} by a neural network. A cost on the mean, for instance, misses a change of variance that a kernel comparison detects. The penalised detectors (Window, BinSeg, BottomUp, PELT, DynP) take their cost as a parameter and span the first two rows.

\noindent\textbf{Main pool vs.\ extended scope.} Real-world series routinely reach $n = 10^4$-$10^6$ samples, where an $O(n^2)$ method may become intractable before it becomes uninteresting. The \textbf{main pool} (Section~\ref{sec:experiments}) gathers \textbf{21 algorithms}: 15 CPD and 6 SD methods. The \textbf{extended-scope pool} (Appendix~\ref{app:padding_strategy}) gathers \textbf{5 algorithms} that cannot run on every series: BOCD and EAgglo are at least quadratic, the neural training of TS-CP$^2$ and tGLAD exceeds our compute budget on parts of the benchmark, and Hidalgo is multivariate-only. We verify that it never beats the main pool on shared cells, so this scoping does not hide a state-of-the-art outlier.

\vspace{-0.1cm}
\subsection{Heterogeneous Time Series Collection}
\label{sec:datasets}
\vspace{-0.1cm}

\texttt{TSB-SEG} bundles \textbf{8 datasets} ($592$ labelled series) used for the systematic evaluation, and a 9th dataset, \textbf{PAMAP2}~\cite{Reiss2012Introducing} (9 series), kept aside as a long-series scalability case study (Section~\ref{sec:experiments}). The 8 main datasets contain univariate (TSSB~\cite{Ermshaus2023ClaSP}, UTSA~\cite{Gharghabi2018Domain}) and multivariate ($d \in [4,10]$) series (ActRecTut~\cite{Bulling2014ActRecTut}, HAS~\cite{Ermshaus2023Human}, MoCap~\cite{CMU2003MoCap}, USC-HAD~\cite{Zhang2012USCHAD}, SKAB~\cite{Katser2020SKAB}, PUMP~\cite{Gaugel2023PrecTime}), drawn from heterogeneous domains such as human activity recognition, industrial monitoring, healthcare, robotics, simulations, biology. Series lengths range from $240$ to $56{,}251$, the number of CPs from $0$ to $78$, and the number of distinct states from $1$ to $43$. Per-dataset statistics are reported in Table~\ref{tab:dataset_summary}, with more details in Appendix~\ref{app:datasets}. All labels are inherited from the original dataset authors, without re-annotation; Appendix~\ref{app:datasets} discusses their provenance and uncertainty.

\vspace{-0.1cm}
\subsection{Evaluation Modes}
\label{sec:exp_strategy}
\vspace{-0.1cm}

Every algorithm is evaluated under two complementary \emph{modes}. In \textbf{Default mode}, each algorithm runs one configuration $\theta_0$, recommended by its authors or the original implementation (Appendix~\ref{app:param_grids}). In \textbf{Parameter-search mode}, we sweep a grid of 2-4 values for the 2-4 most influential parameters, calibrated on pilot series in the same way for every algorithm (Appendix~\ref{app:param_grids}). For each (algorithm, dataset) pair, the configuration with the highest (lowest) dataset-level mean score, applied uniformly to every series of that dataset, gives the \emph{tuned} (\emph{worst-case}) score. Their gap $\Delta = P_{\text{tuned}} - P_{\text{worst}}$, the \emph{tuning gap}, exposes the algorithm's sensitivity to its hyperparameters. Selected on the evaluation labels, the tuned score is optimistic, as in the oracle experiment of TCPDBench~\cite{Burg2020Evaluation}, though less so since one configuration serves a whole dataset. A few methods (e.g.\ AutoPlait) are essentially parameter-free and use a single configuration. Orthogonally, the expected number of events $K \in \{n_{CP}, n_S\}$ is treated as a separate \emph{guidance} dimension since it is hard to estimate from data alone: every algorithm runs in \textbf{Non-Guided} (must infer $K$) and \textbf{Guided} (ground-truth $K$ provided) modes. Guided mode is an oracle analysis: knowing $K$ removes the hardest model-selection step, so it bounds what a method can reach; it is also directly relevant when $K$ is fixed by the protocol (scripted sessions, test cycles) or given by an expert, and the gap between the two modes measures the error due to estimating $K$. Methods that cannot ingest $K$ (e.g.\ PELT) have Guided $\equiv$ Non-Guided.

\vspace{-0.1cm}
\subsection{Evaluation Metrics}
\label{sec:metrics}
\vspace{-0.1cm}

We assess segmentation quality with task-specific metrics (Table~\ref{tab:metrics_overview}; formal definitions in Appendix~\ref{app:metric_defs}). For \textbf{CPD}, we report the standard \textbf{$F_1$-score} (CP localisation within a $1\%$ margin) and \textbf{Covering score} (segment alignment via maximum IoU). Both have known limitations~\cite{chavelli2025}: $F_1$ is discontinuous in the margin (a single-sample shift can flip a sparse-CP score from $1$ to $0$) and standard Covering is asymmetric and insensitive to over-segmentation. Event detection faces the same issue, which SoftED~\cite{sallesSoftEDMetricsSoft2024a} addresses with metrics that give partial credit to detections close to an event; in the same spirit, we complement $F_1$ and Covering with their natural \emph{soft} and \emph{symmetric} counterparts: \textbf{Gaussian $F_1$} (replacing the hit-or-miss indicator with a Gaussian reward kernel centred on each true CP, with a greedy bipartite matching to predictions) and \textbf{Bidirectional Covering (BiC)} (harmonic mean of the prediction-to-truth and truth-to-prediction Covering directions). For \textbf{SD}, the standard \textbf{ARI} and \textbf{AMI} disregard temporal ordering; following~\cite{chavelli2025} we additionally report \textbf{WARI} (a temporally-weighted ARI) and the interpretable \textbf{State Matching Score (SMS)}, which decomposes its penalty into four error types used in RQ2-RQ3: \emph{Delay} (a boundary placed slightly early or late), \emph{Transition} (a spurious state between two true segments), \emph{Isolation} (a spurious fragment inside a true segment, i.e.\ over-fragmentation) and \emph{Missing} (a true segment left undetected). BiC, Gaussian $F_1$ and the Delay class of SMS tolerate small boundary shifts, which limits the influence of annotation uncertainty. The main text reports \textbf{BiC} (CPD) and \textbf{SMS} (SD); the other metrics are reported in Appendix~\ref{app:secondary_metrics}.

\begin{table}[t]
    \centering
    \begin{minipage}[t]{0.48\linewidth}
        \caption{Datasets and their characteristics.\label{tab:dataset_summary}}
        \centering
        \resizebox{\linewidth}{!}{
        \begin{tabular}{llrrrrr}
        \toprule
        \textbf{Dataset} & \#TS & $\boldsymbol{n}$ & $\boldsymbol{d}$ & $\boldsymbol{n_{CP}}$ & $\boldsymbol{n_{S}}$ \\
        \midrule
        MoCap \cite{CMU2003MoCap} & 9 & 5-11k & 4 & 5-10 & 3-9 \\
        HAS \cite{Ermshaus2023Human} & 250 & 0.3-41k & 6-9 & 0-14 & 1-12 \\
        ActRecTut \cite{Bulling2014ActRecTut} & 2 & 31-33k & 10 & 42 & 6 \\
        PUMP \cite{Gaugel2023PrecTime} & 120 & 20-41k & 9 & 50-78 & 41-43 \\
        SKAB \cite{Katser2020SKAB} & 34 & 0.7-1.3k & 8 & 1-2 & 2 \\
        USC-HAD \cite{Zhang2012USCHAD} & 70 & 25-56k & 6 & 11 & 12 \\
        UTSA \cite{Gharghabi2018Domain} & 32 & 2-40k & 1 & 1-2 & 2-3 \\
        TSSB \cite{Ermshaus2023ClaSP} & 75 & 0.2-21k & 1 & 0-8 & 1-7 \\
        PAMAP2 \cite{Reiss2012Introducing} & 9 & 8-447k & 9 & 2-27 & 2-13 \\
        \bottomrule
        \end{tabular}
        }
    \end{minipage}\hfill
    \begin{minipage}[t]{0.465\linewidth}
        \caption{Metrics (primary ones shaded).\label{tab:metrics_overview}}
        \centering
        \resizebox{\linewidth}{!}{
        \begin{tabular}{@{}c l l@{}}
        \toprule
        \textbf{Task} & \textbf{Metric} & \textbf{Key Characteristic}\\
        \midrule
        \multirow{4}{*}{\textbf{CPD}}
            & $F_1$-score                                  & Localisation (1\% margin)  \\
            & Covering                                     & Segment alignment (max IoU)  \\
            & Gaussian $F_1$ & Smooth proximity reward \\
            \rowcolor{gray!20}
            \cellcolor{white}
            & \textbf{BiC}  & Symmetric bidirectional overlap  \\
        \midrule
        \multirow{4}{*}{\textbf{SD}}
            & ARI                       & Clustering agreement  \\
            & AMI                       & Adjusted mutual information  \\
            & WARI~\cite{chavelli2025}  & Temporally-weighted ARI  \\
            \rowcolor{gray!20}
            \cellcolor{white}
            & \textbf{SMS}~\cite{chavelli2025} & Interpretable state matching \\
        \bottomrule
        \end{tabular}
        }
    \end{minipage}
\end{table}

\vspace{-0.1cm}
\section{Experimental Evaluation} \label{sec:experiments}
\vspace{-0.2cm}

All experiments are reproducible with our open-source repository; compute environment and per-trial budgets are given in Appendix~\ref{app:compute}.

\vspace{-0.1cm}
\subsection{RQ1: Overall Accuracy Evaluation}
\vspace{-0.1cm}

\begin{figure}[tb]
    \centering
    \begin{subfigure}[t]{0.48\linewidth}
        \vspace{0pt}
        \includegraphics[width=\linewidth]{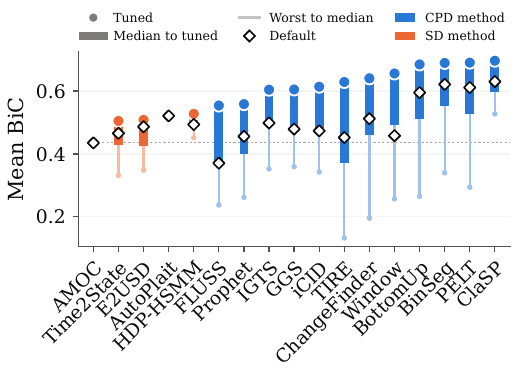}
        \caption{\emph{Non-Guided} strip plot (BiC)}
        \label{fig:cpd_strip_ng}
    \end{subfigure}\hfill
    \begin{subfigure}[t]{0.50\linewidth}
        \vspace{0pt}
        \includegraphics[width=\linewidth]{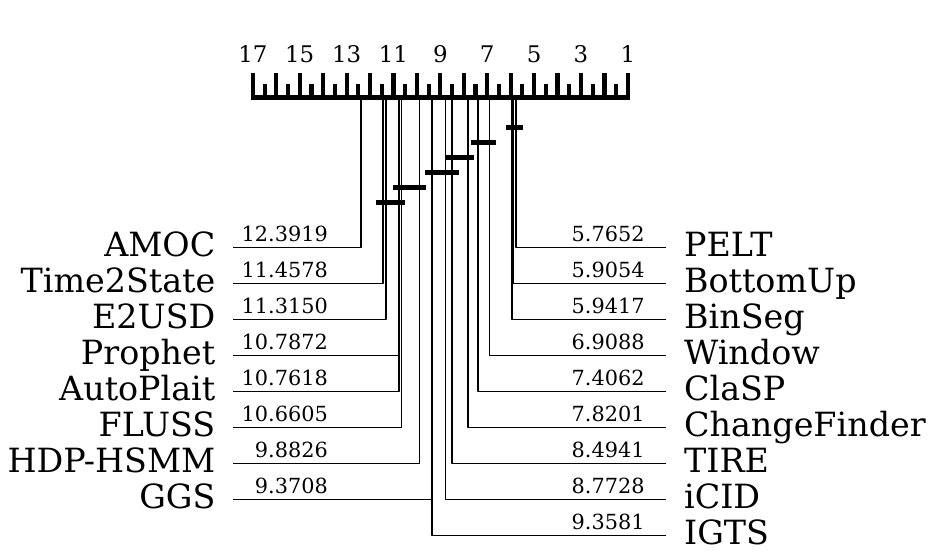}
        \vspace{7pt}
        \caption{\emph{Non-Guided} critical-difference diagram}
        \label{fig:cpd_cd_ng}
    \end{subfigure}\\[1em]
    \begin{subfigure}[t]{0.48\linewidth}
        \includegraphics[width=\linewidth]{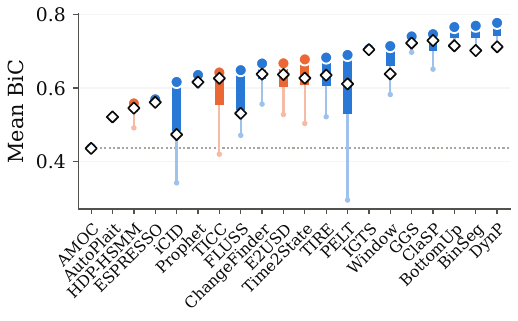}
        \caption{\emph{Guided} strip plot (BiC)}
        \label{fig:cpd_strip_g}
    \end{subfigure}\hfill
    \begin{subfigure}[t]{0.50\linewidth}
        \includegraphics[width=\linewidth]{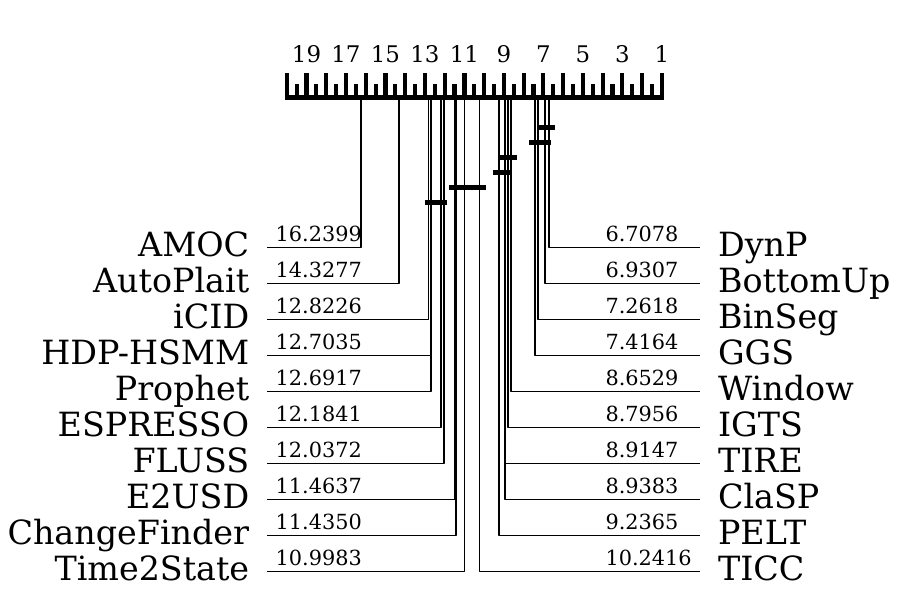}
        \caption{\emph{Guided} critical-difference diagram}
        \label{fig:cpd_cd_g}
    \end{subfigure}
    \caption{\textbf{CPD performance - \emph{Non-Guided} and \emph{Guided} modes.} Strip plots: mean over the 592 series of the tuned configuration (dot), the median and worst-case ones (thick body, thin stem) and the default one (\(\lozenge\)); colour: CPD or SD method; dotted line: AMOC. CD diagrams: mean ranks; bars join rank-consecutive methods that a two-sided Wilcoxon signed-rank test with Holm correction does not separate ($\alpha = 0.1$).}
    \label{fig:cpd_main}
\end{figure}

We evaluate the 21 main-pool algorithms across the 8 main \texttt{TSB-SEG} datasets ($592$ series). The SD methods are also scored on CPD, through the change points their labels induce, except CLaP, which labels the segments of ClaSP and is not scored separately. Three methods (DynP, ESPRESSO, TICC) require the number of change points or states as input and are reported only in the \emph{Guided} mode. For each (algorithm, dataset) pair, we report the \emph{tuned}, median, and \emph{worst-case} configurations (Section~\ref{sec:exp_strategy}) and the \emph{default} one. Figures~\ref{fig:cpd_main} and~\ref{fig:sd_main} report CPD and SD results in both supervision modes; per-dataset and per-domain views are in Appendices~\ref{app:domain_spiders}-\ref{app:rq1_extra}.

\textbf{CPD.} The critical-difference diagram (Figure~\ref{fig:cpd_cd_ng}) places the penalised detectors \textbf{PELT}, \textbf{BottomUp} and \textbf{BinSeg} in a statistically tied top clique (mean ranks $5.77$-$5.94$), followed by two tied pairs, \textbf{Window} and \textbf{ClaSP}, then ChangeFinder and TIRE, TIRE being tied with iCID, IGTS and GGS; the last two cliques (HDP-HSMM, FLUSS, AutoPlait; Prophet, E2USD, Time2State) precede the single-CP reference \textbf{AMOC}. The SD methods, scored on their induced change points, are thus the weakest Non-Guided CPD detectors: lowest mean tuned BiC after AMOC ($0.51$-$0.53$), and significantly behind every CPD method but FLUSS and Prophet. In mean tuned BiC, ClaSP ($0.70$) and the three penalised detectors ($0.68$-$0.69$) are level: among CPD methods, ClaSP leads on TSSB, HAS and UTSA, GGS on MoCap and TIRE on ActRecTut, the penalised detectors (Window included) on PUMP, SKAB and USC-HAD. The mean score and the mean rank disagree because ClaSP is a highly feature-dependent method: it is the strict best on $18\%$ of the series, more than any other, but ranks in the bottom half of the methods on $39\%$ of them (against $19$-$20\%$ for the top clique), so its large wins lift its mean while its frequent losses lower its rank. The strip plot (Figure~\ref{fig:cpd_strip_ng}) reveals a sharply heterogeneous \emph{tuning gap}: TIRE, ChangeFinder, BottomUp, Window, PELT and BinSeg span wide tuned-to-worst ranges ($\Delta \approx 0.35$-$0.50$), so a poor configuration can drive them below AMOC; FLUSS, Prophet, iCID, IGTS and GGS sit in between ($\Delta \approx 0.25$-$0.32$), while ClaSP stays well above AMOC even in its worst configuration ($\Delta \approx 0.17$). Providing $n_{CP}$ (\emph{Guided}, Figures~\ref{fig:cpd_strip_g}-\ref{fig:cpd_cd_g}) helps unevenly: Time2State, E2USD and GGS gain $+0.14$ to $+0.17$ in mean tuned BiC, IGTS, FLUSS, BinSeg, BottomUp, Prophet, Window, TIRE and ClaSP $+0.05$ to $+0.10$, HDP-HSMM, ChangeFinder and iCID at most $+0.03$, while PELT, AMOC and AutoPlait, which cannot use $n_{CP}$, are unchanged. DynP, which requires $n_{CP}$, takes the lead (tuned BiC $0.78$, mean rank $6.71$), statistically tied with BottomUp, itself tied with BinSeg and GGS. Among the methods that require $n_{CP}$, TICC ($0.64$) comes closest to the leaders, while ESPRESSO ($0.57$) stays in the lower half.

\textbf{SD.} The picture is sharper: \textbf{CLaP} attains the lowest mean rank ($2.32$), significantly ahead of the four other methods (mean ranks $3.12$-$3.23$), of which the test separates only Time2State from the rest (Figure~\ref{fig:sd_cd_stack}, top). The strip plot (Figure~\ref{fig:sd_strip_ng}) confirms CLaP's lead: it achieves the highest mean tuned SMS ($0.63$) with a narrow tuning gap, while Time2State and E2USD trail it ($0.50$-$0.51$) with wider gaps, and HDP-HSMM and AutoPlait have the lowest means ($0.48$). Adding the ground-truth number of states (Figures~\ref{fig:sd_strip_g} and~\ref{fig:sd_cd_stack}, bottom) brings TICC into the comparison; it remains below the leaders, and CLaP now leads alone (mean rank $2.63$, tuned SMS $0.71$), significantly ahead of Time2State ($2.92$). Notably, CLaP, the strongest SD method, labels the segments found by ClaSP, a CPD method: in such segment-then-group pipelines, progress in CPD carries over to SD, up to the labeling stage, which remains the bottleneck where states recur (RQ2).

\textbf{Beyond the means.} Two views in Appendix~\ref{app:rq1_extra} complete this picture. Picking the best method of each series would raise the mean score by $0.11$ to $0.16$ over the single best method in every task and mode, so even the leaders leave much to gain on many series. Without any tuning, the Default configurations keep largely the same leaders.

\begin{figure}[tb]
    \centering
    \begin{subfigure}[c]{0.24\linewidth}
        \includegraphics[width=\linewidth]{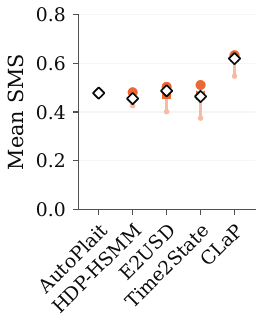}
        \caption{SD \emph{Non-Guided}}
        \label{fig:sd_strip_ng}
    \end{subfigure}\hfill
    \begin{subfigure}[c]{0.27\linewidth}
        \includegraphics[width=\linewidth]{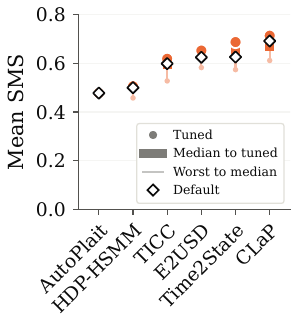}
        \caption{SD \emph{Guided}}
        \label{fig:sd_strip_g}
    \end{subfigure}\hfill
    \begin{subfigure}[c]{0.35\linewidth}
        \centering
        \includegraphics[width=\linewidth]{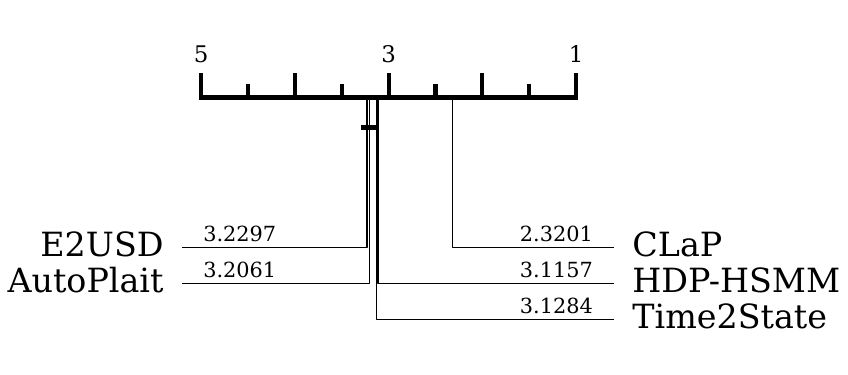}\\[2pt]
        \includegraphics[width=\linewidth]{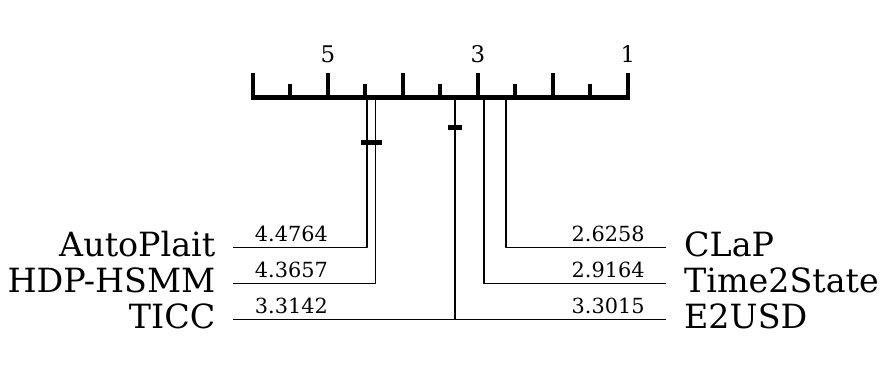}
        \caption{SD \emph{NG} (top) and \emph{G} (bottom).}
        \label{fig:sd_cd_stack}
    \end{subfigure}
    \caption{\textbf{SD performance - \emph{Non-Guided} and \emph{Guided} modes.} (a, b) SMS strip plots and (c) CD diagrams, as in Figure~\ref{fig:cpd_main}; in (c, top), HDP-HSMM is also tied with AutoPlait and E2USD. %
    }
    \label{fig:sd_main}
\end{figure}

\begin{tcolorbox}[colback=gray!5!white,colframe=gray!80!white,boxsep=2pt,top=2pt,bottom=2pt]
\textit{\textbf{Answer to RQ1:} No method dominates uniformly across tasks and modes. Per-dataset hyperparameter tuning is critical for many CPD methods. When available, providing the number of change points or states substantially improves performance for most methods.}
\end{tcolorbox}

\vspace{-0.1cm}
\subsection{RQ2: Change Point Detection vs.\ State Detection}
\vspace{-0.1cm}

\begin{figure}[tb]
    \centering
    \includegraphics[width=\linewidth]{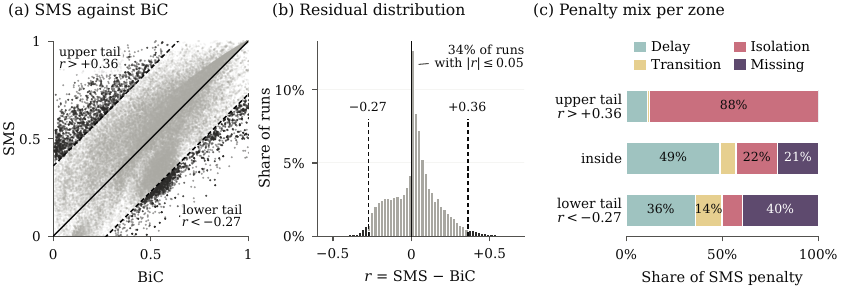}
    {\phantomsubcaption\label{fig:rq2_scatter}}
    {\phantomsubcaption\label{fig:rq2_residual}}
    {\phantomsubcaption\label{fig:rq2_error_mix}}
    \caption{\textbf{CPD vs.\ SD} on every run of the SD methods, both modes. (a) SMS against BiC; the dashed envelope holds $95\%$ of the runs, the tails outside it in black. (b) Distribution of $r=\mathrm{SMS}-\mathrm{BiC}$. (c) Mean share of each SMS error type per zone (default configurations).}
    \vspace{-0.3cm}
    \label{fig:regression_analysis}
\end{figure}

To probe the link between CPD and SD, we restrict attention to SD algorithms 
and depict in Figure~\ref{fig:rq2_scatter} every run \emph{both} as CPD (BiC) and as SD (SMS), pooling the Non-Guided and Guided modes; analysed separately, each mode leads to the same conclusions (Appendix~\ref{app:rq2_by_mode}). 
We report the residual $r = \mathrm{SMS} - \mathrm{BiC}$ in Figure~\ref{fig:rq2_residual}.

\textbf{CPD is a useful but imperfect proxy for SD.} The residual $r$ is sharply concentrated around $0$: $\sim$34\% of runs satisfy $|r|\leq 0.05$ and an empirical $\approx 95\%$ envelope holds $r\in[-0.27,+0.36]$. The agreement is statistical rather than deterministic: the scatter (Figure~\ref{fig:rq2_scatter}) shows non-negligible vertical spread for any given BiC. In practice, a strong CPD score reliably indicates a strong SD score (e.g.\ $\mathrm{BiC}\geq 0.8$ rarely yields $\mathrm{SMS}<0.6$), but exceptions exist in both directions. Since change-point labels are far cheaper than dense state annotations, BiC remains a useful and inexpensive proxy for SD performance \emph{on average}. The proxy is tightest when states do not recur ($|r|\leq 0.05$ for $41\%$ of runs, against $23\%$ when they do) and weakest on series with many recurring states such as PUMP (median $r=-0.18$), where correct boundaries do not guarantee correct labels. Conversely, the SD methods are the weakest CPD detectors (RQ1; CLaP takes its boundaries from ClaSP, a CPD method): their boundaries are a by-product of labelling windows (blocks of samples for HDP-HSMM), placed at that resolution, and lost whenever two adjacent segments are assigned the same state.

\textbf{The asymmetric tails reflect two distinct error regimes.} The SMS error breakdown (Figure~\ref{fig:rq2_error_mix}) clarifies what drives each tail. In the lower tail ($r<-0.27$), the SMS penalty is split between \emph{Missing} ($\approx 40\%$) and \emph{Delay} ($\approx 36\%$) errors: the segmentation is acceptable but the clustering stage mis-labels segments, so SMS collapses while BiC stays high. In the upper tail ($r>+0.36$), the penalty is dominated by \emph{Isolation} ($\approx 88\%$): heavy over-segmentation crashes BiC, but SMS only mildly penalises spurious fragments near a true boundary. Practical headroom therefore lies in clustering improvements (lower tail) and anti-fragmentation regularisation (upper tail).

\begin{tcolorbox}[colback=gray!5!white,colframe=gray!80!white,boxsep=2pt,top=2pt,bottom=2pt]
\textit{\textbf{Answer to RQ2:} BiC and SMS are strongly correlated: 
CPD accuracy is a useful (though imperfect) proxy for SD accuracy. The asymmetric tails decompose into two regimes: a clustering bottleneck ($r<-0.27$) and over-segmentation ($r>+0.36$).}
\end{tcolorbox}

\vspace{-0.1cm}
\subsection{RQ3: Actionable Guidelines}
\vspace{-0.1cm}

We now turn from accuracy to the choices a practitioner faces: runtime, tuning effort and the properties of the series at hand.

\begin{figure}[tb]
    \centering
    \includegraphics[width=\linewidth]{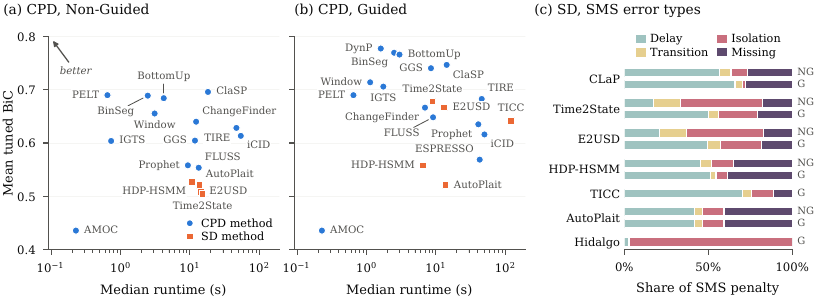}
    {\phantomsubcaption\label{fig:runtime_accuracy_trade_off}}
    {\phantomsubcaption}
    {\phantomsubcaption\label{fig:sms_breakdown}}
    \caption{\textbf{Runtime-accuracy trade-off and SD error breakdown.} (a, b) Mean tuned BiC against median runtime of the tuned configuration (circles: CPD methods; squares: SD methods). (c) Share of each SMS error type in the penalty of each SD method (default configuration; NG: Non-Guided, G: Guided).}
    \vspace{-0.3cm}
    \label{fig:rq3_runtime}
\end{figure}

\textbf{Runtime-accuracy trade-off.} Apart from AMOC, which places a single change point, the penalised detectors and IGTS are the fastest (median runtime of the tuned configuration at most $4.2$\,s); all methods but TICC have a median below $60$\,s, but the 90th percentile exceeds $500$\,s for ClaSP, CLaP, iCID, ESPRESSO and TICC. For CPD, the fastest methods are also among the most accurate (Figure~\ref{fig:rq3_runtime}): PELT reaches BiC $0.69$ in $0.65$\,s, BinSeg and BottomUp are within $0.01$ at $2$-$4$\,s, and only ClaSP scores higher ($0.70$, at $\approx 18$\,s); with $n_{CP}$ given, DynP, BinSeg and BottomUp lead in $1.6$-$3$\,s. For SD, all methods sit in the mid-range of runtimes (Appendix~\ref{app:runtime_sd}), where CLaP, the most accurate, takes a median of $25$-$29$\,s per series.

\textbf{Hyperparameter sensitivity.} A per-algorithm marginal-score view (Appendix~\ref{app:hyperparams}) shows that Time2State, E2USD and iCID are dominated by a single, monotone window-size knob (TIRE and Prophet without $K$ by their threshold) and are otherwise robust; penalty-based methods (Window, BinSeg, BottomUp, PELT) respond mainly to the penalty, peaking near $8\times$ the BIC penalty ($1\times$ for Window), the cost function mattering mostly at low penalties; only TICC exhibits genuine multi-axis sensitivity, so for the other methods a coarse sweep of a single parameter recovers most of the tuning gain. Providing $K$ flattens the response surface for most methods, partially absorbing the per-series tuning effort. Neither multivariate adaptation (ensembling or L2 reduction) dominates the other.

\textbf{Sensitivity to input features.} Spider plots (Appendix~\ref{app:domain_spiders}) and Spearman correlations between tuned score and structural features ($n,d,n_{CP},n_S$; Figure~\ref{fig:feature_sensitivity}) show that the penalised detectors are nearly insensitive to these features in Non-Guided mode ($|\rho| \leq 0.17$), whereas ClaSP excels on univariate series with few events (TSSB) but degrades on series with many events ($\rho \approx -0.5$) and on multivariate inputs ($\rho \approx -0.5$). Time2State and E2USD show the opposite pattern on CPD ($\rho \approx 0.5$ with event counts) and their SMS barely depends on $d$ or $n_{CP}$ ($|\rho| \leq 0.15$, against $\approx -0.55$ for CLaP), making them the most robust SD picks for complex multivariate series in Non-Guided mode.

\textbf{Performance on long series.} On PAMAP2 (up to $447$k samples per series; Appendix~\ref{app:pamap2_case_study}), only $17$ of the $26$ algorithms process all 9 series within the time budget; ClaSP, CLaP and tGLAD finish 1-5 of them and six methods none, so main-pool accuracy does not always carry over to long series. For CPD, the penalised detectors scale \emph{and} perform well: BinSeg, PELT and BottomUp reach tuned BiC $0.68$-$0.72$ Non-Guided ($0.71$-$0.73$ Guided, with DynP), on par with IGTS ($0.70$) and above GGS ($0.61$ Non-Guided). For SD, Time2State and E2USD scale and remain competitive (tuned SMS $0.53$-$0.56$).

\textbf{SMS error-type breakdown.} Decomposing the SMS penalty into its four error types (Figure~\ref{fig:sms_breakdown}) reveals two distinct error profiles: the strongest methods (CLaP, Time2State~G, TICC~G) are dominated by \emph{Delay} ($50$-$70\%$): they recover the right states but misplace boundaries by a small margin, a benign error in many applications; Hidalgo is dominated by \emph{Isolation} ($> 90\%$): its intrinsic-dimension labels over-fragment our series with few channels, whereas the state-sequence HDP-HSMM and AutoPlait split their penalty between \emph{Delay} ($\approx 42$-$51\%$) and \emph{Missing} ($\approx 35$-$41\%$): they under-segment rather than fragment. Without $K$, the window-clustering E2USD and Time2State are dominated by \emph{Isolation} ($46$-$48\%$), i.e.\ over-fragmentation, which supplying $K$ halves, mostly in favour of \emph{Delay}. Practically, residual SD headroom lies in \emph{boundary refinement} for the leaders and \emph{anti-fragmentation regularisation} for the rest.

\begin{tcolorbox}[colback=gray!5!white,colframe=gray!80!white,boxsep=2pt,top=2pt,bottom=2pt]
\textit{\textbf{Answer to RQ3:} For CPD, the fast penalised detectors (PELT, BinSeg, BottomUp; DynP when $K$ is known) are the safest choice and scale to long series; ClaSP is preferable on univariate series with few events. For SD, CLaP leads on simple series, Time2State and E2USD on complex multivariate and long ones. Tuning one window, penalty or threshold parameter captures most of the gain.}
\end{tcolorbox}

\vspace{-0.1cm}
\section{Conclusion and Limitations} \label{sec:conclusion}
\vspace{-0.2cm}

We introduced \texttt{TSB-SEG}, the first large-scale benchmark jointly evaluating Change Point Detection (CPD) and State Detection (SD), covering 26 algorithms and 592 time series through the open-source \texttt{tsseg} library. Our results show that (i) no method consistently dominates, highlighting the need for dataset-specific tuning; (ii) BiC and SMS are strongly correlated, making CPD a useful proxy for SD; and (iii) data characteristics should guide method selection.

\texttt{TSB-SEG} also has limitations. It covers univariate and moderately multivariate series, leaving highly multivariate settings largely unexplored, and its parameter grids, however broad, cannot cover every plausible configuration. It treats change points and states as discrete events, whereas real transitions are often gradual and states can be mixed or hierarchical, which our soft metrics only partly account for. Finally, it focuses on offline segmentation and does not yet address online or streaming settings.

These limitations point to several directions for future work: extending the benchmark to highly multivariate and streaming data, adapting time-series foundation models to segmentation, and automating method selection from data characteristics, for instance by meta-learning.

\begin{ack}
The authors are grateful to the CLEPS infrastructure from Inria of Paris for providing resources and support. John Paparrizos was partially supported by the National Science Foundation under Award No.~2543459.
\end{ack}

\bibliographystyle{unsrtnat}
\bibliography{main}

\begin{thebibliography}{174}
\providecommand{\natexlab}[1]{#1}
\providecommand{\url}[1]{\texttt{#1}}
\expandafter\ifx\csname urlstyle\endcsname\relax
  \providecommand{\doi}[1]{doi: #1}\else
  \providecommand{\doi}{doi: \begingroup \urlstyle{rm}\Url}\fi

\bibitem[Lara and Labrador(2013)]{Lara2013Survey}
Oscar~D. Lara and Miguel~A. Labrador.
\newblock A survey on human activity recognition using wearable sensors.
\newblock \emph{IEEE Communications Surveys \& Tutorials}, 15:\penalty0
  1192--1209, 2013.

\bibitem[Kanawaday and Sane(2017)]{Kanawaday2017Machine}
Ameeth Kanawaday and Aditya Sane.
\newblock Machine learning for predictive maintenance of industrial machines
  using {IoT} sensor data.
\newblock \emph{2017 8th IEEE International Conference on Software Engineering
  and Service Science (ICSESS)}, pages 87--90, 2017.

\bibitem[Woollam et~al.(2022)Woollam, M{\"u}nchmeyer, Tilmann, Rietbrock,
  Lange, Bornstein, Diehl, Giunchi, Haslinger, Jozinovi{\'c}, Michelini, Saul,
  and Soto]{Woollam2022SeisBench}
J.~H. Woollam, Jannes M{\"u}nchmeyer, Frederik Tilmann, Andreas Rietbrock,
  Dietrich Lange, Thomas Bornstein, Tobias Diehl, Carlo Giunchi, Florian
  Haslinger, Dario Jozinovi{\'c}, Alberto Michelini, Joachim Saul, and Hugo
  Soto.
\newblock {SeisBench}—a toolbox for machine learning in seismology.
\newblock \emph{Seismological Research Letters}, 2022.

\bibitem[Ermshaus et~al.(2023{\natexlab{a}})Ermshaus, Sch{\"a}fer, and
  Leser]{Ermshaus2023ClaSP}
Arik Ermshaus, Patrick Sch{\"a}fer, and Ulf Leser.
\newblock {ClaSP}: parameter-free time series segmentation.
\newblock \emph{Data Mining and Knowledge Discovery}, 37:\penalty0 1262 --
  1300, 2023{\natexlab{a}}.

\bibitem[Paparrizos et~al.(2016{\natexlab{a}})Paparrizos, White, and
  Horvitz]{paparrizos2016screening}
John Paparrizos, Ryen~W White, and Eric Horvitz.
\newblock Screening for pancreatic adenocarcinoma using signals from web search
  logs: Feasibility study and results.
\newblock \emph{Journal of oncology practice}, 12\penalty0 (8):\penalty0
  737--744, 2016{\natexlab{a}}.

\bibitem[Paparrizos(2018)]{paparrizos2018fast}
Ioannis Paparrizos.
\newblock \emph{Fast, scalable, and accurate algorithms for time-series
  analysis}.
\newblock PhD thesis, Columbia University, 2018.

\bibitem[Paparrizos et~al.(2016{\natexlab{b}})Paparrizos, White, and
  Horvitz]{paparrizos2016detecting}
John Paparrizos, Ryen~W White, and Eric Horvitz.
\newblock Detecting devastating diseases in search logs.
\newblock In \emph{Proceedings of the 22nd ACM SIGKDD International Conference
  on Knowledge Discovery and Data Mining}, pages 559--568, 2016{\natexlab{b}}.

\bibitem[Liu et~al.(2023)Liu, Mangla, Shaowang, Zhao, Paparrizos, Krishnan, and
  Feamster]{liu2023amir}
Shinan Liu, Tarun Mangla, Ted Shaowang, Jinjin Zhao, John Paparrizos, Sanjay
  Krishnan, and Nick Feamster.
\newblock Amir: Active multimodal interaction recognition from video and
  network traffic in connected environments.
\newblock \emph{Proceedings of the ACM on Interactive, Mobile, Wearable and
  Ubiquitous Technologies}, 7\penalty0 (1):\penalty0 1--26, 2023.

\bibitem[Liu et~al.(2021)Liu, Jiang, Paparrizos, and Elmore]{liu2021decomposed}
Chunwei Liu, Hao Jiang, John Paparrizos, and Aaron~J Elmore.
\newblock Decomposed bounded floats for fast compression and queries.
\newblock \emph{Proceedings of the VLDB Endowment}, 14\penalty0 (11):\penalty0
  2586--2598, 2021.

\bibitem[Liu et~al.(2024{\natexlab{a}})Liu, Paparrizos, and
  Elmore]{liu2024adaedge}
Chunwei Liu, John Paparrizos, and Aaron~J Elmore.
\newblock Adaedge: A dynamic compression selection framework for resource
  constrained devices.
\newblock In \emph{2024 IEEE 40th International Conference on Data Engineering
  (ICDE)}, pages 1506--1519. IEEE, 2024{\natexlab{a}}.

\bibitem[McKeown et~al.(2016)McKeown, Daume~III, Chaturvedi, Paparrizos,
  Thadani, Barrio, Biran, Bothe, Collins, Fleischmann,
  et~al.]{mckeown2016predicting}
Kathy McKeown, Hal Daume~III, Snigdha Chaturvedi, John Paparrizos, Kapil
  Thadani, Pablo Barrio, Or~Biran, Suvarna Bothe, Michael Collins, Kenneth~R
  Fleischmann, et~al.
\newblock Predicting the impact of scientific concepts using full-text
  features.
\newblock \emph{Journal of the Association for Information Science and
  Technology}, 67\penalty0 (11):\penalty0 2684--2696, 2016.

\bibitem[Paparrizos et~al.(2021)Paparrizos, Liu, Barbarioli, Hwang, Edian,
  Elmore, Franklin, and Krishnan]{paparrizos2021vergedb}
John Paparrizos, Chunwei Liu, Bruno Barbarioli, Johnny Hwang, Ikraduya Edian,
  Aaron~J Elmore, Michael~J Franklin, and Sanjay Krishnan.
\newblock Vergedb: A database for iot analytics on edge devices.
\newblock In \emph{CIDR}, 2021.

\bibitem[Goel et~al.(2016)Goel, Soni, Goyal, Paparrizos, Wallach, Diaz, and
  Eisenstein]{goel2016social}
Rahul Goel, Sandeep Soni, Naman Goyal, John Paparrizos, Hanna Wallach, Fernando
  Diaz, and Jacob Eisenstein.
\newblock The social dynamics of language change in online networks.
\newblock In \emph{International conference on social informatics}, pages
  41--57. Springer, 2016.

\bibitem[Krishnan et~al.(2019)Krishnan, Elmore, Franklin, Paparrizos, Shang,
  Dziedzic, and Liu]{krishnan2019artificial}
Sanjay Krishnan, Aaron~J Elmore, Michael Franklin, John Paparrizos, Zechao
  Shang, Adam Dziedzic, and Rui Liu.
\newblock Artificial intelligence in resource-constrained and shared
  environments.
\newblock \emph{ACM SIGOPS Operating Systems Review}, 53\penalty0 (1):\penalty0
  1--6, 2019.

\bibitem[Jiang et~al.(2020)Jiang, Liu, Jin, Paparrizos, and
  Elmore]{jiang2020pids}
Hao Jiang, Chunwei Liu, Qi~Jin, John Paparrizos, and Aaron~J Elmore.
\newblock Pids: attribute decomposition for improved compression and query
  performance in columnar storage.
\newblock \emph{Proceedings of the VLDB Endowment}, 13\penalty0 (6):\penalty0
  925--938, 2020.

\bibitem[Jiang et~al.(2021)Jiang, Liu, Paparrizos, Chien, Ma, and
  Elmore]{jiang2021good}
Hao Jiang, Chunwei Liu, John Paparrizos, Andrew~A Chien, Jihong Ma, and Aaron~J
  Elmore.
\newblock Good to the last bit: Data-driven encoding with codecdb.
\newblock In \emph{Proceedings of the 2021 International Conference on
  Management of Data}, pages 843--856, 2021.

\bibitem[Dziedzic et~al.(2019)Dziedzic, Paparrizos, Krishnan, Elmore, and
  Franklin]{dziedzic2019band}
Adam Dziedzic, John Paparrizos, Sanjay Krishnan, Aaron Elmore, and Michael
  Franklin.
\newblock Band-limited training and inference for convolutional neural
  networks.
\newblock In \emph{International Conference on Machine Learning}, pages
  1745--1754. PMLR, 2019.

\bibitem[Paparrizos et~al.(2023{\natexlab{a}})Paparrizos, Wu, Elmore,
  Faloutsos, and Franklin]{paparrizos2023accelerating}
John Paparrizos, Kaize Wu, Aaron Elmore, Christos Faloutsos, and Michael~J
  Franklin.
\newblock Accelerating similarity search for elastic measures: A study and new
  generalization of lower bounding distances.
\newblock \emph{Proceedings of the VLDB Endowment}, 16\penalty0 (8):\penalty0
  2019--2032, 2023{\natexlab{a}}.

\bibitem[Paparrizos et~al.(2023{\natexlab{b}})Paparrizos, Liu, Elmore, and
  Franklin]{paparrizos2023querying}
John Paparrizos, Chunwei Liu, Aaron~J Elmore, and Michael~J Franklin.
\newblock Querying time-series data: A comprehensive comparison of distance
  measures.
\newblock \emph{Data Engineering}, page~69, 2023{\natexlab{b}}.

\bibitem[Paparrizos et~al.(2022{\natexlab{a}})Paparrizos, Edian, Liu, Elmore,
  and Franklin]{paparrizos2022fast}
John Paparrizos, Ikraduya Edian, Chunwei Liu, Aaron~J Elmore, and Michael~J
  Franklin.
\newblock Fast adaptive similarity search through variance-aware quantization.
\newblock In \emph{2022 IEEE 38th International Conference on Data Engineering
  (ICDE)}, pages 2969--2983. IEEE, 2022{\natexlab{a}}.

\bibitem[Jeung et~al.(2010)Jeung, Sarni, Paparrizos, Sathe, Aberer, Dawes,
  Papaioannou, and Lehning]{jeung2010effective}
Hoyoung Jeung, Sofiane Sarni, Ioannis Paparrizos, Saket Sathe, Karl Aberer,
  Nicholas Dawes, Thanasis~G Papaioannou, and Michael Lehning.
\newblock Effective metadata management in federated sensor networks.
\newblock In \emph{2010 IEEE International Conference on Sensor Networks,
  Ubiquitous, and Trustworthy Computing}, pages 107--114. IEEE, 2010.

\bibitem[Paparrizos et~al.(2011)Paparrizos, Jeung, and
  Aberer]{paparrizos2011advanced}
Ioannis Paparrizos, Hoyoung Jeung, and Karl Aberer.
\newblock Advanced search, visualization and tagging of sensor metadata.
\newblock In \emph{2011 IEEE 27th International Conference on Data
  Engineering}, pages 1356--1359. IEEE, 2011.

\bibitem[Hishida et~al.(2025)Hishida, Liu, Paparrizos, and
  Elmore]{hishida2025beyond}
Kaisei Hishida, Chunwei Liu, John Paparrizos, and Aaron Elmore.
\newblock Beyond compression: A comprehensive evaluation of lossless
  floating-point compression.
\newblock \emph{Proceedings of the VLDB Endowment}, 18\penalty0 (11):\penalty0
  4396--4409, 2025.

\bibitem[d’Hondt et~al.(2024)d’Hondt, Papapetrou, and
  Paparrizos]{d2024beyond}
Jens~E d’Hondt, Odysseas Papapetrou, and John Paparrizos.
\newblock Beyond the dimensions: A structured evaluation of multivariate time
  series distance measures.
\newblock In \emph{2024 IEEE 40th International Conference on Data Engineering
  Workshops (ICDEW)}, pages 107--112. IEEE, 2024.

\bibitem[Sch\"{a}fer(2015)]{boss2015}
Patrick Sch\"{a}fer.
\newblock The {BOSS} is concerned with time series classification in the
  presence of noise.
\newblock \emph{Data Min. Knowl. Discov.}, 29\penalty0 (6):\penalty0
  1505–1530, November 2015.
\newblock ISSN 1384-5810.
\newblock \doi{10.1007/s10618-014-0377-7}.
\newblock URL \url{https://doi.org/10.1007/s10618-014-0377-7}.

\bibitem[Schäfer and Leser(2017)]{Sch_fer_2017}
Patrick Schäfer and Ulf Leser.
\newblock Fast and accurate time series classification with {WEASEL}.
\newblock In \emph{Proceedings of the 2017 ACM on Conference on Information and
  Knowledge Management}, CIKM ’17. ACM, November 2017.
\newblock \doi{10.1145/3132847.3132980}.
\newblock URL \url{http://dx.doi.org/10.1145/3132847.3132980}.

\bibitem[Lines et~al.(2016)Lines, Taylor, and Bagnall]{hivecote}
Jason Lines, Sarah Taylor, and Anthony Bagnall.
\newblock {HIVE-COTE}: The hierarchical vote collective of transformation-based
  ensembles for time series classification.
\newblock In \emph{2016 IEEE 16th International Conference on Data Mining
  (ICDM)}, pages 1041--1046, 2016.
\newblock \doi{10.1109/ICDM.2016.0133}.

\bibitem[Middlehurst et~al.(2024{\natexlab{a}})Middlehurst, Sch{\"a}fer, and
  Bagnall]{middlehurst2024bake}
Matthew Middlehurst, Patrick Sch{\"a}fer, and Anthony Bagnall.
\newblock Bake off redux: a review and experimental evaluation of recent time
  series classification algorithms.
\newblock \emph{Data Mining and Knowledge Discovery}, 38\penalty0 (4):\penalty0
  1958--2031, 2024{\natexlab{a}}.

\bibitem[Paparrizos and Gravano(2015)]{paparrizos2015k}
John Paparrizos and Luis Gravano.
\newblock {k-Shape}: Efficient and accurate clustering of time series.
\newblock In \emph{Proceedings of the 2015 ACM SIGMOD International Conference
  on Management of Data}, pages 1855--1870. ACM, 2015.

\bibitem[Paparrizos et~al.(2024{\natexlab{a}})Paparrizos, Yang, and
  Li]{paparrizos2024bridging}
John Paparrizos, Fan Yang, and Haojun Li.
\newblock Bridging the gap: A decade review of time-series clustering methods.
\newblock \emph{arXiv preprint arXiv:2412.20582}, 2024{\natexlab{a}}.

\bibitem[Paparrizos and Franklin(2019)]{paparrizos2019grail}
John Paparrizos and Michael~J Franklin.
\newblock {GRAIL}: efficient time-series representation learning.
\newblock \emph{Proceedings of the VLDB Endowment}, 12\penalty0 (11):\penalty0
  1762--1777, 2019.

\bibitem[Boniol et~al.(2024{\natexlab{a}})Boniol, Liu, Huang, Palpanas, and
  Paparrizos]{boniol2024divetimeseriesanomalydetection}
Paul Boniol, Qinghua Liu, Mingyi Huang, Themis Palpanas, and John Paparrizos.
\newblock Dive into time-series anomaly detection: A decade review,
  2024{\natexlab{a}}.
\newblock URL \url{https://arxiv.org/abs/2412.20512}.

\bibitem[Paparrizos et~al.(2022{\natexlab{b}})Paparrizos, Kang, Boniol, Tsay,
  Palpanas, and Franklin]{paparrizos2022tsb}
John Paparrizos, Yuhao Kang, Paul Boniol, Ruey~S Tsay, Themis Palpanas, and
  Michael~J Franklin.
\newblock {TSB-UAD}: an end-to-end benchmark suite for univariate time-series
  anomaly detection.
\newblock \emph{Proceedings of the VLDB Endowment}, 15\penalty0 (8):\penalty0
  1697--1711, 2022{\natexlab{b}}.

\bibitem[Liu and Paparrizos(2024)]{liu2024elephant}
Qinghua Liu and John Paparrizos.
\newblock The elephant in the room: Towards a reliable time-series anomaly
  detection benchmark.
\newblock \emph{Advances in Neural Information Processing Systems},
  37:\penalty0 108231--108261, 2024.

\bibitem[Senin et~al.(2014)Senin, Lin, Wang, Oates, Gandhi, Boedihardjo, Chen,
  Frankenstein, and Lerner]{senin2014grammarviz}
Pavel Senin, Jessica Lin, Xing Wang, Tim Oates, Sunil Gandhi, Arnold~P
  Boedihardjo, Crystal Chen, Susan Frankenstein, and Manfred Lerner.
\newblock {GrammarViz} 2.0: a tool for grammar-based pattern discovery in time
  series.
\newblock In \emph{Machine Learning and Knowledge Discovery in Databases:
  European Conference, ECML PKDD 2014, Nancy, France, September 15-19, 2014.
  Proceedings, Part III 14}, pages 468--472. Springer, 2014.

\bibitem[Paparrizos et~al.(2022{\natexlab{c}})Paparrizos, Boniol, Palpanas,
  Tsay, Elmore, and Franklin]{paparrizos2022vus}
John Paparrizos, Paul Boniol, Themis Palpanas, Ruey~S Tsay, Aaron Elmore, and
  Michael~J Franklin.
\newblock Volume under the surface: a new accuracy evaluation measure for
  time-series anomaly detection.
\newblock \emph{Proceedings of the VLDB Endowment}, 15\penalty0 (11):\penalty0
  2774--2787, 2022{\natexlab{c}}.

\bibitem[Boniol and Palpanas(2020)]{10.14778/3407790.3407792}
Paul Boniol and Themis Palpanas.
\newblock {Series2Graph}: graph-based subsequence anomaly detection for time
  series.
\newblock \emph{Proc. VLDB Endow.}, 13\penalty0 (12):\penalty0 1821–1834,
  July 2020.
\newblock ISSN 2150-8097.
\newblock \doi{10.14778/3407790.3407792}.
\newblock URL \url{https://doi.org/10.14778/3407790.3407792}.

\bibitem[Lin et~al.(2003)Lin, Keogh, Lonardi, and Chiu]{lin2003symbolic}
Jessica Lin, Eamonn Keogh, Stefano Lonardi, and Bill Chiu.
\newblock A symbolic representation of time series, with implications for
  streaming algorithms.
\newblock In \emph{Proceedings of the 8th ACM SIGMOD workshop on Research
  issues in data mining and knowledge discovery}, pages 2--11, 2003.

\bibitem[Shieh and Keogh(2008)]{shieh2008isax}
Jin Shieh and Eamonn Keogh.
\newblock {iSAX}: indexing and mining terabyte sized time series.
\newblock In \emph{Proceedings of the 14th ACM SIGKDD International Conference
  on Knowledge Discovery and Data Mining}, KDD '08, page 623–631, New York,
  NY, USA, 2008. Association for Computing Machinery.
\newblock ISBN 9781605581934.
\newblock \doi{10.1145/1401890.1401966}.
\newblock URL \url{https://doi.org/10.1145/1401890.1401966}.

\bibitem[Sch\"{a}fer and H\"{o}gqvist(2012)]{SFA2012}
Patrick Sch\"{a}fer and Mikael H\"{o}gqvist.
\newblock {SFA}: a symbolic {Fourier} approximation and index for similarity
  search in high dimensional datasets.
\newblock In \emph{Proceedings of the 15th International Conference on
  Extending Database Technology}, EDBT '12, page 516–527, New York, NY, USA,
  2012. Association for Computing Machinery.
\newblock ISBN 9781450307901.
\newblock \doi{10.1145/2247596.2247656}.
\newblock URL \url{https://doi.org/10.1145/2247596.2247656}.

\bibitem[Paparrizos et~al.(2024{\natexlab{b}})Paparrizos, Li, Yang, Wu,
  d'Hondt, and Papapetrou]{paparrizos2024survey}
John Paparrizos, Haojun Li, Fan Yang, Kaize Wu, Jens~E d'Hondt, and Odysseas
  Papapetrou.
\newblock A survey on time-series distance measures.
\newblock \emph{arXiv preprint arXiv:2412.20574}, 2024{\natexlab{b}}.

\bibitem[d’Hondt et~al.(2025)d’Hondt, Li, Yang, Papapetrou, and
  Paparrizos]{dhondt2025structured}
Jens~E d’Hondt, Haojun Li, Fan Yang, Odysseas Papapetrou, and John
  Paparrizos.
\newblock A structured study of multivariate time-series distance measures.
\newblock In \emph{Proceedings of the 2025 ACM SIGMOD international conference
  on management of data}, 2025.

\bibitem[Li et~al.(2012)Li, Lin, and Oates]{yuan2012visual}
Yuan Li, Jessica Lin, and Tim Oates.
\newblock Visualizing variable-length time series motifs.
\newblock In \emph{Proceedings of the 2012 SIAM International Conference on
  Data Mining (SDM)}, pages 895--906. SIAM, 2012.
\newblock \doi{10.1137/1.9781611972825.77}.

\bibitem[Feremans et~al.(2025)Feremans, Sch{\"a}fer, and
  Meert]{feremans2025motiplus}
Len Feremans, Patrick Sch{\"a}fer, and Wannes Meert.
\newblock {MotiPlus} and {MotiSet}: Discovering the best set of motiflets in
  time series.
\newblock In \emph{Joint European Conference on Machine Learning and Knowledge
  Discovery in Databases}, pages 473--489. Springer, 2025.

\bibitem[Yang and Paparrizos(2025{\natexlab{a}})]{Yang2025SPARTAN}
Fan Yang and John Paparrizos.
\newblock {SPARTAN}: Data-adaptive symbolic time-series approximation.
\newblock \emph{Proc. ACM Manag. Data}, 3\penalty0 (3), June
  2025{\natexlab{a}}.
\newblock \doi{10.1145/3725357}.
\newblock URL \url{https://doi.org/10.1145/3725357}.

\bibitem[Yang and Paparrizos(2025{\natexlab{b}})]{yang2025sail}
Fan Yang and John Paparrizos.
\newblock {SAIL}: A voyage to symbolic approximation solutions for time-series
  analysis.
\newblock \emph{Proceedings of the VLDB Endowment}, 18\penalty0 (12):\penalty0
  5419--5422, 2025{\natexlab{b}}.

\bibitem[Nie et~al.(2023)Nie, H.~Nguyen, Sinthong, and
  Kalagnanam]{Yuqietal-2023-PatchTST}
Yuqi Nie, Nam H.~Nguyen, Phanwadee Sinthong, and Jayant Kalagnanam.
\newblock A time series is worth 64 words: Long-term forecasting with
  transformers.
\newblock In \emph{International Conference on Learning Representations}, 2023.

\bibitem[Ansari et~al.(2024)Ansari, Stella, Turkmen, Zhang, Mercado, Shen,
  Shchur, Rangapuram, Pineda~Arango, Kapoor, Zschiegner, Maddix, Wang, Mahoney,
  Torkkola, Gordon~Wilson, Bohlke-Schneider, and Wang]{ansari2024chronos}
Abdul~Fatir Ansari, Lorenzo Stella, Caner Turkmen, Xiyuan Zhang, Pedro Mercado,
  Huibin Shen, Oleksandr Shchur, Syama~Syndar Rangapuram, Sebastian
  Pineda~Arango, Shubham Kapoor, Jasper Zschiegner, Danielle~C. Maddix, Hao
  Wang, Michael~W. Mahoney, Kari Torkkola, Andrew Gordon~Wilson, Michael
  Bohlke-Schneider, and Yuyang Wang.
\newblock Chronos: Learning the language of time series.
\newblock \emph{arXiv preprint arXiv:2403.07815}, 2024.

\bibitem[Liu et~al.(2024{\natexlab{b}})Liu, Xu, Zhao, Kong, Prabhakar~Kamarthi,
  Sasanur, Sharma, Cui, Wen, Zhang, et~al.]{liu2024timemmd}
Haoxin Liu, Shangqing Xu, Zhiyuan Zhao, Lingkai Kong, Harshavardhan
  Prabhakar~Kamarthi, Aditya Sasanur, Megha Sharma, Jiaming Cui, Qingsong Wen,
  Chao Zhang, et~al.
\newblock {Time-MMD}: Multi-domain multimodal dataset for time series analysis.
\newblock \emph{Advances in Neural Information Processing Systems},
  37:\penalty0 77888--77933, 2024{\natexlab{b}}.

\bibitem[Liu et~al.(2024{\natexlab{c}})Liu, Zhao, Wang, Kamarthi, and
  Prakash]{liu-etal-2024-lstprompt}
Haoxin Liu, Zhiyuan Zhao, Jindong Wang, Harshavardhan Kamarthi, and B.~Aditya
  Prakash.
\newblock {LSTP}rompt: Large language models as zero-shot time series
  forecasters by long-short-term prompting.
\newblock In Lun-Wei Ku, Andre Martins, and Vivek Srikumar, editors,
  \emph{Findings of the Association for Computational Linguistics: ACL 2024},
  pages 7832--7840, Bangkok, Thailand, August 2024{\natexlab{c}}. Association
  for Computational Linguistics.
\newblock \doi{10.18653/v1/2024.findings-acl.466}.
\newblock URL \url{https://aclanthology.org/2024.findings-acl.466/}.

\bibitem[Liu et~al.(2024{\natexlab{d}})Liu, Kamarthi, Kong, Zhao, Zhang, and
  Prakash]{pmlr-v235-liu24ae}
Haoxin Liu, Harshavardhan Kamarthi, Lingkai Kong, Zhiyuan Zhao, Chao Zhang, and
  B.~Aditya Prakash.
\newblock Time-series forecasting for out-of-distribution generalization using
  invariant learning.
\newblock In Ruslan Salakhutdinov, Zico Kolter, Katherine Heller, Adrian
  Weller, Nuria Oliver, Jonathan Scarlett, and Felix Berkenkamp, editors,
  \emph{Proceedings of the 41st International Conference on Machine Learning},
  volume 235 of \emph{Proceedings of Machine Learning Research}, pages
  31312--31325. PMLR, 21--27 Jul 2024{\natexlab{d}}.
\newblock URL \url{https://proceedings.mlr.press/v235/liu24ae.html}.

\bibitem[Liu et~al.(2025{\natexlab{a}})Liu, Zhao, Li, and
  Prakash]{liu2025evaluating}
Haoxin Liu, Zhiyuan Zhao, Shiduo Li, and B~Aditya Prakash.
\newblock Evaluating system 1 vs. 2 reasoning approaches for zero-shot
  time-series forecasting: A benchmark and insights.
\newblock \emph{arXiv preprint arXiv:2503.01895}, 2025{\natexlab{a}}.

\bibitem[Zhao et~al.(2023)Zhao, Rodriguez, and Prakash]{zhao2023performative}
Zhiyuan Zhao, Alexander Rodriguez, and B~Aditya Prakash.
\newblock Performative time-series forecasting.
\newblock \emph{arXiv preprint arXiv:2310.06077}, 2023.

\bibitem[Zhao et~al.(2025{\natexlab{a}})Zhao, Liu, and
  Prakash]{zhao2025tackling}
Zhiyuan Zhao, Haoxin Liu, and B~Aditya Prakash.
\newblock Tackling time-series forecasting generalization via mitigating
  concept drift.
\newblock \emph{arXiv preprint arXiv:2510.14814}, 2025{\natexlab{a}}.

\bibitem[Zhao et~al.(2025{\natexlab{b}})Zhao, Ni, Xu, Liu, Jin, and
  Prakash]{zhao2025timerecipe}
Zhiyuan Zhao, Juntong Ni, Shangqing Xu, Haoxin Liu, Wei Jin, and B~Aditya
  Prakash.
\newblock {TimeRecipe}: A time-series forecasting recipe via benchmarking
  module level effectiveness.
\newblock \emph{arXiv preprint arXiv:2506.06482}, 2025{\natexlab{b}}.

\bibitem[Boniol et~al.(2021{\natexlab{a}})Boniol, Paparrizos, Palpanas, and
  Franklin]{boniol2021sand}
Paul Boniol, John Paparrizos, Themis Palpanas, and Michael~J Franklin.
\newblock Sand: Streaming subsequence anomaly detection.
\newblock \emph{Proc. VLDB Endow.}, 14\penalty0 (10):\penalty0 1717--1729,
  2021{\natexlab{a}}.

\bibitem[Boniol et~al.(2021{\natexlab{b}})Boniol, Paparrizos, Palpanas, and
  Franklin]{boniol2021sandinaction}
Paul Boniol, John Paparrizos, Themis Palpanas, and Michael~J Franklin.
\newblock Sand in action: Subsequence anomaly detection for streams.
\newblock \emph{Proc. VLDB Endow.}, 14\penalty0 (12):\penalty0 2867--2870,
  2021{\natexlab{b}}.

\bibitem[Paparrizos et~al.(2020)Paparrizos, Liu, Elmore, and
  Franklin]{paparrizos2020debunking}
John Paparrizos, Chunwei Liu, Aaron~J Elmore, and Michael~J Franklin.
\newblock Debunking four long-standing misconceptions of time-series distance
  measures.
\newblock In \emph{Proceedings of the 2020 ACM SIGMOD International Conference
  on Management of Data}, pages 1887--1905, 2020.

\bibitem[Paparrizos and Gravano(2017)]{paparrizos2017fast}
John Paparrizos and Luis Gravano.
\newblock Fast and accurate time-series clustering.
\newblock \emph{ACM Transactions on Database Systems (TODS)}, 42\penalty0
  (2):\penalty0 1--49, 2017.

\bibitem[Liu et~al.(2024{\natexlab{e}})Liu, Boniol, Palpanas, and
  Paparrizos]{liu2024time}
Qinghua Liu, Paul Boniol, Themis Palpanas, and John Paparrizos.
\newblock Time-series anomaly detection: Overview and new trends.
\newblock \emph{Proceedings of the VLDB Endowment (PVLDB)}, 17\penalty0
  (12):\penalty0 4229--4232, 2024{\natexlab{e}}.

\bibitem[Sylligardos et~al.(2023)Sylligardos, Boniol, Paparrizos, Trahanias,
  and Palpanas]{sylligardos2023choose}
Emmanouil Sylligardos, Paul Boniol, John Paparrizos, Panos~E Trahanias, and
  Themis Palpanas.
\newblock Choose wisely: An extensive evaluation of model selection for anomaly
  detection in time series.
\newblock \emph{Proc. VLDB Endow.}, 16\penalty0 (11):\penalty0 3418--3432,
  2023.

\bibitem[Sylligardos et~al.(2025)Sylligardos, Paparrizos, Palpanas, Senellart,
  and Boniol]{sylligardos2025msad}
Emmanouil Sylligardos, John Paparrizos, Themis Palpanas, Pierre Senellart, and
  Paul Boniol.
\newblock Msad: A deep dive into model selection for time series anomaly
  detection.
\newblock \emph{The VLDB Journal}, 34\penalty0 (6):\penalty0 1--25, 2025.

\bibitem[Boniol et~al.(2024{\natexlab{b}})Boniol, Sylligardos, Paparrizos,
  Trahanias, and Palpanas]{boniol2024adecimo}
Paul Boniol, Emmanouil Sylligardos, John Paparrizos, Panos Trahanias, and
  Themis Palpanas.
\newblock Adecimo: Model selection for time series anomaly detection.
\newblock In \emph{2024 IEEE 40th International Conference on Data Engineering
  (ICDE)}, pages 5441--5444. IEEE, 2024{\natexlab{b}}.

\bibitem[Boniol et~al.(2025)Boniol, Krishna, Bruel, Liu, Huang, Palpanas, Tsay,
  Elmore, Franklin, and Paparrizos]{boniol2025vus}
Paul Boniol, Ashwin~K Krishna, Marine Bruel, Qinghua Liu, Mingyi Huang, Themis
  Palpanas, Ruey~S Tsay, Aaron Elmore, Michael~J Franklin, and John Paparrizos.
\newblock Vus: effective and efficient accuracy measures for time-series
  anomaly detection.
\newblock \emph{The VLDB Journal}, 34\penalty0 (3):\penalty0 32, 2025.

\bibitem[Paparrizos et~al.(2025)Paparrizos, Boniol, Liu, and
  Palpanas]{paparrizos2025advances}
John Paparrizos, Paul Boniol, Qinghua Liu, and Themis Palpanas.
\newblock Advances in time-series anomaly detection: Algorithms, benchmarks,
  and evaluation measures.
\newblock In \emph{Proceedings of the 31st ACM SIGKDD Conference on Knowledge
  Discovery and Data Mining V. 2}, pages 6151--6161, 2025.

\bibitem[Paparrizos and Bogireddy(2025)]{paparrizos2025time}
John Paparrizos and Sai Prasanna Teja~Reddy Bogireddy.
\newblock Time-series clustering: A comprehensive study of data mining, machine
  learning, and deep learning methods.
\newblock \emph{Proceedings of the VLDB Endowment}, 18\penalty0 (11):\penalty0
  4380--4395, 2025.

\bibitem[Paparrizos and Reddy(2023)]{paparrizos2023odyssey}
John Paparrizos and Sai Prasanna~Teja Reddy.
\newblock Odyssey: An engine enabling the time-series clustering journey.
\newblock \emph{Proceedings of the VLDB Endowment}, 16\penalty0 (12):\penalty0
  4066--4069, 2023.

\bibitem[Boniol et~al.(2022)Boniol, Paparrizos, Kang, Palpanas, Tsay, Elmore,
  and Franklin]{boniol2022theseus}
Paul Boniol, John Paparrizos, Yuhao Kang, Themis Palpanas, Ruey~S Tsay, Aaron~J
  Elmore, and Michael~J Franklin.
\newblock Theseus: Navigating the labyrinth of time-series anomaly detection.
\newblock \emph{Proc. VLDB Endow.}, 15\penalty0 (12):\penalty0 3702--3705,
  2022.

\bibitem[Boniol et~al.(2023)Boniol, Paparrizos, and Palpanas]{boniol2023new}
Paul Boniol, John Paparrizos, and Themis Palpanas.
\newblock New trends in time series anomaly detection.
\newblock In \emph{EDBT}, pages 847--850, 2023.

\bibitem[Bariya et~al.(2021)Bariya, von Meier, Paparrizos, and
  Franklin]{bariya2021k}
Mohini Bariya, Alexandra von Meier, John Paparrizos, and Michael~J Franklin.
\newblock k-shapestream: Probabilistic streaming clustering for electric grid
  events.
\newblock In \emph{2021 IEEE Madrid PowerTech}, pages 1--6. IEEE, 2021.

\bibitem[Liu et~al.(2025{\natexlab{b}})Liu, Lee, and Paparrizos]{liu2025easyad}
Qinghua Liu, Seunghak Lee, and John Paparrizos.
\newblock Easyad: A demonstration of automated solutions for time-series
  anomaly detection.
\newblock \emph{Proceedings of the VLDB Endowment}, 18\penalty0 (12):\penalty0
  5431--5434, 2025{\natexlab{b}}.

\bibitem[Liu et~al.(2025{\natexlab{c}})Liu, Heshmati, Mai, Abraham, Paparrizos,
  and Ren]{liu2025mllm4ts}
Qinghua Liu, Sam Heshmati, Zheda Mai, Zubin Abraham, John Paparrizos, and Liu
  Ren.
\newblock Mllm4ts: Leveraging vision and multimodal language models for general
  time-series analysis.
\newblock \emph{arXiv preprint arXiv:2510.07513}, 2025{\natexlab{c}}.

\bibitem[Liu et~al.(2025{\natexlab{d}})Liu, Lee, and Paparrizos]{liu2025tsb}
Qinghua Liu, Seunghak Lee, and John Paparrizos.
\newblock Tsb-autoad: Towards automated solutions for time-series anomaly
  detection.
\newblock \emph{Proceedings of the VLDB Endowment}, 18\penalty0 (11):\penalty0
  4364--4379, 2025{\natexlab{d}}.

\bibitem[Boniol et~al.(2024{\natexlab{c}})Boniol, Paparrizos, and
  Palpanas]{boniol2024interactive}
Paul Boniol, John Paparrizos, and Themis Palpanas.
\newblock An interactive dive into time-series anomaly detection.
\newblock In \emph{2024 IEEE 40th International Conference on Data Engineering
  (ICDE)}, pages 5382--5386. IEEE, 2024{\natexlab{c}}.

\bibitem[Giannoulidis et~al.(2025)Giannoulidis, Gounaris, and
  Paparrizos]{giannoulidis2025burst}
Apostolos Giannoulidis, Anastasios Gounaris, and John Paparrizos.
\newblock Burst: Rendering clustering techniques suitable for evolving streams.
\newblock \emph{Proceedings of the VLDB Endowment}, 18\penalty0 (11):\penalty0
  4054--4063, 2025.

\bibitem[Papadopoulos et~al.(2026)Papadopoulos, Giannoulidis, Gounaris, and
  Paparrizos]{papadopoulos2026power}
Anastasios Papadopoulos, Apostolos Giannoulidis, Anastasios Gounaris, and John
  Paparrizos.
\newblock The power of anomaly detection in predictive maintenance:[experiments
  \& analysis].
\newblock \emph{Proceedings of the ACM on Management of Data}, 4\penalty0 (3
  (SIGMOD):\penalty0 1--33, 2026.

\bibitem[Huang et~al.(2026{\natexlab{a}})Huang, Liu, Boniol, and
  Paparrizos]{huang2026hydra}
Mingyi Huang, Qinghua Liu, Paul Boniol, and John Paparrizos.
\newblock Hydra: A multi-level hierarchy-driven approach for robust anomaly
  detection in time series.
\newblock \emph{Proceedings of the ACM on Management of Data}, 4\penalty0 (3
  (SIGMOD):\penalty0 1--30, 2026{\natexlab{a}}.

\bibitem[Huang et~al.(2026{\natexlab{b}})Huang, Liu, Boniol, and
  Paparrizos]{huang2026glassboxad}
Mingyi Huang, Qinghua Liu, Paul Boniol, and John Paparrizos.
\newblock Glassboxad: An interactive system for dissecting hierarchical
  time-series anomaly detection.
\newblock In \emph{Companion of the International Conference on Management of
  Data}, pages 46--49, 2026{\natexlab{b}}.

\bibitem[Paparrizos et~al.(2026)Paparrizos, Boniol, Liu, and
  Palpanas]{paparrizos2026comprehensive}
John Paparrizos, Paul Boniol, Qinghua Liu, and Themis Palpanas.
\newblock A comprehensive guide to time-series anomaly detection.
\newblock In \emph{Proceedings of the Nineteenth ACM International Conference
  on Web Search and Data Mining}, pages 1363--1366, 2026.

\bibitem[Li and Paparrizos(2026)]{li2026mufasa}
Haojun Li and John Paparrizos.
\newblock Mufasa: Fast and accurate multivariate time-series clustering.
\newblock \emph{Proceedings of the ACM on Management of Data}, 4\penalty0 (3
  (SIGMOD):\penalty0 1--29, 2026.

\bibitem[Pastrana et~al.(2026)Pastrana, Li, Mehta, Vyas, Pakazad, Ohlsson, and
  Paparrizos]{pastrana2026read}
Gerardo Pastrana, Haojun Li, Dhruv Mehta, Anoushka Vyas, Sina~Khoshfetrat
  Pakazad, Henrik Ohlsson, and John Paparrizos.
\newblock Read-bench: Benchmarking historical instance retrieval for
  time-series diagnosis.
\newblock \emph{arXiv preprint arXiv:2609.32123}, 2026.

\bibitem[Yin et~al.(2008)Yin, Yang, and Pan]{Yin2008Sensor}
Jie Yin, Qiang Yang, and Jeffrey~Junfeng Pan.
\newblock Sensor-based abnormal human-activity detection.
\newblock \emph{IEEE Transactions on Knowledge and Data Engineering},
  20:\penalty0 1082--1090, 2008.

\bibitem[Chamroukhi et~al.(2013)Chamroukhi, Mohammed, Trabelsi, Oukhellou, and
  Amirat]{chamroukhiJointSegmentationMultivariate2013}
Faicel Chamroukhi, Samer Mohammed, Dorra Trabelsi, Latifa Oukhellou, and Yacine
  Amirat.
\newblock Joint segmentation of multivariate time series with hidden process
  regression for human activity recognition.
\newblock \emph{Neurocomputing}, 120:\penalty0 633--644, 2013.
\newblock ISSN 09252312.
\newblock \doi{10.1016/j.neucom.2013.04.003}.
\newblock URL \url{http://arxiv.org/abs/1312.6956}.

\bibitem[Ermshaus et~al.(2023{\natexlab{b}})Ermshaus, Sch{\"a}fer, Bagnall,
  Guyet, Ifrim, Lemaire, Leser, Leverger, and Malinowski]{Ermshaus2023Human}
Arik Ermshaus, Patrick Sch{\"a}fer, Anthony Bagnall, Thomas Guyet, Georgiana
  Ifrim, Vincent Lemaire, Ulf Leser, Colin Leverger, and Simon Malinowski.
\newblock Human activity segmentation challenge @ {ECML/PKDD}’23.
\newblock In \emph{8th Workshop on Advanced Analytics and Learning on Temporal
  Data}, 2023{\natexlab{b}}.

\bibitem[Tran et~al.(2022)Tran, Doan, Vu, and Vu]{Tran2022Machine}
Minh-Quang Tran, Hoang-Phuong Doan, Viet~Q. Vu, and Lien~T. Vu.
\newblock Machine learning and {IoT}-based approach for tool condition
  monitoring: A review and future prospects.
\newblock \emph{Measurement}, 2022.

\bibitem[Kemp et~al.(2000)Kemp, Zwinderman, Tuk, Kamphuisen, and
  Oberye]{Kemp2000Analysis}
Bob Kemp, Aeilko~H. Zwinderman, Bert Tuk, Hilbert A.~C. Kamphuisen, and
  Josefien J.~L. Oberye.
\newblock Analysis of a sleep-dependent neuronal feedback loop: the slow-wave
  microcontinuity of the {EEG}.
\newblock \emph{IEEE Transactions on Biomedical Engineering}, 47:\penalty0
  1185--1194, 2000.

\bibitem[Ermshaus et~al.(2024)Ermshaus, Sch{\"a}fer, and
  Leser]{Ermshaus2024ClaSS}
Arik Ermshaus, Patrick Sch{\"a}fer, and Ulf Leser.
\newblock Raising the {ClaSS} of streaming time series segmentation.
\newblock \emph{Proceedings of the VLDB Endowment}, 17\penalty0 (8):\penalty0
  1953--1966, 2024.

\bibitem[Truong et~al.(2020)Truong, Oudre, and Vayatis]{Truong2020Selective}
Charles Truong, Laurent Oudre, and Nicolas Vayatis.
\newblock Selective review of offline change point detection methods.
\newblock \emph{Signal Processing}, 167:\penalty0 107299, 2020.

\bibitem[Aminikhanghahi and Cook(2017)]{Aminikhanghahi2017Survey}
Samaneh Aminikhanghahi and Diane~J Cook.
\newblock A survey of methods for time series change point detection.
\newblock \emph{KAIS}, 51\penalty0 (2):\penalty0 339--367, 2017.

\bibitem[Wang et~al.(2023)Wang, Wu, Zhou, and Cai]{Wang2023Time2State}
Chengyu Wang, Kui Wu, Tongqing Zhou, and Zhiping Cai.
\newblock {Time2State}: An unsupervised framework for inferring the latent
  states in time series data.
\newblock \emph{Proceedings of the ACM on Management of Data}, 1:\penalty0 1 --
  18, 2023.

\bibitem[Lai et~al.(2024)Lai, Li, Zhang, Zhao, Qian, and Jensen]{Lai2024E2Usd}
Zhichen Lai, Huan Li, Dalin Zhang, Yan Zhao, Weizhu Qian, and Christian~S.
  Jensen.
\newblock {E2USD}: Efficient-yet-effective unsupervised state detection for
  multivariate time series.
\newblock \emph{Proceedings of the ACM Web Conference 2024}, 2024.

\bibitem[van~den Burg and Williams(2020)]{Burg2020Evaluation}
Gerrit~JJ van~den Burg and Christopher~KI Williams.
\newblock An evaluation of change point detection algorithms.
\newblock \emph{arXiv}, 2020.

\bibitem[Ermshaus et~al.(2025)Ermshaus, Sch{\"a}fer, and
  Leser]{Ermshaus2025CLaP}
Arik Ermshaus, Patrick Sch{\"a}fer, and Ulf Leser.
\newblock {CLaP} - state detection from time series.
\newblock \emph{Proceedings of the VLDB Endowment}, 19\penalty0 (1):\penalty0
  70 -- 83, 2025.

\bibitem[Gharghabi et~al.(2019)Gharghabi, Yeh, Ding, Ding, Hibbing, LaMunion,
  Kaplan, Crouter, and Keogh]{Gharghabi2018Domain}
Shaghayegh Gharghabi, Chin-Chia~Michael Yeh, Yifei Ding, Wei Ding, Paul~R.
  Hibbing, Samuel~R LaMunion, Andrew Kaplan, Scott~E. Crouter, and Eamonn~J.
  Keogh.
\newblock Domain agnostic online semantic segmentation for multi-dimensional
  time series.
\newblock \emph{Data Mining and Knowledge Discovery}, 33:\penalty0 96 -- 130,
  2019.

\bibitem[Katser and Kozitsin(2020)]{Katser2020SKAB}
Iurii~D. Katser and Vyacheslav~O. Kozitsin.
\newblock Skoltech anomaly benchmark ({SKAB}), 2020.

\bibitem[Chavelli et~al.(2026)Chavelli, Ermshaus, Yang, Sch{\"a}fer,
  Paparrizos, and Boniol]{tsseg2026}
F{\'e}lix Chavelli, Arik Ermshaus, Fan Yang, Patrick Sch{\"a}fer, John
  Paparrizos, and Paul Boniol.
\newblock {tsseg}: An interactive toolkit for time series segmentation.
\newblock In \emph{Machine Learning and Knowledge Discovery in Databases.
  Applied Data Science Track, Demo Track and Industrial Track}, Lecture Notes
  in Computer Science, pages 391--394. Springer Nature Switzerland, 2026.
\newblock \doi{10.1007/978-3-032-37685-5_41}.

\bibitem[Chavelli et~al.(2025)Chavelli, Boniol, and Thomazo]{chavelli2025}
F{\'e}lix Chavelli, Paul Boniol, and Micha{\"e}l Thomazo.
\newblock Toward interpretable evaluation measures for time series
  segmentation.
\newblock In \emph{The Thirty-ninth Annual Conference on Neural Information
  Processing Systems}, 2025.
\newblock URL \url{https://openreview.net/forum?id=Gz6dujD5j0}.

\bibitem[Wang et~al.(2024)Wang, Li, Zhou, and Cai]{Wang2024Unsupervised}
Chengyu Wang, Xionglve Li, Tongqing Zhou, and Zhiping Cai.
\newblock Unsupervised time series segmentation: A survey on recent advances.
\newblock \emph{Computers, Materials \& Continua}, 2024.

\bibitem[Basseville and Nikiforov(1993)]{bassevilleDetectionAbruptChanges}
Mich{\`e}le Basseville and Igor~V Nikiforov.
\newblock \emph{Detection of {{Abrupt Changes}}: {{Theory}} and
  {{Application}}}.
\newblock Prentice Hall, Englewood Cliffs, NJ, 1993.

\bibitem[Cao et~al.(2024)Cao, Zhu, Ting, Salim, Li, Yang, and
  Li]{Cao2022Detecting}
Yang Cao, Ye~Zhu, Kai~Ming Ting, Flora~D. Salim, Hong~Xian Li, Luxing Yang, and
  Gang Li.
\newblock Detecting change intervals with isolation distributional kernel.
\newblock \emph{J. Artif. Intell. Res.}, 79:\penalty0 273--306, 2024.
\newblock \doi{10.1613/jair.1.15762}.

\bibitem[Nguyen et~al.(2015)Nguyen, Woon, and Ng]{Nguyen2015Survey}
Hai-Long Nguyen, Yew-Kwong Woon, and Wee-Keong Ng.
\newblock {A Survey on Data Stream Clustering and Classification}.
\newblock \emph{{KAIS}}, 45\penalty0 (3):\penalty0 535--569, 2015.

\bibitem[Middlehurst et~al.(2024{\natexlab{b}})Middlehurst, Ismail-Fawaz,
  Guillaume, Holder, Guijo-Rubio, Bulatova, Tsaprounis, Mentel, Walter,
  Sch{{\"a}}fer, and Bagnall]{aeon}
Matthew Middlehurst, Ali Ismail-Fawaz, Antoine Guillaume, Christopher Holder,
  David Guijo-Rubio, Guzal Bulatova, Leonidas Tsaprounis, Lukasz Mentel, Martin
  Walter, Patrick Sch{{\"a}}fer, and Anthony Bagnall.
\newblock aeon: a {Python} toolkit for learning from time series.
\newblock \emph{Journal of Machine Learning Research}, 25\penalty0
  (289):\penalty0 1--10, 2024{\natexlab{b}}.
\newblock URL \url{http://jmlr.org/papers/v25/23-1444.html}.

\bibitem[L{\"{o}}ning et~al.(2019)L{\"{o}}ning, Bagnall, Ganesh, Kazakov,
  Lines, and Kir{\'{a}}ly]{sktime}
Markus L{\"{o}}ning, Anthony~J. Bagnall, Sajaysurya Ganesh, Viktor Kazakov,
  Jason Lines, and Franz~J. Kir{\'{a}}ly.
\newblock sktime: {A} unified interface for machine learning with time series.
\newblock \emph{CoRR}, abs/1909.07872, 2019.
\newblock URL \url{http://arxiv.org/abs/1909.07872}.

\bibitem[Killick and Eckley(2014)]{JSSv058i03}
Rebecca Killick and Idris~A. Eckley.
\newblock {changepoint}: An {R} package for changepoint analysis.
\newblock \emph{Journal of Statistical Software}, 58\penalty0 (3):\penalty0
  1--19, 2014.
\newblock \doi{10.18637/jss.v058.i03}.
\newblock URL
  \url{https://www.jstatsoft.org/index.php/jss/article/view/v058i03}.

\bibitem[Takeuchi and Yamanishi(2006)]{changefinder2006}
J.~Takeuchi and K.~Yamanishi.
\newblock A unifying framework for detecting outliers and change points from
  time series.
\newblock \emph{IEEE Transactions on Knowledge and Data Engineering},
  18\penalty0 (4):\penalty0 482--492, 2006.
\newblock \doi{10.1109/TKDE.2006.1599387}.

\bibitem[Deldari et~al.(2020)Deldari, Smith, Sadri, and
  Salim]{Deldari2020ESPRESSO}
Shohreh Deldari, Daniel~V. Smith, Amin Sadri, and Flora~D. Salim.
\newblock {ESPRESSO}: Entropy and shape aware time-series segmentation for
  processing heterogeneous sensor data.
\newblock \emph{Proc. ACM Interact. Mob. Wearable Ubiquitous Technol.},
  4:\penalty0 77:1--77:24, 2020.

\bibitem[Ryck et~al.(2021)Ryck, de~Vos, and Bertrand]{DeRyck2020Change}
Tim~De Ryck, Maarten de~Vos, and Alexander Bertrand.
\newblock Change point detection in time series data using autoencoders with a
  time-invariant representation.
\newblock \emph{IEEE Transactions on Signal Processing}, 69:\penalty0
  3513--3524, 2021.

\bibitem[Deldari et~al.(2021)Deldari, Smith, Xue, and Salim]{Deldari2020Time}
Shohreh Deldari, Daniel~V. Smith, Hao Xue, and Flora~Dilys Salim.
\newblock Time series change point detection with self-supervised contrastive
  predictive coding.
\newblock \emph{Proceedings of the Web Conference 2021}, 2021.

\bibitem[Imani and Shrivastava(2023)]{Imani2023tGLAD}
Shima Imani and Harsh Shrivastava.
\newblock {tGLAD}: A sparse graph recovery based approach for multivariate time
  series segmentation.
\newblock In \emph{AALTD@ECML/PKDD}, 2023.

\bibitem[Keogh et~al.(2001)Keogh, Chu, Hart, and Pazzani]{Keogh2001Online}
Eamonn Keogh, Selina Chu, David Hart, and Michael Pazzani.
\newblock An online algorithm for segmenting time series.
\newblock In \emph{Proceedings 2001 IEEE international conference on data
  mining}, pages 289--296. IEEE, 2001.

\bibitem[Matteson and James(2014)]{Matteson2013Nonparametric}
David~S. Matteson and Nicholas~A. James.
\newblock A nonparametric approach for multiple change point analysis of
  multivariate data.
\newblock \emph{Journal of the American Statistical Association}, 109:\penalty0
  334 -- 345, 2014.

\bibitem[Scott and Knott(1974)]{Scott1974Cluster}
A.~J. Scott and Martin Knott.
\newblock A cluster analysis method for grouping means in the analysis of
  variance.
\newblock \emph{Biometrics}, pages 507--512, 1974.

\bibitem[Hallac et~al.(2019)Hallac, Nystrup, and Boyd]{Hallac2016Greedy}
David Hallac, Peter Nystrup, and Stephen~P. Boyd.
\newblock Greedy {Gaussian} segmentation of multivariate time series.
\newblock \emph{Advances in Data Analysis and Classification}, 13:\penalty0 727
  -- 751, 2019.

\bibitem[Sadri et~al.(2017)Sadri, Ren, and Salim]{Sadri2017Information}
Amin Sadri, Yongli Ren, and Flora~D. Salim.
\newblock Information gain-based metric for recognizing transitions in human
  activities.
\newblock \emph{Pervasive and Mobile Computing}, 38:\penalty0 92--109, 2017.
\newblock ISSN 1574-1192.
\newblock \doi{10.1016/j.pmcj.2017.01.003}.
\newblock URL
  \url{https://www.sciencedirect.com/science/article/pii/S1574119217300081}.

\bibitem[Taylor and Letham(2018)]{Taylor2018Forecasting}
Sean~J Taylor and Benjamin Letham.
\newblock Forecasting at scale.
\newblock \emph{The American Statistician}, 72\penalty0 (1):\penalty0 37--45,
  2018.

\bibitem[Hinkley(1970)]{Hinkley1970Inference}
David~V. Hinkley.
\newblock Inference about the change-point in a sequence of random variables.
\newblock \emph{Biometrika}, 57\penalty0 (1):\penalty0 1--17, 1970.
\newblock ISSN 00063444, 14643510.
\newblock URL \url{http://www.jstor.org/stable/2334932}.

\bibitem[Killick et~al.(2012)Killick, Fearnhead, and
  Eckley]{Killick2012Optimal}
Rebecca Killick, Paul Fearnhead, and Idris~A. Eckley.
\newblock Optimal detection of changepoints with a linear computational cost.
\newblock \emph{Journal of the American Statistical Association}, 107:\penalty0
  1590 -- 1598, 2012.

\bibitem[Adams and MacKay(2007)]{Adams2007Bayesian}
Ryan~Prescott Adams and David~JC MacKay.
\newblock Bayesian online changepoint detection.
\newblock \emph{arXiv preprint arXiv:0710.3742}, 2007.

\bibitem[Matsubara et~al.(2014)Matsubara, Sakurai, and
  Faloutsos]{Matsubara2014AutoPlait}
Yasuko Matsubara, Yasushi Sakurai, and Christos Faloutsos.
\newblock {AutoPlait}: automatic mining of co-evolving time sequences.
\newblock \emph{Proceedings of the 2014 ACM SIGMOD International Conference on
  Management of Data}, 2014.

\bibitem[Johnson and Willsky(2013)]{Johnson2013Bayesian}
Matthew~J. Johnson and Alan~S. Willsky.
\newblock Bayesian nonparametric hidden semi-{Markov} models.
\newblock \emph{Journal of Machine Learning Research}, 14:\penalty0 673--701,
  2013.

\bibitem[Hallac et~al.(2017)Hallac, Vare, Boyd, and
  Leskovec]{Hallac2017Toeplitz}
David Hallac, Sagar Vare, Stephen~P. Boyd, and Jure Leskovec.
\newblock Toeplitz inverse covariance-based clustering of multivariate time
  series data.
\newblock \emph{Proceedings of the 23rd ACM SIGKDD International Conference on
  Knowledge Discovery and Data Mining}, 2017.

\bibitem[Allegra et~al.(2020)Allegra, Facco, Denti, Laio, and
  Mira]{Allegra2019Data}
Michele Allegra, Elena Facco, Francesco Denti, Alessandro Laio, and Antonietta
  Mira.
\newblock Data segmentation based on the local intrinsic dimension.
\newblock \emph{Scientific Reports}, 10:\penalty0 16449, 2020.
\newblock \doi{10.1038/s41598-020-72222-0}.

\bibitem[Reiss and Stricker(2012)]{Reiss2012Introducing}
Attila Reiss and Didier Stricker.
\newblock Introducing a new benchmarked dataset for activity monitoring.
\newblock \emph{2012 16th International Symposium on Wearable Computers}, pages
  108--109, 2012.

\bibitem[Bulling et~al.(2014)Bulling, Blanke, and
  Schiele]{Bulling2014ActRecTut}
Andreas Bulling, Ulf Blanke, and Bernt Schiele.
\newblock A tutorial on human activity recognition using body-worn inertial
  sensors.
\newblock \emph{ACM Computing Surveys}, 46\penalty0 (3):\penalty0 33:1--33:33,
  2014.
\newblock ISSN 0360-0300.
\newblock URL \url{https://github.com/andyknownasabu/ActRecTut}.

\bibitem[{Carnegie Mellon University}(2003)]{CMU2003MoCap}
{Carnegie Mellon University}.
\newblock The {CMU Graphics Lab Motion Capture Database}.
\newblock In \emph{CMU Graphics Lab}, 2003.
\newblock http://mocap.cs.cmu.edu.

\bibitem[Zhang and Sawchuk(2012)]{Zhang2012USCHAD}
Mi~Zhang and Alexander~A. Sawchuk.
\newblock {USC-HAD}: a daily activity dataset for ubiquitous activity
  recognition using wearable sensors.
\newblock \emph{Proceedings of the 2012 ACM Conference on Ubiquitous
  Computing}, 2012.

\bibitem[Gaugel and Reichert(2023)]{Gaugel2023PrecTime}
Stefan Gaugel and Manfred Reichert.
\newblock {PrecTime}: A deep learning architecture for precise time series
  segmentation in industrial manufacturing operations.
\newblock \emph{Engineering Applications of Artificial Intelligence},
  122:\penalty0 106078, June 2023.
\newblock ISSN 0952-1976.
\newblock \doi{10.1016/j.engappai.2023.106078}.
\newblock URL \url{http://dx.doi.org/10.1016/j.engappai.2023.106078}.

\bibitem[Salles et~al.(2024)Salles, Lima, Reis, Coutinho, Pacitti, Masseglia,
  Akbarinia, Chen, Garibaldi, Porto, and
  Ogasawara]{sallesSoftEDMetricsSoft2024a}
Rebecca Salles, Janio Lima, Michel Reis, Rafaelli Coutinho, Esther Pacitti,
  Florent Masseglia, Reza Akbarinia, Chao Chen, Jonathan Garibaldi, Fabio
  Porto, and Eduardo Ogasawara.
\newblock {{SoftED}}: {{Metrics}} for soft evaluation of time series event
  detection.
\newblock \emph{Computers \& Industrial Engineering}, 198:\penalty0 110728,
  2024.
\newblock ISSN 03608352.
\newblock \doi{10.1016/j.cie.2024.110728}.
\newblock URL
  \url{https://linkinghub.elsevier.com/retrieve/pii/S0360835224008507}.

\bibitem[Dau et~al.(2019)Dau, Bagnall, Kamgar, Yeh, Zhu, Gharghabi,
  Ratanamahatana, and Keogh]{dauUCRTimeSeries2019}
Hoang~Anh Dau, Anthony Bagnall, Kaveh Kamgar, Chin-Chia~Michael Yeh, Yan Zhu,
  Shaghayegh Gharghabi, Chotirat~Ann Ratanamahatana, and Eamonn Keogh.
\newblock The {{UCR Time Series Archive}}.
\newblock \emph{IEEE/CAA Journal of Automatica Sinica}, 6\penalty0
  (6):\penalty0 1293--1305, 2019.
\newblock ISSN 2329-9266.
\newblock \doi{10.1109/JAS.2019.1911747}.
\newblock URL
  \url{https://www.ieee-jas.net/en/article/doi/10.1109/JAS.2019.1911747}.

\bibitem[Dempster et~al.(2020)Dempster, Petitjean, and
  Webb]{Dempster2020ROCKET}
Angus Dempster, Fran{\c{c}}ois Petitjean, and Geoffrey~I. Webb.
\newblock {ROCKET}: exceptionally fast and accurate time series classification
  using random convolutional kernels.
\newblock \emph{Data Mining and Knowledge Discovery}, 34\penalty0 (5):\penalty0
  1454--1495, 2020.
\newblock \doi{10.1007/s10618-020-00701-z}.

\bibitem[Zhu et~al.(2017)Zhu, Imamura, Nikovski, and Keogh]{Zhu2017Matrix}
Yan Zhu, Makoto Imamura, Daniel Nikovski, and Eamonn Keogh.
\newblock {Matrix Profile VII}: Time series chains: A new primitive for time
  series data mining (best student paper award).
\newblock \emph{2017 IEEE International Conference on Data Mining (ICDM)},
  pages 695--704, 2017.

\bibitem[Eichinger and Kirch(2018)]{Eichinger2018MOSUM}
Birte Eichinger and Claudia Kirch.
\newblock A {MOSUM} procedure for the estimation of multiple random change
  points.
\newblock \emph{Bernoulli}, 24\penalty0 (1):\penalty0 526--564, 2018.
\newblock \doi{10.3150/16-BEJ887}.

\bibitem[Kawahara et~al.(2007)Kawahara, Yairi, and
  Machida]{kawaharaChangePointDetectionTimeSeries2007}
Yoshinobu Kawahara, Takehisa Yairi, and Kazuo Machida.
\newblock Change-{{Point Detection}} in {{Time-Series Data Based}} on
  {{Subspace Identification}}.
\newblock In \emph{Seventh {{IEEE International Conference}} on {{Data Mining}}
  ({{ICDM}} 2007)}, pages 559--564, October 2007.
\newblock \doi{10.1109/ICDM.2007.78}.

\bibitem[Yamada et~al.(2011)Yamada, Suzuki, Kanamori, Hachiya, and
  Sugiyama]{yamadaRelativeDensityRatioEstimation2011}
Makoto Yamada, Taiji Suzuki, Takafumi Kanamori, Hirotaka Hachiya, and Masashi
  Sugiyama.
\newblock Relative {{Density-Ratio Estimation}} for {{Robust Distribution
  Comparison}}, 2011.
\newblock URL \url{http://arxiv.org/abs/1106.4729}.

\bibitem[Liu et~al.(2013)Liu, Yamada, Collier, and
  Sugiyama]{liuChangePointDetectionTimeSeries2013}
Song Liu, Makoto Yamada, Nigel Collier, and Masashi Sugiyama.
\newblock Change-point detection in time-series data by relative density-ratio
  estimation.
\newblock \emph{Neural Networks}, 43:\penalty0 72--83, 2013.
\newblock \doi{10.1016/j.neunet.2013.01.012}.

\bibitem[Chang et~al.(2019)Chang, Li, Yang, and
  Póczos]{changKERNELCHANGEPOINTDETECTION2019}
Wei-Cheng Chang, Chun-Liang Li, Yiming Yang, and Barnabás Póczos.
\newblock Kernel change-point detection with auxiliary deep generative models,
  2019.
\newblock URL \url{https://arxiv.org/abs/1901.06077}.

\bibitem[Yamada et~al.(2013)Yamada, Kimura, Naya, and
  Sawada]{yamadaChangePointDetectionFeaturea}
Makoto Yamada, Akisato Kimura, Futoshi Naya, and Hiroshi Sawada.
\newblock Change-{{Point Detection}} with {{Feature Selection}} in
  {{High-Dimensional Time-Series Data}}.
\newblock In \emph{IJCAI}, 2013.

\bibitem[Fryzlewicz(2018)]{fryzlewiczTailgreedyBottomupData2018}
Piotr Fryzlewicz.
\newblock Tail-greedy bottom-up data decompositions and fast multiple
  change-point detection.
\newblock \emph{The Annals of Statistics}, 46\penalty0 (6B), December 2018.
\newblock ISSN 0090-5364.
\newblock \doi{10.1214/17-AOS1662}.

\bibitem[Wu et~al.(2024)Wu, Gundimeda, Mou, and
  Quinn]{wuUnsupervisedChangePoint}
Daoping Wu, Suhas Gundimeda, Shaoshuai Mou, and Christopher~J Quinn.
\newblock Unsupervised {{Change Point Detection}} in {{Multivariate Time
  Series}}.
\newblock In \emph{International Conference on Artificial Intelligence and
  Statistics}, 2024.

\bibitem[Fryzlewicz(2014)]{Fryzlewicz2014Wild}
Piotr Fryzlewicz.
\newblock Wild binary segmentation for multiple change-point detection.
\newblock \emph{The Annals of Statistics}, 42\penalty0 (6):\penalty0 2243,
  2014.

\bibitem[Kov{\'a}cs et~al.(2023)Kov{\'a}cs, B{\"u}hlmann, Li, and
  Munk]{kovacsSeededBinarySegmentation2023}
S~Kov{\'a}cs, P~B{\"u}hlmann, H~Li, and A~Munk.
\newblock Seeded binary segmentation: A general methodology for fast and
  optimal changepoint detection.
\newblock \emph{Biometrika}, 110\penalty0 (1):\penalty0 249--256, March 2023.
\newblock ISSN 1464-3510.
\newblock \doi{10.1093/biomet/asac052}.

\bibitem[Baranowski et~al.(2019)Baranowski, Chen, and
  Fryzlewicz]{baranowskiNarrowestOverThresholdDetectionMultiple2019}
Rafal Baranowski, Yining Chen, and Piotr Fryzlewicz.
\newblock Narrowest-{{Over-Threshold Detection}} of {{Multiple Change Points}}
  and {{Change-Point-Like Features}}.
\newblock \emph{Journal of the Royal Statistical Society Series B: Statistical
  Methodology}, 81\penalty0 (3):\penalty0 649--672, July 2019.
\newblock ISSN 1369-7412.
\newblock \doi{10.1111/rssb.12322}.

\bibitem[Wang and Samworth(2018)]{Wang2018Inspect}
Tengyao Wang and Richard~J. Samworth.
\newblock High dimensional change point estimation via sparse projection.
\newblock \emph{Journal of the Royal Statistical Society Series B: Statistical
  Methodology}, 80\penalty0 (1):\penalty0 57--83, 2018.
\newblock \doi{10.1111/rssb.12243}.

\bibitem[Londschien et~al.(2023)Londschien, B{\"u}hlmann, and
  Kov{\'a}cs]{londschienRandomForestsChange}
Malte Londschien, Peter B{\"u}hlmann, and Solt Kov{\'a}cs.
\newblock Random {{Forests}} for {{Change Point Detection}}.
\newblock \emph{Journal of Machine Learning Research}, 24\penalty0
  (216):\penalty0 1--45, 2023.

\bibitem[Chen and Zhang(2015)]{chenGraphbasedChangepointDetection2015}
Hao Chen and Nancy Zhang.
\newblock Graph-based change-point detection.
\newblock \emph{The Annals of Statistics}, 43\penalty0 (1), February 2015.
\newblock ISSN 0090-5364.
\newblock \doi{10.1214/14-AOS1269}.

\bibitem[Sejdinovic et~al.(2013)Sejdinovic, Sriperumbudur, Gretton, and
  Fukumizu]{Sejdinovic2013Equivalence}
Dino Sejdinovic, Bharath Sriperumbudur, Arthur Gretton, and Kenji Fukumizu.
\newblock Equivalence of distance-based and {RKHS}-based statistics in
  hypothesis testing.
\newblock \emph{The Annals of Statistics}, 41\penalty0 (5):\penalty0
  2263--2291, 2013.

\bibitem[Harchaoui and
  {L{\'e}vy-Leduc}(2010)]{harchaouiMultipleChangePointEstimation2010}
Z.~Harchaoui and C.~{L{\'e}vy-Leduc}.
\newblock Multiple {{Change-Point Estimation With}} a {{Total Variation
  Penalty}}.
\newblock \emph{Journal of the American Statistical Association}, 105\penalty0
  (492):\penalty0 1480--1493, December 2010.
\newblock ISSN 0162-1459, 1537-274X.
\newblock \doi{10.1198/jasa.2010.tm09181}.

\bibitem[Vert and Bleakley(2010)]{vertFastDetectionMultiple}
Jean-Philippe Vert and Kevin Bleakley.
\newblock Fast detection of multiple change-points shared by many signals using
  group {{LARS}}.
\newblock In \emph{Advances in Neural Information Processing Systems}, 2010.

\bibitem[Kim et~al.(2009)Kim, Koh, Boyd, and
  Gorinevsky]{kimEll1TrendFiltering2009}
Seung-Jean Kim, Kwangmoo Koh, Stephen Boyd, and Dimitry Gorinevsky.
\newblock {$\ell_1$} {{Trend Filtering}}.
\newblock \emph{SIAM Review}, 51\penalty0 (2):\penalty0 339--360, May 2009.
\newblock ISSN 0036-1445, 1095-7200.
\newblock \doi{10.1137/070690274}.

\bibitem[Maidstone et~al.(2017)Maidstone, Hocking, Rigaill, and
  Fearnhead]{maidstoneOptimalMultipleChangepoint2017}
Robert Maidstone, Toby Hocking, Guillem Rigaill, and Paul Fearnhead.
\newblock On optimal multiple changepoint algorithms for large data.
\newblock \emph{Statistics and Computing}, 27\penalty0 (2):\penalty0 519--533,
  March 2017.
\newblock ISSN 1573-1375.
\newblock \doi{10.1007/s11222-016-9636-3}.

\bibitem[Frick et~al.(2014)Frick, Munk, and
  Sieling]{frickMultiscaleChangePoint2014}
Klaus Frick, Axel Munk, and Hannes Sieling.
\newblock Multiscale {{Change Point Inference}}.
\newblock \emph{Journal of the Royal Statistical Society Series B: Statistical
  Methodology}, 76\penalty0 (3):\penalty0 495--580, June 2014.
\newblock ISSN 1369-7412.
\newblock \doi{10.1111/rssb.12047}.

\bibitem[Draayer et~al.(2021)Draayer, Cao, and Hao]{Draayer2021Reevaluating}
Erick Draayer, Huiping Cao, and Yifan Hao.
\newblock Reevaluating the change point detection problem with segment-based
  {Bayesian} online detection.
\newblock \emph{Proceedings of the 30th ACM International Conference on
  Information \& Knowledge Management}, 2021.

\bibitem[Fearnhead(2006)]{Fearnhead2006Exact}
Paul Fearnhead.
\newblock Exact and efficient {Bayesian} inference for multiple changepoint
  problems.
\newblock \emph{Statistics and Computing}, 16\penalty0 (2):\penalty0 203--213,
  2006.
\newblock \doi{10.1007/s11222-006-8450-8}.

\bibitem[Zhao et~al.(2019)Zhao, Wulder, Hu, Bright, Wu, Qin, Li, Toman,
  Mallick, Zhang, and Brown]{Zhao2019BEAST}
Kaiguang Zhao, Michael~A. Wulder, Tongxi Hu, Ryan Bright, Qiusheng Wu, Haiming
  Qin, Yang Li, Elizabeth Toman, Bani Mallick, Xuesong Zhang, and Molly Brown.
\newblock Detecting change-point, trend, and seasonality in satellite time
  series data to track abrupt changes and nonlinear dynamics: A {Bayesian}
  ensemble algorithm.
\newblock \emph{Remote Sensing of Environment}, 232:\penalty0 111181, 2019.
\newblock ISSN 0034-4257.
\newblock \doi{10.1016/j.rse.2019.04.034}.
\newblock URL
  \url{https://www.sciencedirect.com/science/article/pii/S0034425719301853}.

\bibitem[Knoblauch and Damoulas(2018)]{knoblauchSpatiotemporalBayesianOnline}
Jeremias Knoblauch and Theodoros Damoulas.
\newblock Spatio-temporal {{Bayesian On}}-line {{Changepoint Detection}} with
  {{Model Selection}}.
\newblock In \emph{International Conference on Machine Learning}, 2018.

\bibitem[Saat{\c c}i et~al.(2010)Saat{\c c}i, Turner, and
  Rasmussen]{saatciGaussianProcessChange}
Yunus Saat{\c c}i, Ryan Turner, and Carl~Edward Rasmussen.
\newblock Gaussian {{Process Change Point Models}}.
\newblock In \emph{International Conference on Machine Learning}, 2010.

\bibitem[Barry and Hartigan(1993)]{barryBayesianAnalysisChange2026}
Daniel Barry and J~A Hartigan.
\newblock A {{Bayesian Analysis}} for {{Change Point Problems}}.
\newblock \emph{Journal of the American Statistical Association}, 1993.

\bibitem[Chib(1998)]{chibEstimationComparisonMultiple1998}
Siddhartha Chib.
\newblock Estimation and comparison of multiple change-point models.
\newblock \emph{Journal of Econometrics}, 86\penalty0 (2):\penalty0 221--241,
  October 1998.
\newblock ISSN 0304-4076.
\newblock \doi{10.1016/S0304-4076(97)00115-2}.

\bibitem[Ko et~al.(2015)Ko, Chong, and Ghosh]{koDirichletProcessHidden2015}
Stanley I.~M. Ko, Terence T.~L. Chong, and Pulak Ghosh.
\newblock Dirichlet {{Process Hidden Markov Multiple Change}}-point {{Model}}.
\newblock \emph{Bayesian Analysis}, 10\penalty0 (2), June 2015.
\newblock ISSN 1936-0975.
\newblock \doi{10.1214/14-BA910}.

\bibitem[Fearnhead and Liu(2007)]{Fearnhead2007Online}
Paul Fearnhead and Zhen Liu.
\newblock On-line inference for multiple changepoint problems.
\newblock \emph{Journal of the Royal Statistical Society Series B: Statistical
  Methodology}, 69\penalty0 (4):\penalty0 589--605, 2007.
\newblock \doi{10.1111/j.1467-9868.2007.00601.x}.

\bibitem[Johnson and Willsky(2010)]{Johnson2010THierarchical}
Matthew~J. Johnson and Alan~S. Willsky.
\newblock The hierarchical {Dirichlet} process hidden semi-{Markov} model.
\newblock In \emph{Conference on Uncertainty in Artificial Intelligence}, 2010.

\bibitem[Rabiner(1989)]{rabinerTutorialHiddenMarkov1989}
L.R. Rabiner.
\newblock A tutorial on hidden {{Markov}} models and selected applications in
  speech recognition.
\newblock \emph{Proceedings of the IEEE}, 77\penalty0 (2):\penalty0 257--286,
  February 1989.
\newblock ISSN 1558-2256.
\newblock \doi{10.1109/5.18626}.

\bibitem[Fox et~al.(2011)Fox, Sudderth, Jordan, and
  Willsky]{foxStickyHDPHMMApplication2011}
Emily~B. Fox, Erik~B. Sudderth, Michael~I. Jordan, and Alan~S. Willsky.
\newblock A sticky {{HDP-HMM}} with application to speaker diarization.
\newblock \emph{The Annals of Applied Statistics}, 5\penalty0 (2A):\penalty0
  1020--1056, June 2011.
\newblock ISSN 1932-6157, 1941-7330.
\newblock \doi{10.1214/10-AOAS395}.

\bibitem[Fox et~al.(2014)Fox, Hughes, Sudderth, and
  Jordan]{foxJointModelingMultiple2014}
Emily~B. Fox, Michael~C. Hughes, Erik~B. Sudderth, and Michael~I. Jordan.
\newblock Joint modeling of multiple time series via the beta process with
  application to motion capture segmentation.
\newblock \emph{The Annals of Applied Statistics}, 8\penalty0 (3), September
  2014.
\newblock ISSN 1932-6157.
\newblock \doi{10.1214/14-AOAS742}.

\bibitem[Linderman et~al.(2017)Linderman, Johnson, Miller, Adams, Blei, and
  Paninski]{lindermanBayesianLearningInference2017}
Scott Linderman, Matthew Johnson, Andrew Miller, Ryan Adams, David Blei, and
  Liam Paninski.
\newblock Bayesian {{Learning}} and {{Inference}} in {{Recurrent Switching
  Linear Dynamical Systems}}.
\newblock In \emph{Proceedings of the 20th {{International Conference}} on
  {{Artificial Intelligence}} and {{Statistics}}}, pages 914--922. PMLR, April
  2017.

\bibitem[Dong et~al.(2020)Dong, Seybold, Murphy, and Bui]{pmlr-v119-dong20e}
Zhe Dong, Bryan Seybold, Kevin Murphy, and Hung Bui.
\newblock Collapsed amortized variational inference for switching nonlinear
  dynamical systems.
\newblock In Hal~Daumé III and Aarti Singh, editors, \emph{Proceedings of the
  37th International Conference on Machine Learning}, volume 119 of
  \emph{Proceedings of Machine Learning Research}, pages 2638--2647. PMLR,
  13--18 Jul 2020.
\newblock URL \url{https://proceedings.mlr.press/v119/dong20e.html}.

\bibitem[Hido et~al.(2008)Hido, Id{\'e}, Kashima, Kubo, and
  Matsuzawa]{Hido2008Unsupervised}
Shohei Hido, Tsuyoshi Id{\'e}, Hisashi Kashima, Harunobu Kubo, and Hirofumi
  Matsuzawa.
\newblock Unsupervised change analysis using supervised learning.
\newblock In \emph{Pacific-Asia Conference on Knowledge Discovery and Data
  Mining}, 2008.

\bibitem[Bonifati et~al.(2022)Bonifati, Buono, Guerra, and
  Tiano]{bonifatiTime2FeatLearningInterpretable2022}
Angela Bonifati, Francesco~Del Buono, Francesco Guerra, and Donato Tiano.
\newblock {{Time2Feat}}: Learning interpretable representations for
  multivariate time series clustering.
\newblock \emph{Proceedings of the VLDB Endowment}, 16\penalty0 (2):\penalty0
  193--201, October 2022.
\newblock ISSN 2150-8097.
\newblock \doi{10.14778/3565816.3565822}.

\bibitem[Cheng et~al.(2020)Cheng, Aeron, Hughes, Hussey, and
  Miller]{chengOptimalTransportBased2020}
Kevin~C. Cheng, Shuchin Aeron, Michael~C. Hughes, Erika Hussey, and Eric~L.
  Miller.
\newblock Optimal transport based change point detection and time series
  segment clustering.
\newblock In \emph{{{ICASSP}} 2020 - 2020 {{IEEE International Conference}} on
  {{Acoustics}}, {{Speech}} and {{Signal Processing}} ({{ICASSP}})}, pages
  6034--6038. IEEE, May 2020.
\newblock \doi{10.1109/ICASSP40776.2020.9053344}.

\bibitem[Gentile et~al.(2026)Gentile, Huang, and
  Murphy]{gentileWassersteinBasedIdentificationMetastable2026}
David Gentile, Joshua Huang, and James~M. Murphy.
\newblock {{Wasserstein}}-based identification of metastable states in time
  series data via change point detection and segment clustering.
\newblock arXiv preprint arXiv:2603.01989, March 2026.

\bibitem[Horvath et~al.(2024)Horvath, Issa, and
  Muguruza]{horvathClusteringMarketRegimes2021}
Blanka Horvath, Zacharia Issa, and Aitor Muguruza.
\newblock Clustering {{Market Regimes}} using the {{Wasserstein Distance}}.
\newblock \emph{Journal of Computational Finance}, 28\penalty0 (1):\penalty0
  1--39, 2024.

\bibitem[Sinn et~al.(2013)Sinn, Keller, and
  Chen]{sinnSegmentationClassificationTime2013}
M.~Sinn, K.~Keller, and B.~Chen.
\newblock Segmentation and classification of time series using ordinal pattern
  distributions.
\newblock \emph{The European Physical Journal Special Topics}, 222\penalty0
  (2):\penalty0 587--598, June 2013.
\newblock ISSN 1951-6401.
\newblock \doi{10.1140/epjst/e2013-01861-8}.

\bibitem[Carpentier et~al.(2024)Carpentier, Feremans, Meert, and
  Verbeke]{Carpentier2024Pattern}
Louis Carpentier, Len Feremans, Wannes Meert, and Mathias Verbeke.
\newblock Pattern-based time series semantic segmentation with gradual state
  transitions.
\newblock In \emph{SDM}, 2024.

\bibitem[Iss()]{IssuesLzz19980125Awesometimeseriessegmentationpapers}
Lzz19980125/awesome-time-series-segmentation-papers.
\newblock GitHub repository.
\newblock URL
  \url{https://github.com/lzz19980125/awesome-time-series-segmentation-papers}.

\end{thebibliography}

\newpage
\appendix
\raggedbottom

\section{Reader's Guide} \label{app:guide}

The appendix is organised in four parts: Part~I documents the design and protocol of the benchmark, Part~II gives the additional results behind each research question, Part~III explains how to extend \texttt{TSB-SEG}, and Part~IV gathers the per-(algorithm, dataset) result tables. Table~\ref{tab:guide} maps each appendix to the part of the main text it supports.

\begin{table}[h]
    \centering
    \small
    \caption{Reader's guide to the appendix.}
    \label{tab:guide}
    \begin{tabularx}{\linewidth}{l X l}
    \toprule
    \textbf{App.} & \textbf{Content} & \textbf{Supports} \\
    \midrule
    \multicolumn{3}{l}{\emph{Part I - Benchmark design and protocol}} \\
    \ref{app:notation} & Notations & Section~\ref{sec:fandRW} \\
    \ref{app:datasets} & Datasets, provenance and uncertainty of the labels, licences & Section~\ref{sec:datasets} \\
    \ref{app:method_details} & Per-method descriptions & Section~\ref{sec:benchmark_methods} \\
    \ref{app:mts_adapt} & Re-implementations and univariate/multivariate adaptations & Section~\ref{sec:benchmark} \\
    \ref{app:metric_defs} & Formal definitions of the metrics & Section~\ref{sec:metrics} \\
    \ref{app:param_grids} & Parameter grids and penalties & Section~\ref{sec:exp_strategy} \\
    \ref{app:compute} & Compute environment, budgets, missing series & Section~\ref{sec:experiments} \\
    \ref{app:padding_strategy} & Extended-scope pool & Section~\ref{sec:benchmark_methods} \\
    \midrule
    \multicolumn{3}{l}{\emph{Part II - Additional results}} \\
    \ref{app:metric_corr} & Correlation between metrics & Section~\ref{sec:metrics} \\
    \ref{app:domain_spiders} & Per-domain and per-dataset spider plots & RQ1, RQ3 \\
    \ref{app:rq1_extra} & Per-dataset score distributions, per-series oracle, ranking without tuning & RQ1 \\
    \ref{app:rq2_by_mode} & CPD vs.\ SD in each supervision mode & RQ2 \\
    \ref{app:runtime_sd} & Runtime distributions & RQ3 \\
    \ref{app:hyperparams} & Hyperparameter influence & RQ3 \\
    \ref{app:feature_sensitivity} & Sensitivity to series features & RQ3 \\
    \ref{app:pamap2_case_study} & Long series (PAMAP2) & RQ3 \\
    \midrule
    \multicolumn{3}{l}{\emph{Part III - Extending the benchmark}} \\
    \ref{app:extending} & Adding algorithms and datasets & - \\
    \midrule
    \multicolumn{3}{l}{\emph{Part IV - Result tables}} \\
    \ref{app:main_pool_full_grid} & Per-(algorithm, dataset) scores of the main pool & RQ1 \\
    \ref{app:secondary_metrics} & The same tables under secondary metrics & RQ1 \\
    \bottomrule
    \end{tabularx}
\end{table}

\section{Notations} \label{app:notation}

We summarise the main symbols and notations used in this paper in Table~\ref{notation}.

\begin{table}[h]
    \centering
    \caption{List of symbols and notations used in this paper.}
    \label{notation}
    \begin{tabular}{c l}
    \toprule
    Symbol & Definition \\
    \midrule
    $T$ & Time series $T \in \mathbb{R}^{n \times d}$ \\
    $n$ & Time series length \\
    $d$ & Time series number of dimensions \\
    $T_{s,e}$ & Subsequence $(t_s,\dots,t_e)$ of $T$ \\
    $\ell$ & Subsequence / window length \\
    $\mathcal{C}$ & Set of change points \{$CP_i$\}\\
    $n_{CP}$ & Number of change points $|\mathcal{C}|$ \\
    $S$ & Segment sequence $\{[CP_i,CP_{i+1})\}_{0 \leq i \leq n_{CP}}$ ($CP_0$, $CP_{n_{CP}+1}$: ends of $T$) \\
    $\mathcal{S}$ & Set of discrete states $\{s_1, \dots, s_{n_S}\}$ \\
    $n_{S}$ & Number of unique states $|\mathcal{S}|$ \\
    $Q$ & State sequence $(s_{t_1}, \dots, s_{t_n}) \in \mathcal{S}^n$ \\
    $K$ & Number of events provided in Guided mode ($n_{CP}$ or $n_S$) \\
    \bottomrule
    \end{tabular}
\end{table}

\section{Datasets} \label{app:datasets}

\texttt{TSB-SEG} bundles 8 datasets used for the systematic evaluation, plus PAMAP2~\cite{Reiss2012Introducing} kept aside as a long-series scalability case study (see RQ3 and Appendix~\ref{app:pamap2_case_study}). The 8 main datasets cover three domain families: \emph{Human Activity Recognition} (ActRecTut~\cite{Bulling2014ActRecTut}, HAS~\cite{Ermshaus2023Human}, MoCap~\cite{CMU2003MoCap}, USC-HAD~\cite{Zhang2012USCHAD}), \emph{industrial monitoring} (SKAB~\cite{Katser2020SKAB}, anomaly-style transitions on industrial machine signals; PUMP~\cite{Gaugel2023PrecTime}, 120 hydraulic-pump End-of-Line cycles measured by 9 sensors with up to 43 operational states per cycle), and \emph{general-purpose collections} drawn from heterogeneous fields (TSSB~\cite{Ermshaus2023ClaSP}, derived from the UEA/UCR classification archives~\cite{dauUCRTimeSeries2019}; UTSA~\cite{Gharghabi2018Domain}, spanning sports, medicine and environmental monitoring). All datasets are annotated at the state level, enabling consistent CPD and SD evaluation.\\ \textbf{Provenance and uncertainty of the labels.} We use the state labels released by the original authors, without re-annotation: activity labels recorded with the protocol or annotated by hand for ActRecTut, HAS, MoCap and PAMAP2, operating states of the test program for PUMP, normal and anomalous phases for SKAB, the class of each concatenated segment for TSSB and USC-HAD (one recording per activity), and the regime labels of the original sources for UTSA. Boundaries of concatenated series are exact; hand-placed boundaries are uncertain by a few samples to seconds, and transitions between activities are gradual, so the state of the samples around a boundary is partly a matter of convention. Three choices limit the influence of this uncertainty on our conclusions: the margin of the $F_1$-score ($1\%$ of the series length), the Gaussian kernel of the Gaussian $F_1$, and the \emph{Delay} class of SMS, which charges small boundary shifts less than structural errors; and every ranking is reported under several metrics (Appendix~\ref{app:secondary_metrics}). Dataset licence terms are reported in Table~\ref{tab:license}.

\begin{table}[ht]
  \centering
  \caption{Licences of datasets used in our experiments.}
  \label{tab:license}
  \resizebox{\textwidth}{!}{%
    \begin{tabular}{ll}
        \toprule
        \textbf{Dataset} & \textbf{Licence / Usage Terms} \\
        \midrule
        PAMAP2~\cite{Reiss2012Introducing} & CC BY 4.0 \\
        USC-HAD~\cite{Zhang2012USCHAD} & Licence not explicit; presented as freely available in the original paper \\
        TSSB~\cite{Ermshaus2023ClaSP} & BSD 3-Clause License \\
        UTSA~\cite{Gharghabi2018Domain} & No explicit licence at the source; redistributed under BSD 3-Clause in the CLaP repository \\
        ActRecTut~\cite{Bulling2014ActRecTut} & GPL-3.0-or-later \\
        MoCap~\cite{CMU2003MoCap} & Free for research; commercial use allowed in products, but resale (even converted) is prohibited \\
        HAS~\cite{Ermshaus2023Human} & CC BY-NC-SA 4.0 \\
        SKAB~\cite{Katser2020SKAB} & GPL-3.0 \\
        PUMP~\cite{Gaugel2023PrecTime} & CC BY 4.0 \\
        \bottomrule
    \end{tabular}%
  }
\end{table}

\section{Per-method Descriptions} \label{app:method_details}

We describe here every method of Figure~\ref{fig:taxonomy}, column by column, that is, by search procedure. In each column we first describe the algorithms evaluated in \texttt{TSB-SEG}, then cite the representative methods we do not evaluate. The two axes extend the decomposition of CPD methods into a cost function and a search method~\cite{Truong2020Selective}; we leave out its third element, the constraint on the number of changes, which our Guided and Non-Guided modes cover (Section~\ref{sec:exp_strategy}). Our implementation choices are gathered in Appendix~\ref{app:mts_adapt}.

\paragraph{Homogeneity criterion (rows).} The cost of~\cite{Truong2020Selective} is a measure of the homogeneity of a segment; the rows classify the methods by what must stay the same within a segment or a state. A \emph{model of the observations} is compared through its fit or likelihood, or through fixed statistics (mean, median, covariance, regression or autoregressive coefficients, trend, subspace, Gaussian process, local dimension). A \emph{model-free distribution} comparison contrasts empirical distributions without a model (characteristic kernel, energy distance, Wasserstein distance, density ratio, classifier, graph). \emph{Subsequence shapes} are compared to their nearest neighbours (matrix-profile arcs, $k$-nearest-neighbour classifiers), by a convolutional-kernel classifier (ROCKET~\cite{Dempster2020ROCKET} in CLaP) or through symbolic patterns. \emph{Learned representations} embed windows (time steps in SNLDS) with a trained network. The penalised detectors (Window, BinSeg, BottomUp, PELT, DynP) take their segment cost as a parameter: the $\ell_1$, $\ell_2$ and cosine costs (the cosine cost is the $\ell_2$ cost on normalised vectors, i.e.\ the linear kernel) compare a model, whereas the RBF kernel, which is characteristic, maps the signal to a Hilbert space where any change of distribution becomes a mean shift; these detectors therefore span the first two rows.

\paragraph{CPD - Profile and peaks.} These methods score every time point, usually by comparing two adjacent windows or arcs over the whole series, and report the peaks of the score. The score is computed once over the whole series, whereas binary segmentation recomputes its gain in each part: ClaSP, whose classification score profile is such a gain, therefore belongs to binary segmentation. \textbf{Window}~\cite{Truong2020Selective} (popularised in the 1990s) slides two adjacent windows along the series and detects the peaks of their discrepancy. \textbf{ChangeFinder}~\cite{changefinder2006} reduces CPD to outlier detection through two stacked sequentially discounting autoregressive (SDAR) models. \textbf{iCID}~\cite{Cao2022Detecting} represents each window by its kernel mean embedding under the Isolation Distributional Kernel and scores one minus the normalised kernel similarity between adjacent windows, choosing the isolation parameter that minimises the approximate entropy of the scores; it detects change \emph{intervals}, which generalises CPD to changes spread over time. \textbf{FLUSS}~\cite{Gharghabi2018Domain} uses motif-arc proximity to build the arc curve, whose local minima reveal CPs of homogeneous regions; \textbf{ESPRESSO}~\cite{Deldari2020ESPRESSO} extends FLUSS's arc curve with time-series chains~\cite{Zhu2017Matrix} and an entropy-based segmentation. With the deep-learning shift, self-supervised representations became prominent: \textbf{TS-CP$^2$}~\cite{Deldari2020Time} learns embeddings via Contrastive Predictive Coding to distinguish time-adjacent from time-distant pairs, and \textbf{TIRE}~\cite{DeRyck2020Change} learns an autoencoder Time-Invariant Representation to detect changes in time and frequency domains. \textbf{tGLAD}~\cite{Imani2023tGLAD} recovers temporal conditional independence graphs across sliding windows via a multi-task uGLAD solver, and segments the sequence by tracking the first- and second-order trajectories of these graphs (precision matrices, hence the model row). Not evaluated: MOSUM~\cite{Eichinger2018MOSUM}, a moving-sum statistic; subspace ID~\cite{kawaharaChangePointDetectionTimeSeries2007}, a change score obtained by subspace identification of a state-space model; RuLSIF~\cite{yamadaRelativeDensityRatioEstimation2011,liuChangePointDetectionTimeSeries2013}, which estimates a relative density ratio between adjacent windows; and KL-CPD~\cite{changKERNELCHANGEPOINTDETECTION2019}, a kernel two-sample test with a learned deep kernel. Previous studies found aHSIC~\cite{yamadaChangePointDetectionFeaturea}, RuLSIF and KL-CPD to be dominated by methods included in our benchmark~\cite{Deldari2020Time,Cao2022Detecting}.

\paragraph{CPD - Bottom-up.} These methods start from a fine partition and merge the most similar adjacent segments. \textbf{BottomUp}~\cite{Keogh2001Online} starts with a fine grid and iteratively merges segments. \textbf{EAgglo}~\cite{Matteson2013Nonparametric} merges adjacent segments agglomeratively, preserving time order, and keeps the segmentation that maximises a goodness-of-fit statistic built on the energy distance between segments. Not evaluated: TGUH~\cite{fryzlewiczTailgreedyBottomupData2018}, which merges several pairs of neighbouring regions per pass through an unbalanced Haar decomposition, and MDL greedy~\cite{wuUnsupervisedChangePoint}, a greedy minimum-description-length method that fits autoregressive models on windows, groups similar windows into subsequences and greedily removes the subsequences whose removal shortens the description.

\paragraph{CPD - Top-down (binary segmentation).} These methods search one change point per interval, split where the gain is largest and recurse on the two parts; the intervals are whole segments or random or deterministic sub-intervals. \textbf{BinSeg}~\cite{Scott1974Cluster} extends AMOC to multiple CPs by applying the single-CP search recursively to the resulting sub-signals. \textbf{GGS}~\cite{Hallac2016Greedy} models each segment as a multivariate Gaussian with a regularised covariance and maximises the total log-likelihood greedily, adding one breakpoint at a time and readjusting the previous ones after each addition. \textbf{IGTS}~\cite{Sadri2017Information} segments multivariate series top-down, splitting at the points that maximise the Information Gain (entropy reduction between segments). \textbf{ClaSP}~\cite{Ermshaus2023ClaSP} casts CPD as a collection of self-supervised subsequence-classification problems, reports the split maximising cross-validation accuracy, and recurses on the two parts. Not evaluated: wild binary segmentation (WBS)~\cite{Fryzlewicz2014Wild}, seeded binary segmentation (Seeded BS)~\cite{kovacsSeededBinarySegmentation2023}, narrowest-over-threshold (NOT)~\cite{baranowskiNarrowestOverThresholdDetectionMultiple2019} and Inspect~\cite{Wang2018Inspect}, with its sparse projections, which search sub-intervals for changes in mean or slope; changeforest~\cite{londschienRandomForestsChange}, whose gain is a classifier log-likelihood ratio; gSeg~\cite{chenGraphbasedChangepointDetection2015}, a graph-based scan applied recursively for multiple changes; and E-Divisive~\cite{Matteson2013Nonparametric}, the divisive counterpart of EAgglo: with a fixed number of changes, it is binary segmentation with an energy-distance cost, i.e.\ a kernel cost~\cite{Sejdinovic2013Equivalence}, a family we evaluate with the RBF kernel; without it, its permutation test costs $O(d n^2 + R\, n_{CP}\, n^2)$, beyond our budget on the long series.

\paragraph{CPD - Convex relaxation.} These methods fit a piecewise model with an $\ell_1$ penalty on its jumps, so that most jumps are exactly zero. \textbf{Prophet}~\cite{Taylor2018Forecasting} models trend changes with a sparse prior on rate changes at potential grid locations. Not evaluated: TV Lasso~\cite{harchaouiMultipleChangePointEstimation2010}, the Lasso on the jumps of a piecewise-constant signal, i.e.\ a total-variation penalty; fast group LARS~\cite{vertFastDetectionMultiple}, its multivariate extension for change points shared by many signals; and $\ell_1$ trend filtering~\cite{kimEll1TrendFiltering2009}, whose changes of slope are change points.

\paragraph{CPD - Optimal (dynamic programming).} These methods compute the exact minimum of a sum of segment costs, penalised or with a fixed number of changes. The foundation lies in \textbf{AMOC}~\cite{Hinkley1970Inference} (At Most One Change point), which searches for a single breakpoint by minimising the sum of squared errors. \textbf{PELT}~\cite{Killick2012Optimal} prunes the exact dynamic-programming search down to a linear expected cost; \textbf{DynP}~\cite{Truong2020Selective} solves the optimal segmentation when the number of CPs is known. Both minimise a sum of segment costs; kernel change-point detection (KCPD), often listed as a method of its own, runs these same two solvers with a kernel cost. Not evaluated: FPOP and SNIP~\cite{maidstoneOptimalMultipleChangepoint2017}, which speed up the penalised and fixed-$K$ recursions with functional and inequality pruning, and SMUCE~\cite{frickMultiscaleChangePoint2014}, which minimises the number of changes subject to a multiscale test, also by dynamic programming.

\paragraph{CPD - Bayesian.} These methods compute or sample the posterior over segmentations. \textbf{BOCD}~\cite{Adams2007Bayesian} introduces a Bayesian recursive message-passing scheme to infer the most recent CP, later extended for gradual changes~\cite{Draayer2021Reevaluating}; its exact offline counterpart~\cite{Fearnhead2006Exact} returns the posterior probability of a change at every time step. Not evaluated: BEAST~\cite{Zhao2019BEAST}, which averages piecewise trend and seasonal models by Markov chain Monte Carlo, performed very poorly in early experiments and was therefore discarded; BOCPDMS~\cite{knoblauchSpatiotemporalBayesianOnline}, BOCD with model selection among vector autoregressions, and GP-BOCPD~\cite{saatciGaussianProcessChange}, BOCD with Gaussian-process regimes; the product partition model (PPM)~\cite{barryBayesianAnalysisChange2026}, sampled by Markov chain Monte Carlo as in the \texttt{bcp} package; and the change-point HMM~\cite{chibEstimationComparisonMultiple1998} and DPHMM~\cite{koDirichletProcessHidden2015}, left-to-right hidden Markov models whose state either stays or moves to the next regime, so that each state is one segment.

\paragraph{One dynamic programme.} Because the score of a segmentation is a sum of segment costs, or the log of a product of segment likelihoods, the optimal, Bayesian and state-sequence searches run the same recursion over the position of the last change point: taking its minimum gives the best segmentation (PELT, DynP), taking its sum gives the posterior (BOCD); a left-to-right hidden Markov model is a product partition model, and allowing a return to a previous state turns it into the state-sequence search, with Viterbi for the minimum and forward--backward for the sum. The headers of Figure~\ref{fig:taxonomy} draw this recursion: the nodes are candidate change points and the arcs segments, so that a segmentation is a path; the optimal search keeps the cheapest path, the Bayesian one weighs every path (arc width: posterior probability of the segment; dot size: posterior probability of a change), and the state-sequence search follows a path through the states over time. With a geometric prior on segment lengths (a change at each step with probability $p$), a segmentation with $m$ changes has prior probability $p^m (1-p)^{n-1-m} \propto e^{-\beta m}$, with $\beta = \log((1-p)/p)$; the negative logarithm of the maximum a posteriori recursion of~\cite{Fearnhead2007Online} is then the optimal-partitioning recursion that PELT accelerates, with the negative log marginal likelihood of a segment as cost and $\beta$ as penalty.

\paragraph{SD - State sequence (dynamic programming).} These methods give a state to each step, window or block, tied to the next by a transition probability or a switching cost, and infer the whole sequence jointly. \textbf{AutoPlait}~\cite{Matsubara2014AutoPlait} groups HMM states into a Multi-Level Chain Model and uses Minimum Description Length (MDL) to choose both the segmentation and the number of regimes. \textbf{HDP-HSMM}~\cite{Johnson2010THierarchical,Johnson2013Bayesian} extends Hidden Semi-Markov Models with a Hierarchical Dirichlet Process prior, allowing nonparametric estimation of the number of states, with Gaussian emissions and explicit (shifted Poisson) state durations. \textbf{TICC}~\cite{Hallac2017Toeplitz} clusters sliding-window subsequences, each cluster being a Gaussian Markov random field with a Toeplitz-structured sparse inverse covariance, alternating the estimation of these covariances with a temporally consistent assignment of the windows (switching penalty $\beta$), computed by a Viterbi-type dynamic programme. Not evaluated: the Gaussian HMM~\cite{rabinerTutorialHiddenMarkov1989}; the sticky HDP-HMM~\cite{foxStickyHDPHMMApplication2011}; BP-AR-HMM~\cite{foxJointModelingMultiple2014}, a beta-process autoregressive HMM that shares behaviours across several series; rSLDS~\cite{lindermanBayesianLearningInference2017}, recurrent switching linear dynamical systems; and SNLDS~\cite{pmlr-v119-dong20e}, switching nonlinear dynamical systems with neural transition and emission functions, the latter in the learned-representation row.

\paragraph{SD - Segment, then group.} These methods first segment the series with a CPD method, then group the segments into states. \textbf{CLaP}~\cite{Ermshaus2025CLaP} starts from the segments found by ClaSP and shifts focus from feature space to label space: it exploits the self-supervised classification gain~\cite{Hido2008Unsupervised} to merge confusable segment classes, thereby clustering segments directly in label space. Not evaluated: ClaSP2Feat, the baseline of~\cite{Ermshaus2025CLaP}, which clusters the ClaSP segments with the features of Time2Feat~\cite{bonifatiTime2FeatLearningInterpretable2022}; CLaP dominates it in~\cite{Ermshaus2025CLaP}, which is why we did not keep it. Also not evaluated: optimal transport~\cite{chengOptimalTransportBased2020} (unnamed by its authors), which places change points at the peaks of a Wasserstein two-sample statistic between adjacent windows, averaged over channels, then groups the segments by spectral clustering of their pairwise Wasserstein distances, the number of states being given; a recent preprint~\cite{gentileWassersteinBasedIdentificationMetastable2026} follows the same pipeline and infers the number of states.

\paragraph{SD - Window clustering.} These methods label each window or point by a clustering, independently of its neighbours in time; time order is ignored or only enters the features, e.g.\ through contrastive pairs of adjacent windows, or a smoothing of the labels. \textbf{Hidalgo}~\cite{Allegra2019Data} clusters multivariate points by their local intrinsic dimension; its prior links nearest neighbours in the observation space, not consecutive time steps. \textbf{Time2State}~\cite{Wang2023Time2State} trains a contrastive-learning encoder on sliding windows, with sub-windows of the same window as positive pairs, and clusters the embeddings with a Dirichlet-process Gaussian mixture (DPGMM), which infers the number of states; \textbf{E2USD}~\cite{Lai2024E2Usd} optimises this pipeline with FFT-based compression, a Decomposed Dual-view Embedding Module, and False-Negative-Cancellation contrastive learning. Not evaluated: WK-means~\cite{horvathClusteringMarketRegimes2021}, a Wasserstein $k$-means that clusters the empirical distributions of windows of a return series; ordinal patterns~\cite{sinnSegmentationClassificationTime2013}, which clusters windows by their distribution of ordinal patterns; and PaTSS~\cite{Carpentier2024Pattern}, which embeds each time step with the frequent sequential patterns mined from multi-resolution symbolic windows and showed weak performance in previous work~\cite{chavelli2025}.

\paragraph{Empty cells.} Some cells of Figure~\ref{fig:taxonomy} are empty in the published literature that we reviewed. Nearest-neighbour shape comparisons involve pairs of subsequences from the whole series and do not decompose into a sum of segment costs, a likelihood or a penalised parameter, which the optimal, Bayesian and convex searches need; such methods could be built (e.g.\ with a self-similarity segment cost), but we found none. Likewise, convex and Bayesian searches need parameters or a likelihood, so model-free distributions only enter them through nonparametric priors. We found no published unsupervised method combining learned representations with bottom-up, binary, convex, optimal or Bayesian searches, nor one grouping CPD segments by learned representations. For a broader and continuously updated overview of segmentation methods and benchmarks, including recent ones not evaluated here, we refer the reader to the community reading list of~\cite{IssuesLzz19980125Awesometimeseriessegmentationpapers}.

\section{Adaptation and Implementation Details} \label{app:mts_adapt}

Integrating $26$ heterogeneous methods under a unified API required several reimplementations and adaptations. As mentioned in Section~\ref{sec:benchmark}, $13$ of the $26$ methods are absent from public toolkits (\texttt{aeon}~\cite{aeon}, \texttt{sktime}~\cite{sktime}, \texttt{ruptures}~\cite{Truong2020Selective}, and the R package \texttt{changepoint}~\cite{JSSv058i03}) and were implemented or adapted in this work, including ports of methods originally available only in MATLAB (ESPRESSO, iCID), a wrapper around the original C code of AutoPlait, extensions of univariate methods to multivariate inputs (FLUSS, Prophet, ChangeFinder, BOCD) and of IGTS to univariate inputs. For these methods, the results we obtain are consistent with the relative performance reported in the original papers and in the literature, although their datasets and protocols differ from those of our benchmark. Hidalgo is inherently incompatible with univariate inputs and is restricted to its intended multivariate setting. ESPRESSO, although described as essentially parameter-free, and TICC require the number of segments or states in the available implementations; like DynP, they are therefore evaluated in Guided mode only. AutoPlait chooses the number of regimes by MDL and takes no such input: like PELT and AMOC, it scores the same in both modes.

\paragraph{Implementation choices.} The penalised detectors (Window, BinSeg, BottomUp, PELT, DynP) take their segment cost as a parameter; we run them with the $\ell_1$, $\ell_2$, cosine and RBF costs (Appendix~\ref{app:param_grids}), and report kernel change-point detection (KCPD) under PELT and DynP, which it runs with a kernel cost. For BOCD, we use the exact offline inference of the product-partition model~\cite{Fearnhead2006Exact}, which returns the posterior probability of a change at every time step. iCID detects change intervals; we evaluate the change points it returns. Without $K$, the detectors that rank candidate change points stop at a threshold, one of their grid axes: ChangeFinder keeps the peaks of its score above its mean plus $k$ standard deviations, at least $1\%$ of $n$ apart; FLUSS keeps the minima of its corrected arc curve below a threshold, as its online variant FLOSS does~\cite{Gharghabi2018Domain}; TIRE keeps the peaks of its dissimilarity above a prominence threshold; Prophet fits a rate change $\delta$ at $200$ evenly spaced candidates and keeps those whose $|\delta|$ exceeds a multiple of the median $|\delta|$. With $K$, each keeps its $K$ strongest candidates (ChangeFinder among the peaks above $k = 2$). For HDP-HSMM, we use the weak-limit Gibbs sampler of~\cite{Johnson2013Bayesian}, with the prior mean duration set to a third of the series length. HDP-HSMM and Prophet are fitted, on series longer than $2{,}000$ samples, on the means of blocks of $\lceil n/2000 \rceil$ consecutive samples, and their labels or candidate change points are expanded back, so that boundaries fall on block edges: on a set of compared series the loss was negligible, whereas sampling HDP-HSMM at full length is intractable (without truncation of the durations, one sweep costs $O(n^2)$ per state).

\subsection{Univariate and Multivariate Adaptations}

\begin{figure}[h]
    \centering
    \includegraphics[width=0.8\linewidth]{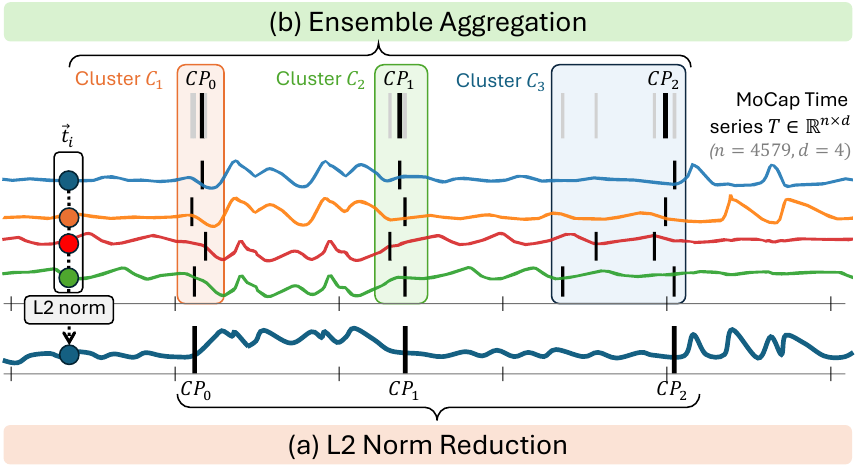}
    \caption{Multivariate adaptations for univariate methods (FLUSS, Prophet, ChangeFinder, BOCD): \emph{L2 Norm Reduction} collapses the multivariate signal into a univariate L2 norm series; \emph{Ensemble Aggregation} runs the detector independently per channel and merges the nearby change points of the channels (the $K$ best-supported ones when $K$ is given; shown with $K = 3$).}
    \label{fig:adaptMTS}
\end{figure}

\paragraph{FLUSS, Prophet, ChangeFinder and BOCD.} These univariate methods rely on two adaptations illustrated in Figure~\ref{fig:adaptMTS}: (i) \emph{L2 Norm Reduction} transforms the multivariate signal into a univariate L2-norm series, preserving joint-magnitude changes; (ii) \emph{Ensemble Aggregation} runs the detector independently per channel, pools all detected change points, clusters them by temporal proximity using a tolerance window, and returns the median timestamp of each cluster: with $K$, of the $K$ clusters supported by the most channels; without $K$, of every cluster, since a majority vote loses the changes that show on few channels. Prophet instead averages its candidate scores over the channels. Both adaptations are part of the grid for FLUSS, Prophet and ChangeFinder; BOCD uses the L2 reduction.

\paragraph{IGTS.} The original IGTS paper~\cite{Sadri2017Information} prescribes a normalise-and-complement preprocessing step (Section~4.4, Eq.~12-13) for \emph{all} inputs: each channel $c_i$ is normalised to sum to one and its complement $\max(c_i)-c_i$ is appended, doubling the channel count from $d$ to $2d$. This transformation ensures that Shannon entropy varies across segments even on univariate signals and on multivariate signals with positively-correlated channels, where it would otherwise be constant and mask change points. The \texttt{aeon}~\cite{aeon} implementation only applies the augmentation to univariate inputs; we apply it unconditionally, which restores the behaviour described in the paper on multivariate data and makes IGTS applicable to univariate series. We additionally implement the cumulative-sum optimisation (Eq.~5-6), which precomputes $F_i(t) = \sum_{j \leq t} c_{i_j}$ once and replaces each segment-sum recomputation by a single subtraction. This reduces each information-gain evaluation from $\mathcal{O}(d \cdot n)$ to $\mathcal{O}(k \cdot d)$, $k$ being the number of segments.

\paragraph{Hidalgo.} Hidalgo~\cite{Allegra2019Data} segments multivariate data into regions of differing local intrinsic dimension (ID) and does not natively support univariate inputs (the intrinsic dimension of a scalar series is $1$ everywhere). We attempted a univariate adaptation via Takens time-delay embedding,
\[
X_t = (x_t,\, x_{t+\tau},\, \dots,\, x_{t+(m-1)\tau}),
\]
so that different dynamical regimes (periodic, chaotic, noise) could occupy manifolds of different dimension in $\mathbb{R}^m$. We swept $m \in \{3, 5, 10\}$, $\tau \in \{1, 2\}$ and the number of Gibbs iterations $n_{\mathrm{iter}} \in \{300, 500, 1000\}$ on TSSB and UTSA, with $q = 3$ and $\zeta = 0.8$ as recommended by the original paper. The mean ARI over TSSB was $\approx 0.01$ (essentially random), and the only configuration that recovered a non-trivial ARI ($\approx 0.57$) was a synthetic sine-vs-white-noise signal with a genuine change in intrinsic dimension. Per-segment ID estimates obtained with the TWO-NN estimator confirm the diagnosis: on TSSB the median ID gap between consecutive ground-truth segments is $1.19$ (only $23\%$ of series with a gap $> 2$), and on UTSA it drops to $0.72$. Hidalgo's Gibbs sampler does not discriminate ID gaps of this magnitude, and the Potts smoothing term cannot create a discriminative signal where none exists. We therefore restrict Hidalgo to its intended multivariate use case.

\section{Evaluation Metrics: Formal Definitions} \label{app:metric_defs}

\subsection{Change Point Detection (CPD) Metrics}

\paragraph{$F_1$-score.} The standard margin-based evaluation metric for change point detection. A predicted change point is considered a true positive if it falls within a fixed tolerance window of a ground-truth change point. While widely used, its binary nature means it provides no partial credit for predictions that are just outside the margin, making it highly discontinuous.

\paragraph{Gaussian $F_1$.}
\label{def:gaussF}
To address the discontinuity of the standard $F_1$-score and its sensitivity to the arbitrary tolerance hyperparameter, we introduce a soft alternative. Let $\mathcal{C} = \{CP_i\}_{1 \leq i \leq n_{CP}}$ be the set of true CPs and $\hat{\mathcal{C}} = \{\hat{CP}_j\}$ the set of predicted CPs. We define a Gaussian reward for each $CP_i$:
\[
G_i(\hat{CP}) = \exp\!\left( -\frac{(\hat{CP} - CP_i)^2}{2\sigma^2} \right)
\]
with $\sigma$ a fraction of the time-series length $n$ (default $\sigma = 0.01\,n$). A greedy bipartite matching $\mathcal{M}$ between $\mathcal{C}$ and $\hat{\mathcal{C}}$ maximises the total reward:
\[
W = \sum_{(CP_i, \hat{CP}_j) \in \mathcal{M}} G_i(\hat{CP}_j)
\]
The soft precision and recall are $W / |\hat{\mathcal{C}}|$ and $W / |\mathcal{C}|$; the Gaussian $F_1$ is their harmonic mean. This provides smooth partial credit instead of binary accept/reject decisions.

\paragraph{Covering Score.} A classical (unidirectional) segment-based metric. Given source and target segment sets $S_{src}$ and $S_{tgt}$, the directional Covering score measures how well each source segment is covered by the target:
\[
\mathrm{Covering}(S_{src} \to S_{tgt}) = \frac{\sum_{s \in S_{src}} |s| \max_{t \in S_{tgt}} \operatorname{IoU}(s, t)}{\sum_{s \in S_{src}} |s|}
\]
By default, Covering is computed only as $\mathrm{Covering}(S_{GT} \to S_{pred})$. This uni-directional formulation is problematic because it is largely insensitive to over-segmentation: a prediction set containing many spuriously fragmented segments can still correctly cover the ground-truth segments.

\paragraph{Bidirectional Covering (BiC).}
\label{def:bicov}
To penalise methods that over-segment the true regions, we extend the classical Covering metric symmetrically. While classical Covering only assesses $\mathrm{Covering}(S_{GT} \to S_{pred})$, we additionally compute the prediction coverage $\mathrm{Covering}(S_{pred} \to S_{GT})$ to measure how well predicted regions map back to the truth. 

BiC is the harmonic mean of both directions. It matches the classical metric when predictions are perfectly aligned, but penalises over-segmentation: each fragment has a low IoU with the larger ground-truth segment it lies in, which drags the $\mathrm{Covering}(S_{pred} \to S_{GT})$ component down.

\subsection{State Detection (SD) Metrics}

\paragraph{Adjusted Rand Index (ARI).} A standard point-based clustering metric evaluating the similarity between ground-truth and predicted state assignments. It relies on the Rand Index (RI), which computes the fraction of agreeing pairs via a contingency matrix \(C = [n_{ij}]\), where $n_{ij}$ counts the timestamps assigned to true state $i$ and predicted state $j$. ARI adjusts this for chance:
\[
\mathrm{ARI} = \frac{\mathrm{RI} - \mathbb{E}[\mathrm{RI}]}{1 - \mathbb{E}[\mathrm{RI}]}
\]
While widely used, standard ARI (and similar metrics like AMI) treats each time step as an independent data point, systematically ignoring the inherent temporal ordering and structure of time series segmentation errors.

\paragraph{Weighted Adjusted Rand Index (WARI) \cite{chavelli2025}.} Introduced to address the structural blindness of point-based metrics, WARI is a temporal-aware extension of the ARI. It assigns a distance-to-boundary weight $w_i = 1 + \alpha\, d_i$ to each timestamp $i$, where $d_i$ is the distance to the nearest true change point and $\alpha$ controls the penalty slope (default $\alpha=0.1$). The contingency counts $n_{ij}$ are replaced by weighted sums $\widetilde{n}_{ij} = \sum_{k} w_k$. This gives more importance to errors that structurally disrupt a segment (e.g., splitting a state into two) and naturally downweights slight boundary misplacements.

\paragraph{State Matching Score (SMS) \cite{chavelli2025}.} SMS is an interpretable measure dedicated to SD. SMS first uses Hungarian matching to optimally align predicted and true discrete states. It then identifies error blocks and assigns a specific penalty to each based on its context (length, distance to boundaries, type of error). It decomposes errors into four structurally distinct types: \emph{Delay} (slight boundary misalignment), \emph{Transition} (brief transitional states between true regions), \emph{Isolation} (over-fragmentation of segments), and \emph{Missing} (completely undetected regions). The final SMS is calculated on a weighted sum of these penalties. We use the default weights of~\cite{chavelli2025}: $w_{\text{delay}}=0.1$, $w_{\text{transition}}=0.3$, $w_{\text{isolation}}=0.8$ and $w_{\text{missing}}=0.5$.

\begin{figure}[h]
    \centering
    \includegraphics[width=0.8\linewidth]{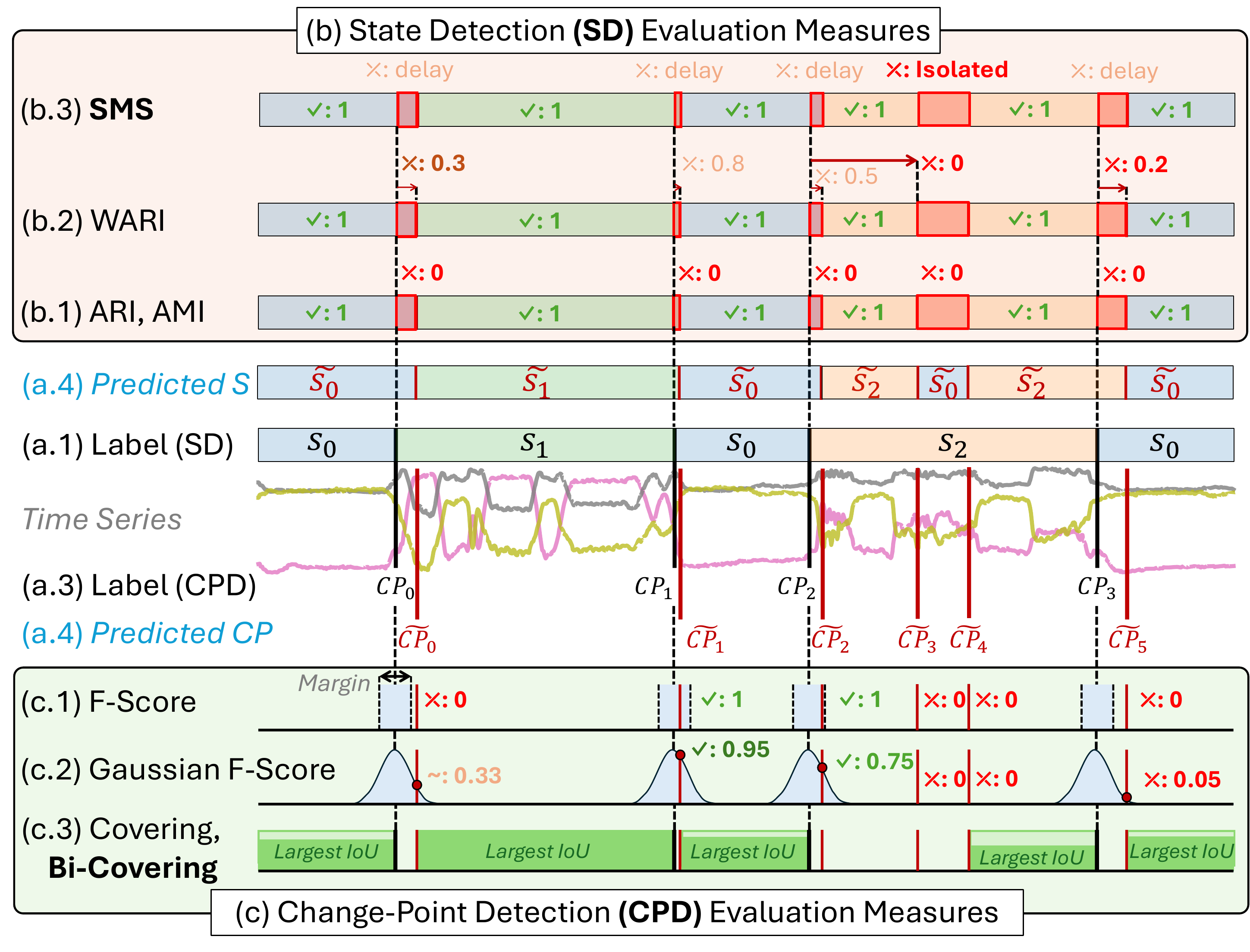}
    \caption{Evaluation measures in \textbf{TSB-SEG}.}
    \label{fig:measures}
\end{figure}

A synthetic illustration of the evaluation metrics used in \texttt{TSB-SEG} is provided in Figure~\ref{fig:measures}.

\section{Parameter Grids} \label{app:param_grids}

\paragraph{Parameter Grids Setup.} For each algorithm we identified the 2-4 hyperparameters that drive its behaviour, defined a small grid of 2-4 plausible values per parameter, drawn from the recommendations of the original papers and from values commonly used in the literature, and took the Cartesian product as the search space. Every grid was then calibrated on pilot series, in the same way for all algorithms, so that no method benefits from a more careful calibration than another. AutoPlait is essentially parameter-free and is evaluated in a single configuration, and the costliest extended-scope methods (BOCD, EAgglo, TS-CP$^2$) in their default configuration only. The penalised detectors (PELT, BinSeg, BottomUp, Window) take their penalty as a multiple $X$ of the BIC penalty $u \log n$, $u$ being the cost of one sample of the whole series ($d$ for the $\ell_2$ cost on standardised channels), so that one value transfers across lengths, dimensions and costs: $X \in \{1, 2, 8, 32\}$ ($\{1/16, 1/4, 1, 4\}$ for Window), with the $\ell_1$, $\ell_2$, RBF and cosine costs, a minimum segment length of $2$ and a candidate step of $5$; in Guided mode $n_{CP}$ replaces the penalty. As their penalty has no scale-free default in the original papers, their Default mode uses the $\ell_2$ cost with $X = 8$ ($X = 1/4$ for Window), chosen on a pilot of 27 benchmark series. The Default configuration $\theta_0$ need not be a grid configuration: when it is one (e.g.\ for the penalised detectors, Prophet, IGTS and HDP-HSMM), the Default score is that of its grid cell; otherwise $\theta_0$ is run on its own, and its score can then exceed the tuned one (in 5 of the 384 (algorithm, dataset, mode) cells of Appendix~\ref{app:main_pool_full_grid}, by at most $0.02$). Table~\ref{tab:param_grids} gives, for every method, the swept parameters, their value in Default mode and the values evaluated in each mode, as read from the logged runs; in Guided mode, the parameter that sets the number of change points, segments or states receives the ground-truth $K$ of each series and replaces the Non-Guided stopping rule (a penalty, a threshold or a maximum number of events). ChangeFinder, FLUSS, TIRE and Prophet have no scale-free rule to stop without $K$ in their original implementations (a hard-coded threshold, a fixed number of regimes, or none): their threshold (Appendix~\ref{app:mts_adapt}) was calibrated in the same way, on pilots of 23-24 benchmark series, and its grid of 3-4 values brackets the calibrated default. As the pilot series are drawn from the benchmark, the calibrated Default configurations and the grids are slightly optimistic, for every method alike, as are the tuned scores. Descriptions of every parameter are released with the configuration files of the \texttt{TSB-SEG} repository\footnote{\url{https://github.com/fchavelli/TSB-SEG}}; we made a best effort to keep the grids fair and representative of standard usage, while acknowledging that exhaustive coverage of every plausible combination is out of reach.

\begingroup
\setlength{\tabcolsep}{3pt}\renewcommand{\arraystretch}{1.05}\footnotesize
\begin{longtable}{@{}l c l l >{\raggedright\arraybackslash}p{3.2cm} >{\raggedright\arraybackslash}p{2.6cm}@{}}
\caption{\textbf{Hyperparameter grids, as evaluated.} For each method, the parameters swept in Parameter-search mode, their value in Default mode and the values tried in each mode; \#cfg is the number of configurations per dataset (Non-Guided/Guided). In Guided mode the parameter marked with $n_{CP}$, $n_{CP}{+}1$ (number of segments) or $n_S$ receives the ground-truth number of events of each series, and ``-'' marks a Non-Guided stopping rule that it replaces; ``same'': the Non-Guided values. The penalties of PELT, BinSeg, BottomUp and Window are multiples $X$ of the BIC penalty $u \log n$ (Appendix~\ref{app:param_grids}). Time2State and E2USD skip the window~64/step~100 combinations (step larger than the window). PELT, AMOC, iCID, EAgglo and tGLAD take no $K$: their Guided grid is their Non-Guided one (PELT and AMOC are not rerun). ClaSP distances: euclidean, z-normalised euclidean and complexity-invariant (CID) euclidean; window-size selection: SuSS, FFT or ACF; L2 norm and ensembling: the multivariate adaptations of Appendix~\ref{app:mts_adapt}. tGLAD's default runs used 100 epochs, 2000 on ActRecTut, MoCap and SKAB. \#cfg ``-'': default configuration only (BOCD, EAgglo, TS-CP$^2$: main parameters shown); \emph{n/a}: mode not evaluated.}\label{tab:param_grids} \\
\toprule
\textbf{Method} & \textbf{\#cfg} & \textbf{Parameter} & \textbf{Default} & \textbf{Non-Guided values} & \textbf{Guided values} \\
\midrule
\endfirsthead
\multicolumn{6}{l}{\emph{(continued from previous page)}} \\
\toprule
\textbf{Method} & \textbf{\#cfg} & \textbf{Parameter} & \textbf{Default} & \textbf{Non-Guided values} & \textbf{Guided values} \\
\midrule
\endhead
\midrule \multicolumn{6}{r}{\emph{(continued on next page)}} \\
\endfoot
\bottomrule
\endlastfoot
\multicolumn{6}{l}{\emph{CPD, distribution-based}} \\*
\midrule
AMOC & 3/3 & \texttt{min\_size} & 5 & \{5, 15, 30\} & same \\
\addlinespace[2pt]
BinSeg & 16/4 & \texttt{model} & $\ell_2$ & \{$\ell_1$, $\ell_2$, RBF, cosine\} & same \\*
 &  & \texttt{penalty} & 8 & \{1, 2, 8, 32\} & - \\*
 &  & \texttt{n\_cps} & - & - & $n_{CP}$ \\
\addlinespace[2pt]
BottomUp & 16/4 & \texttt{model} & $\ell_2$ & \{$\ell_1$, $\ell_2$, RBF, cosine\} & same \\*
 &  & \texttt{penalty} & 8 & \{1, 2, 8, 32\} & - \\*
 &  & \texttt{n\_cps} & - & - & $n_{CP}$ \\
\addlinespace[2pt]
Window & 48/12 & \texttt{model} & $\ell_2$ & \{$\ell_1$, $\ell_2$, RBF, cosine\} & same \\*
 &  & \texttt{pen} & 0.25 & \{0.0625, 0.25, 1, 4\} & - \\*
 &  & \texttt{width} & 100 & \{50, 100, 200\} & same \\*
 &  & \texttt{n\_cps} & - & - & $n_{CP}$ \\
\addlinespace[2pt]
ChangeFinder & 54/54 & \texttt{discount} & 0.005 & \{0.005, 0.01, 0.05\} & same \\*
 &  & \texttt{multivariate\_strategy} & L2 norm & \{L2 norm, ensembling\} & same \\*
 &  & \texttt{order} & 5 & 5 & \{3, 5, 10\} \\*
 &  & \texttt{smooth\_window} & 7 & \{5, 7, 15\} & same \\*
 &  & \texttt{threshold\_factor} & 4 & \{3, 4, 5\} & 2 \\*
 &  & \texttt{n\_cps} & - & - & $n_{CP}$ \\
\addlinespace[2pt]
PELT & 16/16 & \texttt{model} & $\ell_2$ & \{$\ell_1$, $\ell_2$, RBF, cosine\} & same \\*
 &  & \texttt{penalty} & 8 & \{1, 2, 8, 32\} & same \\
\addlinespace[2pt]
DynP & -/4 & \texttt{model} & $\ell_2$ & \multicolumn{1}{c}{\emph{n/a}} & \{$\ell_1$, $\ell_2$, RBF, cosine\} \\*
 &  & \texttt{n\_cps} & - & \multicolumn{1}{c}{\emph{n/a}} & $n_{CP}$ \\
\addlinespace[2pt]
Prophet & 16/4 & \texttt{changepoint\_prior\_scale} & 500 & \{50, 500\} & same \\*
 &  & \texttt{delta\_threshold} & 3 & \{1.5, 2.5, 3, 4\} & - \\*
 &  & \texttt{multivariate\_strategy} & L2 norm & \{L2 norm, ensembling\} & same \\*
 &  & \texttt{n\_changepoints} & - & - & $n_{CP}$ \\
\addlinespace[2pt]
iCID & 27/27 & \texttt{alpha} & 0.5 & \{0.1, 0.5, 0.9\} & same \\*
 &  & \texttt{t} & 200 & \{100, 200, 300\} & same \\*
 &  & \texttt{window\_size} & 50 & \{30, 50, 100\} & same \\
\addlinespace[2pt]
GGS & 9/3 & \texttt{k\_max} & 10 & \{5, 10, 20\} & $n_{CP}$ \\*
 &  & \texttt{lamb} & 1 & \{0.1, 1, 10\} & same \\
\midrule
\multicolumn{6}{l}{\emph{CPD, subsequence-based}} \\*
\midrule
IGTS & 9/3 & \texttt{k\_max} & 10 & \{5, 10, 25\} & $n_{CP}$ \\*
 &  & \texttt{step} & 5 & \{1, 5, 10\} & same \\
\addlinespace[2pt]
FLUSS & 54/18 & \texttt{exclusion\_factor} & 5 & \{3, 5, 10\} & same \\*
 &  & \texttt{multivariate\_strategy} & ensembling & \{L2 norm, ensembling\} & same \\*
 &  & \texttt{threshold} & 0.65 & \{0.5, 0.65, 0.8\} & - \\*
 &  & \texttt{window\_size} & 10 & \{10, 25, 50\} & same \\*
 &  & \texttt{n\_segments} & - & - & $n_{CP}{+}1$ \\
\addlinespace[2pt]
ESPRESSO & -/9 & \texttt{chain\_len} & 3 & \multicolumn{1}{c}{\emph{n/a}} & \{1, 3, 5\} \\*
 &  & \texttt{window\_size} & 64 & \multicolumn{1}{c}{\emph{n/a}} & \{32, 64, 128\} \\*
 &  & \texttt{n\_segments} & - & \multicolumn{1}{c}{\emph{n/a}} & $n_{CP}{+}1$ \\
\addlinespace[2pt]
TIRE & 108/27 & \texttt{domain} & both & \{FD, TD, both\} & same \\*
 &  & \texttt{max\_epochs} & 20 & \{20, 50, 100\} & same \\*
 &  & \texttt{prominence\_threshold} & 0.5 & \{0.05, 0.2, 0.5, 0.7\} & - \\*
 &  & \texttt{window\_size} & 20 & \{10, 20, 50\} & same \\*
 &  & \texttt{n\_segments} & - & - & $n_{CP}{+}1$ \\
\addlinespace[2pt]
ClaSP & 6/6 & \texttt{distance} & z-norm. & \{euclidean, z-norm., CID\} & same \\*
 &  & \texttt{window\_size} & SuSS & \{SuSS, FFT\} & same \\*
 &  & \texttt{n\_segments} & learn & learn & $n_{CP}{+}1$ \\
\midrule
\multicolumn{6}{l}{\emph{SD}} \\*
\midrule
AutoPlait & 1/1 & \texttt{n\_cps} & - & - & $n_{CP}$ \\
\addlinespace[2pt]
HDP-HSMM & 27/9 & \texttt{alpha} & 6 & \{1, 3, 6\} & same \\*
 &  & \texttt{dur\_alpha} & 10 & \{2, 10, 20\} & same \\*
 &  & \texttt{n\_max\_states} & 20 & \{10, 20, 40\} & $n_S$ \\
\addlinespace[2pt]
TICC & -/27 & \texttt{beta} & 3000 & \multicolumn{1}{c}{\emph{n/a}} & \{100, 1000, 5000\} \\*
 &  & \texttt{lambda\_parameter} & 0.001 & \multicolumn{1}{c}{\emph{n/a}} & \{0.0001, 0.001, 0.01\} \\*
 &  & \texttt{window\_size} & 3 & \multicolumn{1}{c}{\emph{n/a}} & \{3, 5, 10\} \\*
 &  & \texttt{n\_states} & - & \multicolumn{1}{c}{\emph{n/a}} & $n_S$ \\
\addlinespace[2pt]
Time2State & 24/24 & \texttt{depth} & 10 & \{1, 5, 10\} & same \\*
 &  & \texttt{step} & 50 & \{20, 50, 100\} & same \\*
 &  & \texttt{window\_size} & 256 & \{64, 128, 256\} & same \\*
 &  & \texttt{n\_states} & 20 & 20 & $n_S$ \\
\addlinespace[2pt]
E2USD & 24/24 & \texttt{depth} & 1 & \{1, 5, 10\} & same \\*
 &  & \texttt{step} & 50 & \{20, 50, 100\} & same \\*
 &  & \texttt{window\_size} & 256 & \{64, 128, 256\} & same \\*
 &  & \texttt{n\_states} & 20 & 20 & $n_S$ \\
\addlinespace[2pt]
CLaP & 6/6 & \texttt{classifier} & ROCKET & \{ROCKET, MR-Hydra\} & same \\*
 &  & \texttt{window\_size} & SuSS & \{SuSS, FFT, ACF\} & same \\*
 &  & \texttt{n\_change\_points} & - & - & $n_{CP}$ \\
\midrule
\multicolumn{6}{l}{\emph{Extended-scope pool}} \\*
\midrule
BOCD & -/- & \texttt{hazard\_lambda} & 300 & 300 & same \\*
 &  & \texttt{cp\_prob\_threshold} & 0.05 & 0.05 & same \\*
 &  & \texttt{min\_distance} & 25 & 25 & same \\*
 &  & \texttt{max\_cps} & - & - & $n_{CP}$ \\
\addlinespace[2pt]
EAgglo & -/- & \texttt{alpha} & 1 & 1 & same \\*
 &  & \texttt{penalty} & none & none & same \\
\addlinespace[2pt]
TS-CP$^2$ & -/- & \texttt{window\_size} & 128 & 128 & same \\*
 &  & \texttt{code\_size} & 32 & 32 & same \\*
 &  & \texttt{epochs} & 100 & 100 & same \\*
 &  & \texttt{temperature} & 0.1 & 0.1 & same \\*
 &  & \texttt{n\_cps} & - & - & $n_{CP}$ \\
\addlinespace[2pt]
tGLAD & 18/18 & \texttt{epochs} & 100/2000 & \{100, 500\} & same \\*
 &  & \texttt{threshold} & 0.5 & \{0.1, 0.3, 0.5\} & same \\*
 &  & \texttt{window\_size} & 32 & \{32, 64, 128\} & same \\
\addlinespace[2pt]
Hidalgo & -/6 & \texttt{q} & 3 & \multicolumn{1}{c}{\emph{n/a}} & \{2, 3, 5\} \\*
 &  & \texttt{zeta} & 0.8 & \multicolumn{1}{c}{\emph{n/a}} & \{0.5, 0.8\} \\*
 &  & \texttt{K\_states} & - & \multicolumn{1}{c}{\emph{n/a}} & $n_S$ \\
\end{longtable}
\endgroup

\section{Experimental Details and Compute Environment} \label{app:compute}

The evaluation behind the three research questions relies on the two modes of Section~\ref{sec:exp_strategy}: the Default configuration and a grid search over a fixed, predefined set of parameter values per method. We selected parameters as carefully as possible, but acknowledge that some may have been missed or under-tuned; we prioritised fairness and consistency while keeping the search space manageable. Each method from the main pool is evaluated across all series from the 9 datasets (8 main + PAMAP2) in these two modes, with the results from PAMAP2 used for scalability analysis only. Across the 8 main datasets, methods that complete most of a dataset but miss a few series (e.g. timeouts) are kept in the main figures and in every mean or rank over the $592$ series, with their missing $(\text{dataset}, \text{trial})$ entries imputed as a score of $0$ on every reported metric (tuned, median, worst-case and default); the per-(algorithm, dataset) tables of Appendices~\ref{app:main_pool_full_grid} and~\ref{app:secondary_metrics} instead average the completed series only. This is consistent with how a practitioner would summarise such a run: a method that fails to return on a series within the allotted budget effectively scores zero on that series. Padding is applied uniformly across modes (Non-Guided / Guided, CPD / SD) so that every algorithm is averaged over the same $592$ series. Two methods are padded, both in Guided mode: ESPRESSO ($20.8\%$ of the series, on PUMP and USC-HAD) and TICC ($13.5\%$, on PUMP only).
All experiments were executed on a high-performance computing cluster comprising Cascade Lake nodes with dual Intel Xeon 5218 CPUs (2$\times$16 cores at 2.4\,GHz, hyper-threading enabled) and 192\,GB RAM. Each run of one configuration on one dataset was allocated 8 CPU cores, 32\,GB RAM and at most 24 hours. Each configuration runs once per series, with a fixed seed for the methods whose configuration sets one (e.g.\ TIRE, Time2State, E2USD); we do not estimate the variance over seeds, which the averaging over the series of a dataset partly absorbs. All means and ranks weight every series equally, so HAS ($250$ of the $592$ series) weighs most; per-dataset results are in Appendices~\ref{app:domain_spiders} and~\ref{app:main_pool_full_grid}.

\section{Extended-Scope Pool} \label{app:padding_strategy}

The \textbf{extended-scope pool} gathers 5 methods that legitimately cannot be evaluated under the same budget as the main pool: \textbf{BOCD} and \textbf{EAgglo} have at least quadratic runtime and quadratic memory (both build $n \times n$ matrices that exceed memory beyond $n \gtrsim 20{,}000$, and our exact BOCD inference is cubic in time); \textbf{TS-CP$^2$} and \textbf{tGLAD} are linear-time at inference but their training exceeds the 24~h budget on parts of HAS, PUMP, USC-HAD and, for TS-CP$^2$, TSSB and UTSA; \textbf{Hidalgo} is multivariate-only by design.

Tables~\ref{tab:excluded_algos_cpd}-\ref{tab:excluded_algos_sd} summarise their per-dataset behaviour on the (algorithm, dataset, mode) cells they actually populate: completion, Default and Tuned scores and runtime (columns defined in the captions).

Because Hidalgo is multivariate-only by design but produces both a CPD and an SD output, it appears in both Table~\ref{tab:excluded_algos_cpd} (with BiC) and Table~\ref{tab:excluded_algos_sd} (with SMS).

\begin{table}[h]
    \centering
    \caption{\textbf{Per-dataset coverage and score of the five extended-scope algorithms with a CPD output.} \texttt{Compl.}\ = completed series / dataset cardinality. \emph{Default} columns report the mean Bidirectional Covering (BiC) under the author-recommended configuration; \emph{Tuned} columns report the mean BiC of the configuration maximising the dataset-level mean over the parameter sweep. $\tilde{t}$ is the median wall-clock per finished trial (in seconds). NG = Non-Guided, G = Guided. ``-'' = (mode, dataset) pair not evaluated.}
    \resizebox{\textwidth}{!}{\begin{tabular}{l l c c c c c c c c c c c c c}
\toprule
 &  &  & \multicolumn{3}{c}{\textbf{Default, NG}} & \multicolumn{3}{c}{\textbf{Default, G}} & \multicolumn{3}{c}{\textbf{Tuned, NG}} & \multicolumn{3}{c}{\textbf{Tuned, G}} \\
\cmidrule(lr){4-6} \cmidrule(lr){7-9} \cmidrule(lr){10-12} \cmidrule(lr){13-15}
\textbf{Algorithm} & \textbf{Dataset} & \textbf{Series} & Compl. & BiC & $\tilde{t}$ [s] & Compl. & BiC & $\tilde{t}$ [s] & Compl. & BiC & $\tilde{t}$ [s] & Compl. & BiC & $\tilde{t}$ [s] \\
\midrule
\multirow{8}{*}{tGLAD} & actrectut & 2 & 2/2 & 0.04 & 36706 & 2/2 & 0.04 & 33986 & 2/2 & 0.49 & 9885 & 2/2 & 0.46 & 9736 \\
 & has & 250 & 206/250 & 0.46 & 288 & 171/250 & 0.45 & 343 & 40/250 & 0.53 & 1462 & 43/250 & 0.53 & 1377 \\
 & mocap & 9 & 9/9 & 0.05 & 7304 & 9/9 & 0.06 & 5256 & 9/9 & 0.46 & 1841 & 9/9 & 0.49 & 1795 \\
 & pump & 120 & 57/120 & 0.07 & 1367 & 55/120 & 0.06 & 1575 & 13/120 & 0.52 & 6820 & 12/120 & 0.51 & 6800 \\
 & skab & 34 & 34/34 & 0.10 & 1326 & 34/34 & 0.10 & 1301 & 34/34 & 0.57 & 309 & 34/34 & 0.60 & 329 \\
 & tssb & 75 & 75/75 & 0.27 & 191 & 75/75 & 0.28 & 240 & 75/75 & 0.33 & 18.7 & 75/75 & 0.34 & 16.5 \\
 & usc-had & 70 & 44/70 & 0.13 & 1796 & 42/70 & 0.13 & 1891 & 9/70 & 0.48 & 10241 & 7/70 & 0.50 & 12580 \\
 & utsa & 32 & 32/32 & 0.16 & 585 & 32/32 & 0.16 & 548 & 32/32 & 0.12 & 64.6 & 32/32 & 0.13 & 71.0 \\
\midrule
\multirow{8}{*}{TS-CP$^2$} & actrectut & 2 & 2/2 & 0.47 & 10924 & 2/2 & 0.45 & 10931 & - & - & - & - & - & - \\
 & has & 250 & 32/250 & 0.46 & 1727 & 33/250 & 0.56 & 1630 & - & - & - & - & - & - \\
 & mocap & 9 & 9/9 & 0.47 & 3151 & 9/9 & 0.32 & 2516 & - & - & - & - & - & - \\
 & pump & 120 & 10/120 & 0.30 & 8133 & 10/120 & 0.17 & 8144 & - & - & - & - & - & - \\
 & skab & 34 & 34/34 & 0.41 & 237 & 34/34 & 0.55 & 245 & - & - & - & - & - & - \\
 & tssb & 75 & 50/75 & 0.46 & 1222 & 48/75 & 0.61 & 1166 & - & - & - & - & - & - \\
 & usc-had & 70 & 6/70 & 0.69 & 14199 & 6/70 & 0.29 & 13447 & - & - & - & - & - & - \\
 & utsa & 32 & 20/32 & 0.26 & 3728 & 20/32 & 0.68 & 3636 & - & - & - & - & - & - \\
\midrule
\multirow{8}{*}{EAgglo} & actrectut & 2 & 0/2 & - & - & 0/2 & - & - & - & - & - & - & - & - \\
 & has & 250 & 8/250 & 0.33 & 33.8 & 8/250 & 0.33 & 33.4 & - & - & - & - & - & - \\
 & mocap & 9 & 9/9 & 0.27 & 108 & 9/9 & 0.27 & 105 & - & - & - & - & - & - \\
 & pump & 120 & 9/120 & 0.42 & 995 & 9/120 & 0.42 & 1036 & - & - & - & - & - & - \\
 & skab & 34 & 34/34 & 0.00 & 6.9 & 34/34 & 0.00 & 3.7 & - & - & - & - & - & - \\
 & tssb & 75 & 75/75 & 0.02 & 22.3 & 75/75 & 0.02 & 26.7 & - & - & - & - & - & - \\
 & usc-had & 70 & 0/70 & - & - & 0/70 & - & - & - & - & - & - & - & - \\
 & utsa & 32 & 28/32 & 0.02 & 320 & 28/32 & 0.02 & 278 & - & - & - & - & - & - \\
\midrule
\multirow{8}{*}{BOCD} & actrectut & 2 & 0/2 & - & - & 0/2 & - & - & - & - & - & - & - & - \\
 & has & 250 & 8/250 & 0.45 & 718 & 8/250 & 0.50 & 738 & - & - & - & - & - & - \\
 & mocap & 9 & 9/9 & 0.16 & 3213 & 9/9 & 0.16 & 3310 & - & - & - & - & - & - \\
 & pump & 120 & 1/120 & 0.04 & 56912 & 1/120 & 0.04 & 65789 & - & - & - & - & - & - \\
 & skab & 34 & 34/34 & 0.64 & 63.5 & 34/34 & 0.64 & 67.0 & - & - & - & - & - & - \\
 & tssb & 75 & 28/75 & 0.43 & 391 & 28/75 & 0.43 & 397 & - & - & - & - & - & - \\
 & usc-had & 70 & 0/70 & - & - & 0/70 & - & - & - & - & - & - & - & - \\
 & utsa & 32 & 5/32 & 0.51 & 1443 & 5/32 & 0.51 & 1494 & - & - & - & - & - & - \\
\midrule
\multirow{6}{*}{Hidalgo} & actrectut & 2 & - & - & - & 2/2 & 0.01 & 10846 & - & - & - & 2/2 & 0.31 & 16506 \\
 & has & 250 & - & - & - & 8/250 & 0.14 & 757 & - & - & - & 8/250 & 0.14 & 731 \\
 & mocap & 9 & - & - & - & 9/9 & 0.05 & 1855 & - & - & - & 9/9 & 0.14 & 2746 \\
 & pump & 120 & - & - & - & 2/120 & 0.06 & 41896 & - & - & - & 2/120 & 0.11 & 32144 \\
 & skab & 34 & - & - & - & 34/34 & 0.08 & 151 & - & - & - & 34/34 & 0.22 & 134 \\
 & usc-had & 70 & - & - & - & 3/70 & 0.00 & 24346 & - & - & - & 4/70 & 0.37 & 18350 \\
\bottomrule
\end{tabular}
}
    \label{tab:excluded_algos_cpd}
\end{table}

\begin{table}[h]
    \centering
    \caption{\textbf{Per-dataset coverage and score of the SD-side extended-scope algorithm (Hidalgo).} Same column convention as Table~\ref{tab:excluded_algos_cpd}, with the SD primary metric (State Matching Score, SMS) replacing BiC. Hidalgo is multivariate-only by design, hence the absent (TSSB, UTSA) cells, and intrinsically Guided so the Non-Guided columns are uniformly empty.}
    \resizebox{\textwidth}{!}{\begin{tabular}{l l c c c c c c c c c c c c c}
\toprule
 &  &  & \multicolumn{3}{c}{\textbf{Default, NG}} & \multicolumn{3}{c}{\textbf{Default, G}} & \multicolumn{3}{c}{\textbf{Tuned, NG}} & \multicolumn{3}{c}{\textbf{Tuned, G}} \\
\cmidrule(lr){4-6} \cmidrule(lr){7-9} \cmidrule(lr){10-12} \cmidrule(lr){13-15}
\textbf{Algorithm} & \textbf{Dataset} & \textbf{Series} & Compl. & SMS & $\tilde{t}$ [s] & Compl. & SMS & $\tilde{t}$ [s] & Compl. & SMS & $\tilde{t}$ [s] & Compl. & SMS & $\tilde{t}$ [s] \\
\midrule
\multirow{6}{*}{Hidalgo} & actrectut & 2 & - & - & - & 2/2 & 0.25 & 10846 & - & - & - & 2/2 & 0.32 & 16506 \\
 & has & 250 & - & - & - & 8/250 & 0.39 & 757 & - & - & - & 8/250 & 0.39 & 776 \\
 & mocap & 9 & - & - & - & 9/9 & 0.25 & 1855 & - & - & - & 9/9 & 0.32 & 2746 \\
 & pump & 120 & - & - & - & 2/120 & 0.17 & 41896 & - & - & - & 1/120 & 0.17 & 72877 \\
 & skab & 34 & - & - & - & 34/34 & 0.59 & 151 & - & - & - & 34/34 & 0.59 & 159 \\
 & usc-had & 70 & - & - & - & 3/70 & 0.22 & 24346 & - & - & - & 3/70 & 0.22 & 23105 \\
\bottomrule
\end{tabular}
}
    \label{tab:excluded_algos_sd}
\end{table}

\paragraph{tGLAD - a budget-vs-quality tension.} In its default configuration, tGLAD scores poorly (BiC $0.04$-$0.46$) and takes from about $190$\,s (TSSB) to $37{,}000$\,s (ActRecTut) per series, leaving parts of HAS, PUMP and USC-HAD unfinished. These default runs use $100$ training epochs, except on ActRecTut, MoCap and SKAB, where only runs with $2000$ epochs are available (Table~\ref{tab:param_grids}), and this longer training still leaves BiC at $0.04$-$0.10$. Tuning the window size, the threshold and the number of epochs lifts BiC to $0.46$-$0.60$ on the six multivariate datasets, still well below the main-pool leaders on each of them, and leaves it at $0.12$-$0.34$ on the univariate TSSB and UTSA. The per-trial cost grows quasi-linearly with the number of epochs and roughly linearly with $n$, so the tuned Non-Guided configuration completes only $40/250$ series on HAS, $13/120$ on PUMP and $9/70$ on USC-HAD within our 24\,h budget per configuration and dataset; its tuned scores on these three datasets are therefore means over these completed series only. tGLAD thus combines a moderate accuracy with a tunable regime that a fixed compute budget cannot accommodate on long series, hence its place in the extended-scope pool.

\paragraph{TS-CP$^2$ - an unscalable method.} TS-CP$^2$ also requires a high number of epochs to converge. Its default configuration alone takes a median of $240$\,s (SKAB) to $14{,}200$\,s (USC-HAD) per series, so the grid sweep was not even dispatched: increasing the number of epochs would have exceeded the limits of our benchmark. Default scores remain limited (BiC $0.26$-$0.69$ depending on dataset), suggesting that TS-CP$^2$ is not competitive with the main-pool clique even on the cells it does complete, matching previous work conclusions \cite{Cao2022Detecting}.

\paragraph{EAgglo - $O(n^2)$ memory \emph{and} structurally weak detector.} Started, as in the reference implementation, from one cluster per observation, EAgglo's hierarchical agglomeration builds an explicit $n \times n$ distance matrix and exhausts the cluster's per-job RAM budget on series with $n \gtrsim 2 \cdot 10^4$ samples, hence the systematic 0/2 coverage on ActRecTut and 0/70 on USC-HAD (the original paper applies it to a large sample only after grouping the observations into initial clusters). On the cells it does complete, EAgglo is also \emph{intrinsically} weak because it massively over-segments: BiC collapses to $0.004$ on SKAB, $0.022$ on TSSB and $0.016$ on UTSA, where its F1 precision stays below $0.01$ at a recall above $0.9$, i.e.\ it returns a hundred or more change points per true one. Its number of change points is the one that maximises a goodness-of-fit statistic, the sum of the energy divergences between adjacent segments, to which every boundary adds a non-negative term and which carries no penalty on the number of segments by default; the method moreover assumes independent observations. It scores reasonably only on PUMP (BiC $0.42$) and HAS ($0.33$), and never approaches the main-pool leaders. We did not attempt grid tuning because no parameter setting can compensate for the algorithm's $O(n^2)$ scaling on the longer series.

\paragraph{BOCD: at least quadratic exact Bayesian inference in our implementation.} Our implementation runs the exact offline inference of the BOCD model~\cite{Fearnhead2006Exact}, which evaluates every candidate segment, then the posterior of every change-point position, and therefore costs $O(n^2)$ memory and $O(n^3)$ time; this makes it fail on USC-HAD ($n \approx 5 \cdot 10^4$), ActRecTut ($n \approx 3 \cdot 10^4$) and all but one PUMP series. On the smaller cells it completes, BOCD is not particularly competitive (BiC $0.64$ on SKAB, $0.43$ on TSSB, $0.51$ on UTSA, $0.45$ on HAS for the 8 series it processes), and cannot reach the main pool's coverage. We did not run a grid because the inference's hyperpriors are conservative and the per-series cost on PUMP already exceeds 16\,h/series at default.

\paragraph{Hidalgo (CPD).} Hidalgo's intrinsic-dimension prior is geometric and produces an SD-style label sequence; we evaluate the implied CPD by reading off the boundaries between consecutive runs of the same label. The default configuration is uniformly weak as a CPD detector (BiC between $0.003$ on USC-HAD and $0.14$ on HAS), reflecting that intrinsic dimension is a poor summary statistic for boundary localisation on series with few channels. The grid sweep recovers some performance on SKAB ($0.22$) and USC-HAD ($0.37$, on $4/70$ series only) but Hidalgo never approaches main-pool CPD detectors.

\paragraph{Hidalgo (SD).} As an SD method Hidalgo performs poorly on our series: its SMS reaches $0.59$ on SKAB (with full coverage) and $0.39$ on the 8 HAS series it completes, and stays in the $0.17$-$0.32$ range elsewhere, well below the SD main pool. This verdict holds at least for series with few channels ($d \leq 10$ in \texttt{TSB-SEG}). Hidalgo was designed for high-dimensional data (e.g.\ molecular dynamics trajectories or brain imaging~\cite{Allegra2019Data}), whose regions can differ widely in intrinsic dimension; its estimation of the local intrinsic dimension may well be more effective in high dimension, a setting our benchmark does not cover (Section~\ref{sec:conclusion}).

\paragraph{Take-away.} The extended-scope evaluation confirms the scoping logic of the main paper: none of these five methods Pareto-dominates the main-pool leaders on the cells they actually complete, and several (EAgglo, Hidalgo as a CPD detector) are outright weaker. The extended-scope pool therefore documents legitimate but specialised tools, not hidden state-of-the-art baselines.

\section{Pairwise Correlation of Evaluation Metrics} \label{app:metric_corr}

\begin{figure}[h]
    \centering
    \includegraphics[width=0.85\linewidth]{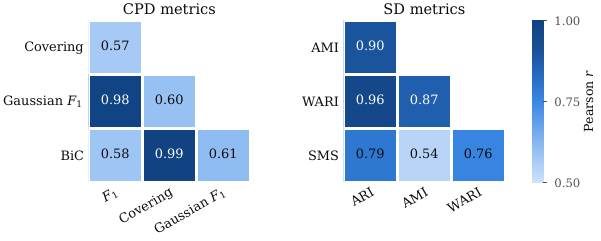}
    \caption{\textbf{Pairwise Pearson correlation between evaluation metrics} over all grid runs (\emph{Non-Guided} and \emph{Guided}) of every algorithm on the 8 main datasets ($438{,}176$ runs with all CPD metrics, $105{,}316$ with all SD metrics). Left: CPD metrics; right: SD metrics.}
    \label{fig:corr_metrics}
\end{figure}

Figure~\ref{fig:corr_metrics} reports the pairwise Pearson correlation between the CPD metrics (left) and the SD metrics (right), aggregated over \emph{all grid runs} (both \emph{Non-Guided} and \emph{Guided}) of every algorithm on the 8 main datasets of \texttt{TSB-SEG}. The CPD metrics are the standard $F_1$-score, the Covering score, our proposed Gaussian $F_1$ and BiC (Appendix~\ref{app:metric_defs}); the SD metrics are the Adjusted Rand Index (ARI), the Adjusted Mutual Information (AMI), the Weighted Adjusted Rand Index (WARI) and the State Matching Score (SMS)~\cite{chavelli2025}. All within-task correlations are positive: the lowest off-diagonal value is $0.57$ for CPD ($F_1$ vs.\ Covering) and $0.54$ for SD (AMI vs.\ SMS). Several metric pairs are essentially redundant: $F_1$ and Gaussian $F_1$ ($r = 0.98$) as well as Covering and BiC ($r = 0.99$) on the CPD side, and ARI/WARI ($r = 0.96$) on the SD side. SMS is the most distinct SD metric, which motivates its use alongside the clustering metrics ARI, AMI and WARI. High correlations do not make the rankings metric-invariant, but the leaders are the same under every metric (Appendix~\ref{app:secondary_metrics}), which supports reporting \textbf{BiC} (CPD) and \textbf{SMS} (SD) in the main text.

\section{Per-Domain and Per-Dataset Spider Plots} \label{app:domain_spiders}

\begin{figure}[H]
    \centering
    \begin{subfigure}[t]{0.49\linewidth}
        \includegraphics[width=\linewidth]{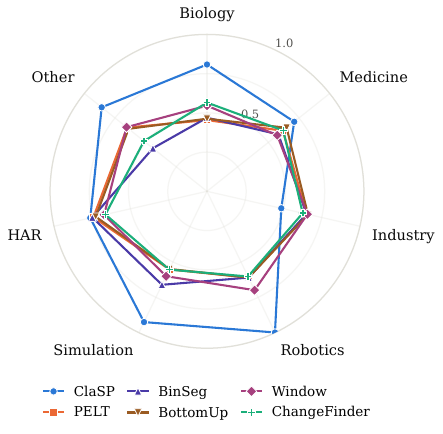}
        \caption{CPD Non-Guided.}
    \end{subfigure}\hfill
    \begin{subfigure}[t]{0.49\linewidth}
        \includegraphics[width=\linewidth]{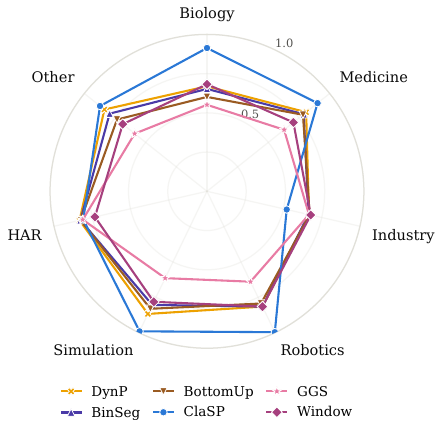}
        \caption{CPD Guided.}
    \end{subfigure}

    \vspace{0.4em}
    \begin{subfigure}[t]{0.49\linewidth}
        \includegraphics[width=\linewidth]{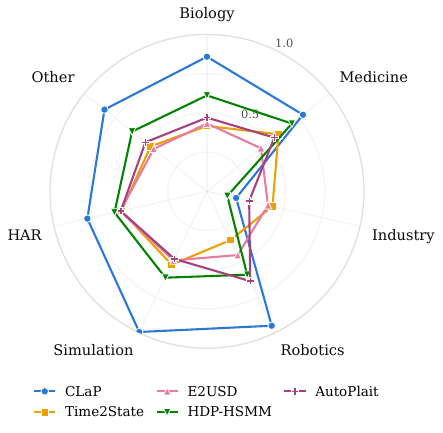}
        \caption{SD Non-Guided.}
    \end{subfigure}\hfill
    \begin{subfigure}[t]{0.49\linewidth}
        \includegraphics[width=\linewidth]{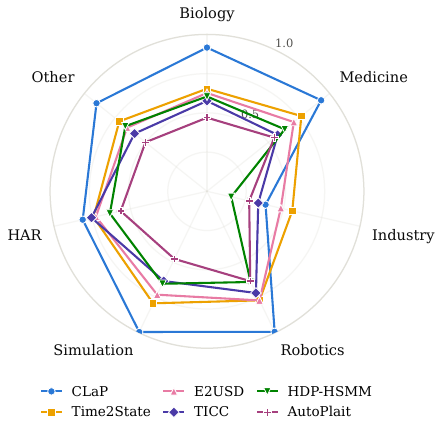}
        \caption{SD Guided.}
    \end{subfigure}
    \caption{Per-domain spider plots of the six best algorithms (tuned score) for CPD and SD, in both supervision modes. Within a task, a method keeps its colour and marker across the Non-Guided and Guided panels.}
    \label{fig:domain_spiders}
\end{figure}

\begin{figure}[H]
    \centering
    \begin{subfigure}[t]{0.49\linewidth}
        \includegraphics[width=\linewidth]{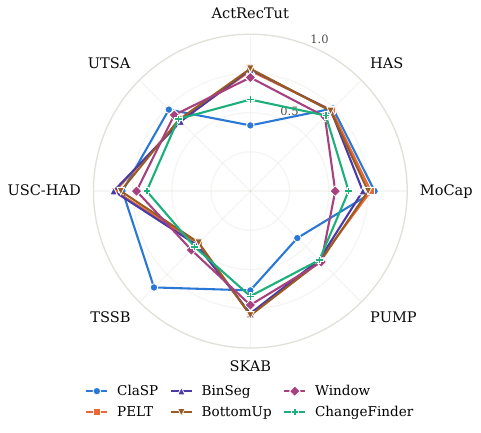}
        \caption{CPD Non-Guided.}
    \end{subfigure}\hfill
    \begin{subfigure}[t]{0.49\linewidth}
        \includegraphics[width=\linewidth]{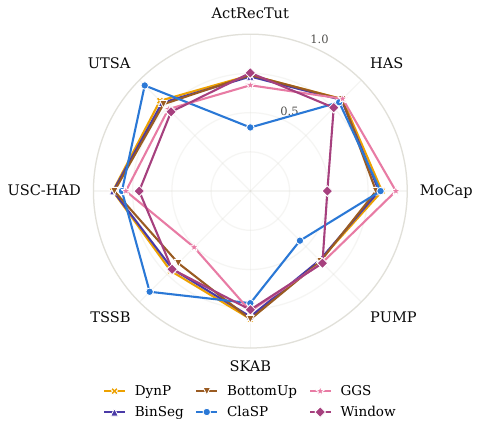}
        \caption{CPD Guided.}
    \end{subfigure}

    \vspace{0.4em}
    \begin{subfigure}[t]{0.49\linewidth}
        \includegraphics[width=\linewidth]{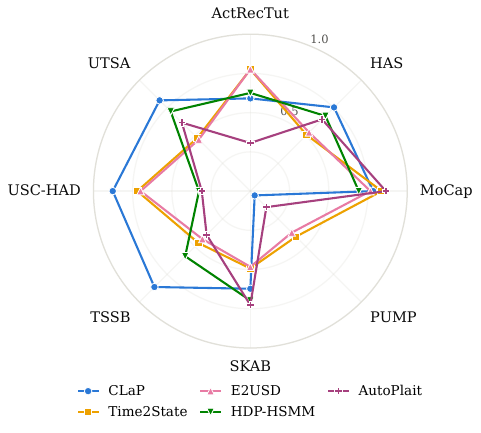}
        \caption{SD Non-Guided.}
    \end{subfigure}\hfill
    \begin{subfigure}[t]{0.49\linewidth}
        \includegraphics[width=\linewidth]{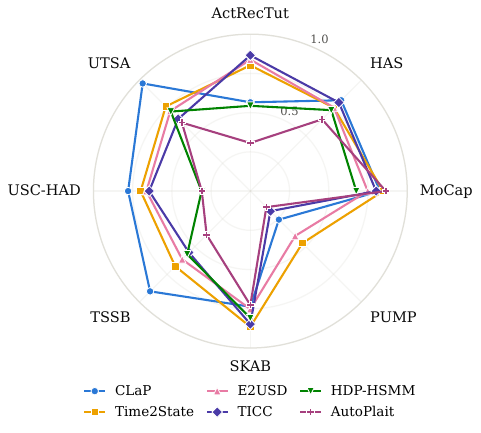}
        \caption{SD Guided.}
    \end{subfigure}
    \caption{Per-dataset spider plots of the six best algorithms (tuned score) for CPD and SD, in both supervision modes. Same colours and markers as Figure~\ref{fig:domain_spiders}.}
    \label{fig:dataset_spiders}
\end{figure}

For completeness, Figure~\ref{fig:domain_spiders} reports the per-domain spider plots of the six best algorithms (by mean tuned score) in each (task, supervision) combination (all the SD methods on the SD side), while Figure~\ref{fig:dataset_spiders} details their performance at the individual dataset level. They confirm the conclusions drawn in RQ1: no single method dominates across all configurations.

\section{Additional Views of the RQ1 Results} \label{app:rq1_extra}

The figures of RQ1 report means and ranks over the 592 series. This appendix shows how the scores spread within each dataset, how much a per-series choice of method could gain over the single best method, and how the methods rank without any tuning.

\begin{figure}[H]
    \centering
    \includegraphics[width=\linewidth]{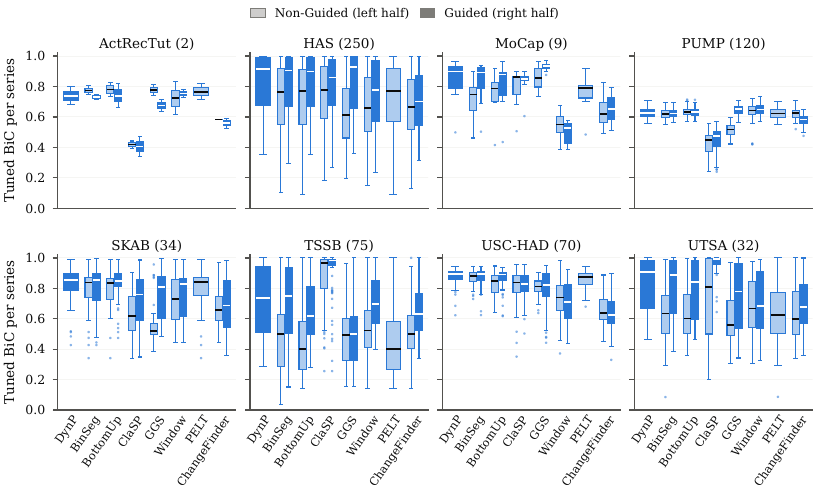}
    \caption{\textbf{Per-dataset distribution of the per-series tuned BiC} of the six best CPD methods of each mode, ordered by their best mean score over the two modes; number of series in parentheses. Light left half: \emph{Non-Guided}; dark right half: \emph{Guided}; a full-width box is drawn for a method run in one mode only (DynP) or that cannot use $n_{CP}$ (PELT). Boxes span the interquartile range, with the median as a line; whiskers extend to $1.5$ times the interquartile range, and dots mark the series beyond.}
    \label{fig:dataset_boxplots_cpd}
\end{figure}

\begin{figure}[H]
    \centering
    \includegraphics[width=\linewidth]{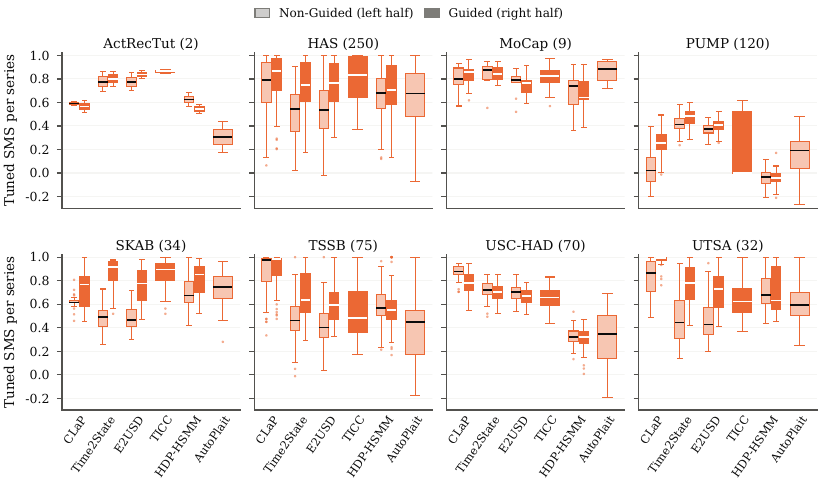}
    \caption{\textbf{Per-dataset distribution of the per-series tuned SMS} of the SD methods, with the conventions of Figure~\ref{fig:dataset_boxplots_cpd}; TICC runs in \emph{Guided} mode only, and AutoPlait cannot use $n_S$.}
    \label{fig:dataset_boxplots_sd}
\end{figure}

\paragraph{Per-dataset score distributions.} Figures~\ref{fig:dataset_boxplots_cpd} and~\ref{fig:dataset_boxplots_sd} give, for each dataset, the distribution of the per-series tuned score of the six best CPD methods of each mode (eight methods in all, as in the spider plots) and of every SD method. The interquartile range of a method is typically $0.2$-$0.35$ on HAS, TSSB and UTSA, against $0.05$-$0.15$ on PUMP and USC-HAD, whose series each follow one recording protocol. The feature-dependent behaviour of ClaSP (RQ1) is clearest on UTSA, where a quarter of its scores fall below $0.5$ and $41\%$ reach $0.9$ or more. Missing series count as $0$ (Appendix~\ref{app:compute}): they pull TICC's PUMP box down to $0$ in Figure~\ref{fig:dataset_boxplots_sd} ($80$ of the $120$ series).

\paragraph{Per-series oracle.} Table~\ref{tab:oracle} compares, in each task and mode, the single best method (highest mean tuned score) with an oracle that picks the best method of each series, among the three best methods or among all of them. Picking per series would raise the mean score by $0.11$ to $0.16$ in every task and mode, which quantifies the answer to RQ1: the single best method is strictly best on less than half of the series ($4$-$40\%$), and the methods that win the others are not interchangeable with it. With $n_{CP}$, the three best CPD methods (DynP, BinSeg and BottomUp) are nearly redundant (oracle $0.793$ against $0.778$), and the gain comes from the other methods. This oracle selects with the labels of each series and takes a maximum over noisy scores: it is an upper bound on what an automated method selection, e.g.\ meta-learned from the series features (Section~\ref{sec:conclusion}), could reach.

\begin{table}[H]
    \centering
    \small
    \caption{\textbf{Single best method against a per-series oracle} (mean tuned BiC for CPD, SMS for SD, over the 592 series). \emph{Single best}: highest mean tuned score; \emph{strict best on}: share of the series on which it beats every other method (ties are frequent among the penalised detectors given $n_{CP}$); \emph{oracle}: mean over the series of the best score among the three best methods or among all methods of the mode.}
    \label{tab:oracle}
    \resizebox{\linewidth}{!}{\begin{tabular}{l l c l c c c c}
\toprule
\textbf{Task} & \textbf{Mode} & \textbf{\#Methods} & \textbf{Single best} & \textbf{Score} & \textbf{Strict best on} & \textbf{Oracle, top 3} & \textbf{Oracle, all} \\
\midrule
CPD & Non-Guided & 17 & ClaSP & 0.696 & 18\% & 0.805 & 0.858 \\
CPD & Guided & 20 & DynP & 0.778 & 4\% & 0.793 & 0.884 \\
SD & Non-Guided & 5 & CLaP & 0.633 & 40\% & 0.732 & 0.772 \\
SD & Guided & 6 & CLaP & 0.714 & 37\% & 0.796 & 0.821 \\
\bottomrule
\end{tabular}
}
\end{table}

\begin{figure}[H]
    \centering
    \begin{subfigure}[t]{0.49\linewidth}
        \includegraphics[width=\linewidth]{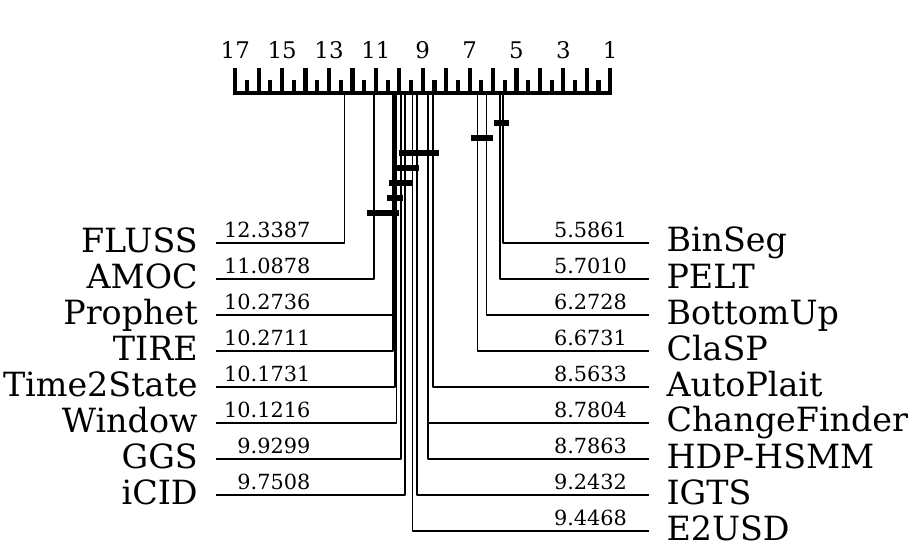}
        \caption{CPD, \emph{Non-Guided}.}
    \end{subfigure}\hfill
    \begin{subfigure}[t]{0.49\linewidth}
        \includegraphics[width=\linewidth]{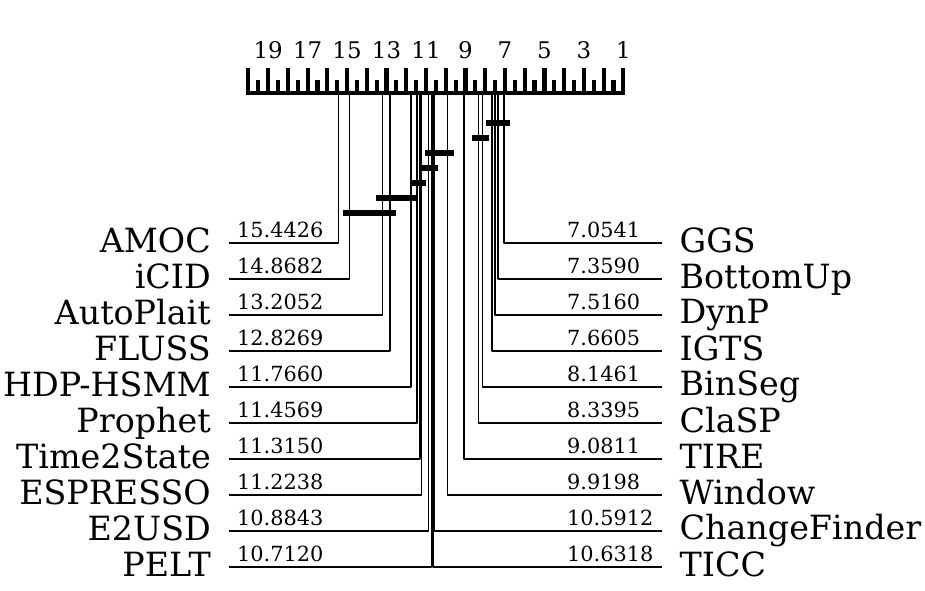}
        \caption{CPD, \emph{Guided}.}
    \end{subfigure}

    \vspace{0.6em}
    \begin{subfigure}[t]{0.42\linewidth}
        \includegraphics[width=\linewidth]{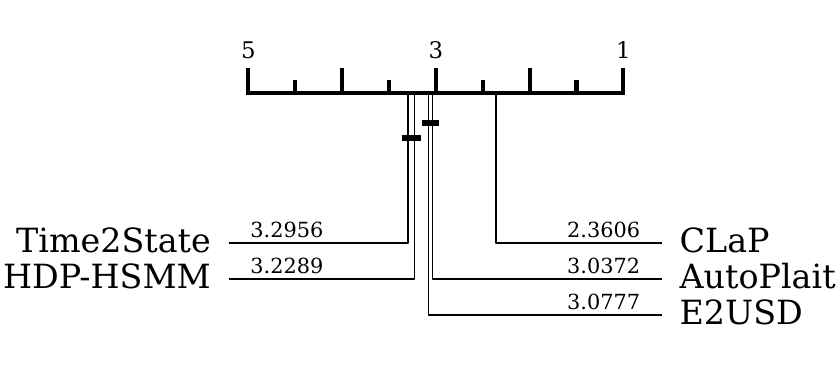}
        \caption{SD, \emph{Non-Guided}.}
    \end{subfigure}\hfill
    \begin{subfigure}[t]{0.42\linewidth}
        \includegraphics[width=\linewidth]{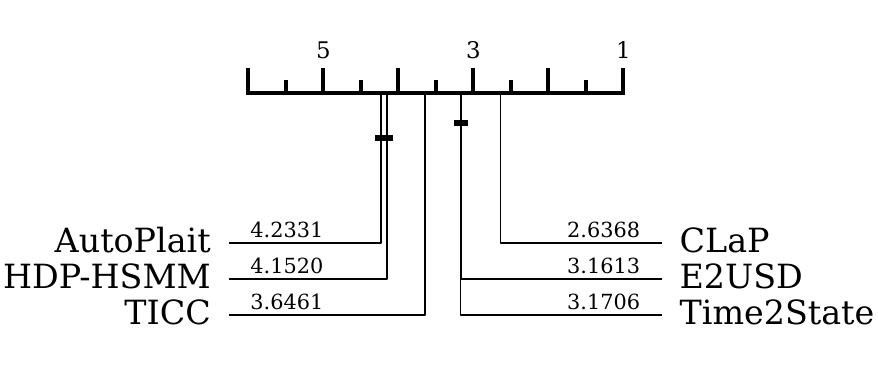}
        \caption{SD, \emph{Guided}.}
    \end{subfigure}
    \caption{\textbf{Critical-difference diagrams of the Default configurations} over the 592 series, with the test of Figure~\ref{fig:cpd_main} (two-sided Wilcoxon signed-rank test with Holm correction, $\alpha = 0.1$).}
    \label{fig:cd_default}
\end{figure}

\paragraph{Ranking without tuning.} Figure~\ref{fig:cd_default} repeats the critical-difference diagrams of RQ1 on the Default configurations, the ranking a practitioner obtains without any tuning. The leaders largely persist. In CPD without $n_{CP}$, BinSeg and PELT form a tied lead (mean ranks $5.59$-$5.70$), followed by BottomUp and ClaSP ($6.27$-$6.67$), whereas Window, fourth once tuned (mean rank $6.91$), falls to $10.12$. With $n_{CP}$, GGS, BottomUp, DynP and IGTS are statistically tied at the top ($7.05$-$7.66$). In SD, CLaP leads alone in both modes (mean rank $2.36$ without $n_S$, $2.64$ with it).

\section{CPD vs.\ SD in Each Supervision Mode} \label{app:rq2_by_mode}

Figure~\ref{fig:regression_analysis} pools the runs of both supervision modes. Table~\ref{tab:rq2_by_mode} repeats its statistics for each mode separately. Envelopes are computed within each mode; the tail compositions use the pooled thresholds ($r<-0.27$, $r>+0.36$) on the runs for which the SMS error decomposition is available. Both modes lead to the same conclusions: a residual concentrated around zero, a BiC-SMS correlation of $0.76$-$0.79$, rare large SMS drops at high BiC, a lower tail split between \emph{Delay} and \emph{Missing} errors and an upper tail dominated by \emph{Isolation}. The Guided envelope is wider on the upper side ($+0.40$ against $+0.31$), partly because of the methods that run only in this mode ($+0.36$ without TICC and Hidalgo).

\begin{table}[h]
    \centering
    \small
    \caption{RQ2 statistics in each supervision mode (SD-capable methods, 8 main datasets, all default and grid runs). The Guided runs of AutoPlait are its Non-Guided ones ($546$ runs), counted once in the pooled column.}
    \label{tab:rq2_by_mode}
    \begin{tabular}{lccc}
    \toprule
     & \textbf{Pooled} & \textbf{Non-Guided} & \textbf{Guided} \\
    \midrule
    Runs & $97{,}076$ & $46{,}825$ & $50{,}797$ \\
    Pearson correlation of BiC and SMS & $0.79$ & $0.79$ & $0.76$ \\
    Runs with $|r| \leq 0.05$ & $33.8\%$ & $33.2\%$ & $34.6\%$ \\
    $95\%$ envelope of $r$ & $[-0.27, +0.36]$ & $[-0.26, +0.31]$ & $[-0.28, +0.40]$ \\
    $\mathrm{SMS} < 0.6$ among runs with $\mathrm{BiC} \geq 0.8$ & $3.7\%$ & $2.7\%$ & $4.0\%$ \\
    Lower tail ($r<-0.27$): \emph{Delay} / \emph{Missing} share & $36\%$ / $40\%$ & $39\%$ / $37\%$ & $36\%$ / $40\%$ \\
    Upper tail ($r>+0.36$): \emph{Isolation} share & $88\%$ & $88\%$ & $88\%$ \\
    \bottomrule
    \end{tabular}
\end{table}

The quality of the proxy also depends on the series and on the method. It is tightest when no state recurs ($|r| \leq 0.05$ for $41\%$ of the runs, against $23\%$ when states recur) and weakest on PUMP, whose $41$-$43$ states recur within every cycle (median $r=-0.18$): boundaries can be right while labels are wrong. Among methods, CLaP, whose labelling is the most accurate, keeps $60\%$ of its runs within $|r| \leq 0.05$, against $26$-$27\%$ for E2USD and Time2State.

\section{Runtime-Accuracy Distributions} \label{app:runtime_sd}

\begin{figure}[H]
    \centering
    \includegraphics[width=\linewidth]{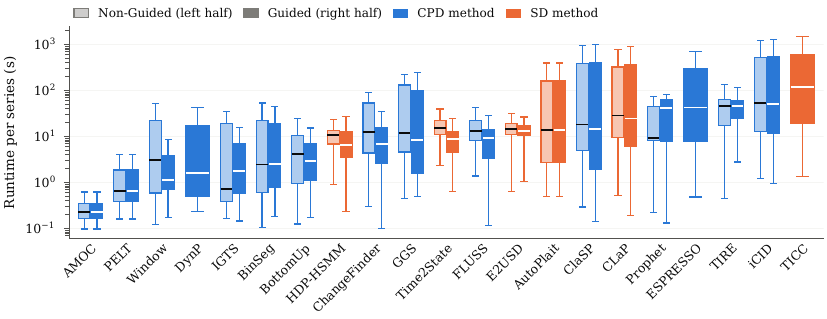}
    \caption{\textbf{Per-series runtime distribution on the CPD task} (tuned configuration on BiC, 8 main datasets), for the CPD methods (blue) and the SD methods scored on CPD (orange), sorted by median runtime. The light left half of each box reports the \emph{Non-Guided} runtimes, the dark right half the \emph{Guided} ones; a method run in one mode only is drawn as a full-width box. Boxes span the interquartile range, whiskers $1.5$ times it.}
    \label{fig:runtime_split_cpd_app}
\end{figure}

\begin{figure}[H]
    \centering
    \begin{subfigure}[t]{0.30\linewidth}
        \includegraphics[width=\linewidth]{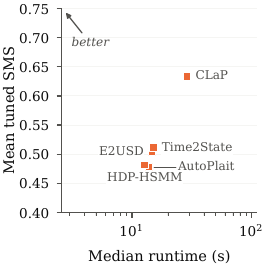}
        \caption{Non-Guided.}
    \end{subfigure}
    \begin{subfigure}[t]{0.30\linewidth}
        \includegraphics[width=\linewidth]{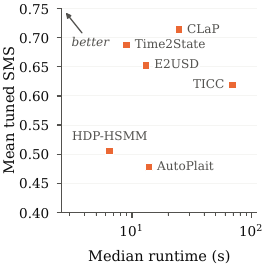}
        \caption{Guided.}
    \end{subfigure}\hfill
    \begin{subfigure}[t]{0.36\linewidth}
        \centerline{\includegraphics[width=\linewidth]{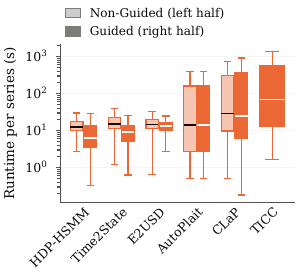}}
        \caption{Runtime distribution.}
        \label{fig:runtime_by_method_sd}
    \end{subfigure}
    \caption{\textbf{Runtime profile of SD methods on \texttt{TSB-SEG}}. (a, b) Mean tuned SMS against the median runtime of the tuned configuration, as in Figure~\ref{fig:rq3_runtime}; (c) per-series runtime of the tuned configuration, with the conventions of Figure~\ref{fig:runtime_split_cpd_app}.}
    \label{fig:rq3_runtime_sd}
\end{figure}

Figure~\ref{fig:runtime_split_cpd_app} reports the per-series runtime distributions for CPD methods, corresponding to the aggregated runtime-accuracy trade-off shown in RQ3 (Figure~\ref{fig:rq3_runtime}). Figure~\ref{fig:rq3_runtime_sd} reports the equivalent runtime-accuracy trade-off and per-series runtime distributions for the SD task.

Unlike CPD, where the fastest methods are also among the most accurate, the SD methods all sit in the mid-range of runtimes: E2USD, AutoPlait, Time2State and HDP-HSMM take a median of $6$-$15$\,s per series, CLaP $25$-$29$\,s and TICC $69$\,s. Supplying the ground-truth number of states (\emph{Guided}) tends to reduce runtimes for most of these methods. Combined with the SMS scores reported in Section~\ref{sec:experiments} (RQ1), CLaP is the most accurate SD method in both modes, at a moderate cost.

\section{Hyperparameter Influence}
\label{app:hyperparams}

\begin{figure}[H]
    \centering
    \includegraphics[width=\linewidth]{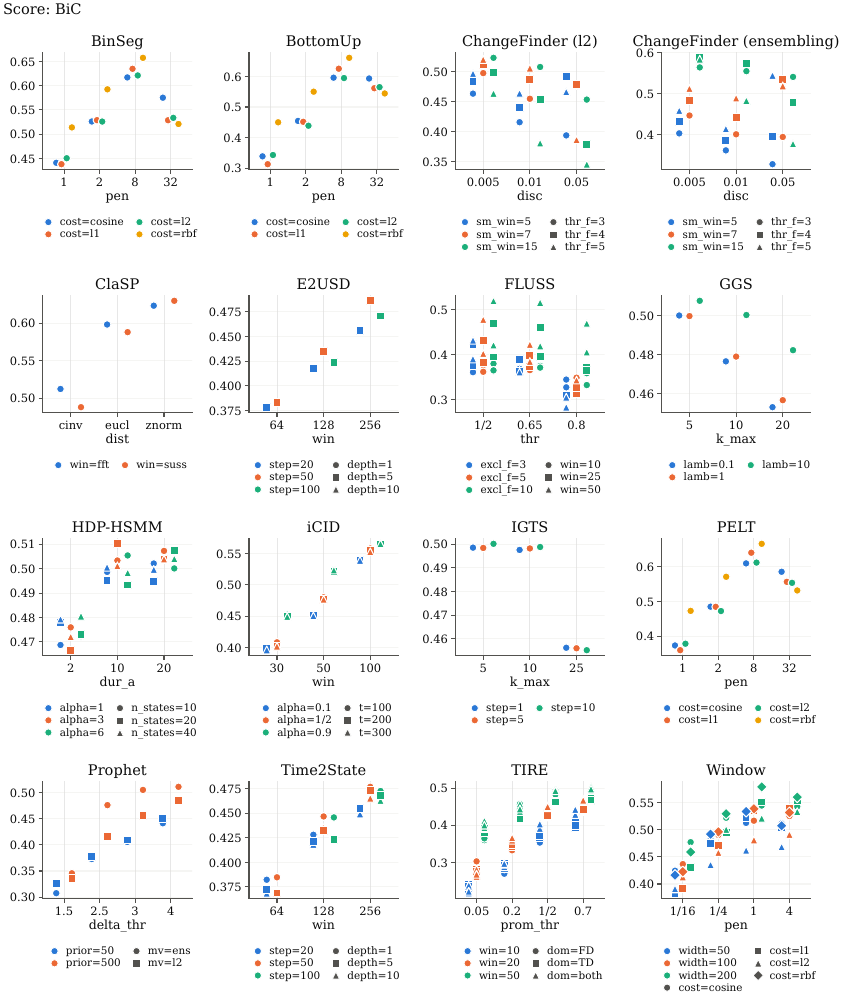}
    \caption{\textbf{Per-algorithm hyperparameter influence on CPD performance, Non-Guided mode.} Score: BiC (top left). For each algorithm, the most influential parameter (largest cross-value spread of marginal means) is used as the x-axis; colour and marker shape encode up to two additional parameters (legend below each panel). Parameter and value names are abbreviated for readability (e.g.\ \texttt{win} = \texttt{window\_size}, \texttt{cost} = segment cost, \texttt{pen} = multiple $X$ of the BIC penalty, \texttt{thr} = threshold, \texttt{cinv} = \texttt{cinvariant\_euclidean\_distance}); the full mapping is given in the source code released with the benchmark.}
    \label{fig:hyperparams_cpd_ng}
\end{figure}

\begin{figure}[H]
    \centering
    \includegraphics[width=\linewidth]{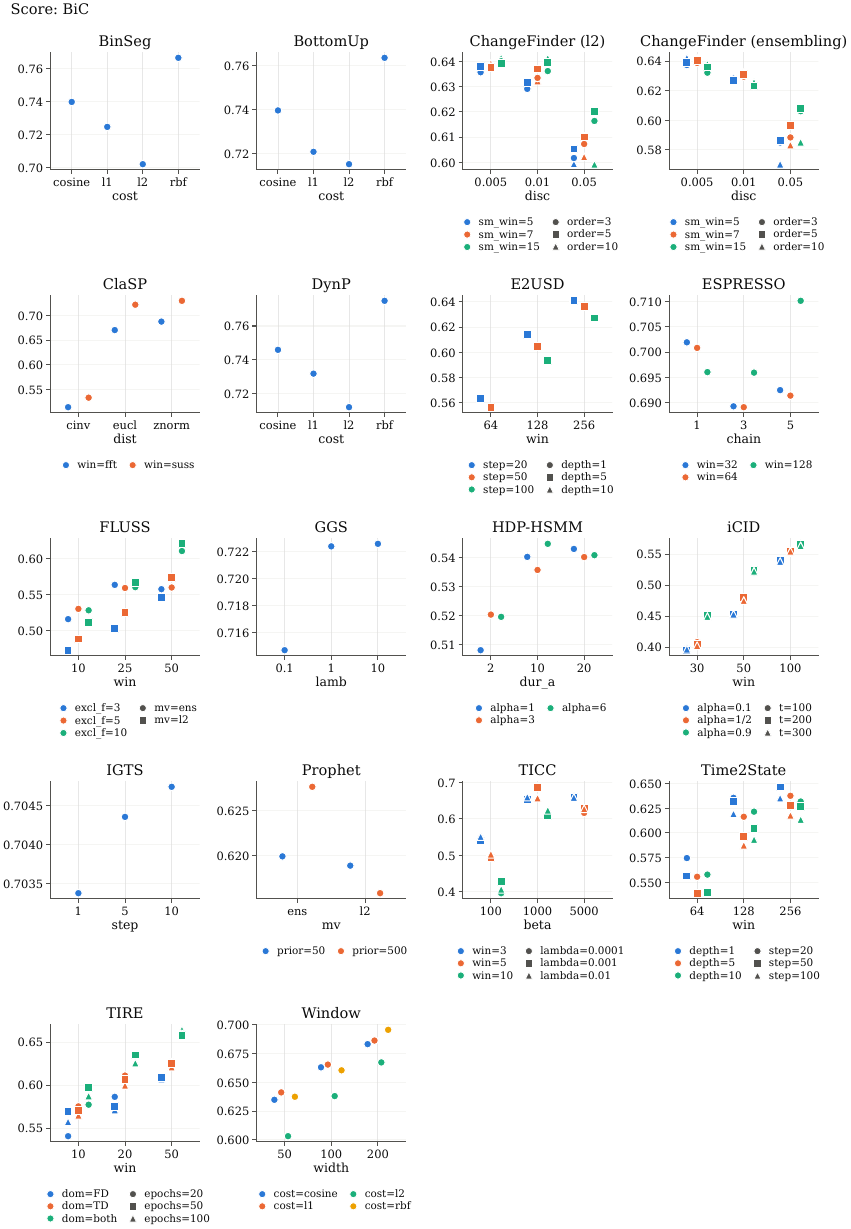}
    \caption{\textbf{Per-algorithm hyperparameter influence on CPD performance, Guided mode.} Same conventions as Figure~\ref{fig:hyperparams_cpd_ng}; PELT, which cannot use $n_{CP}$, is shown there only. Compared to the Non-Guided figure, knowing $n_{CP}$ generally lifts the response surface and shrinks its vertical spread; TICC, run in this mode only, remains strongly responsive to its grid.}
    \label{fig:hyperparams_cpd_g}
\end{figure}

\begin{figure}[H]
    \centering
    \includegraphics[width=\linewidth]{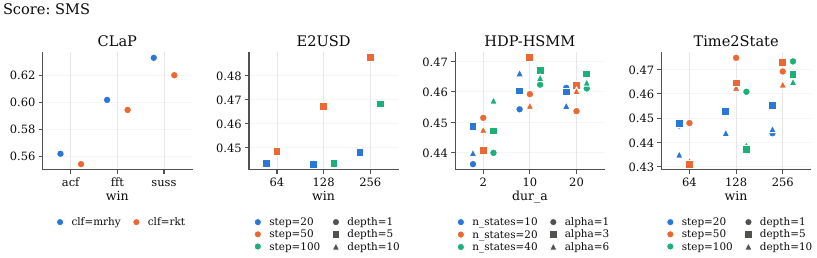}
    \caption{\textbf{Per-algorithm hyperparameter influence on SD performance, Non-Guided mode.} Score: SMS (top left). Same conventions as Figure~\ref{fig:hyperparams_cpd_ng}.}
    \label{fig:hyperparams_sd_ng}
\end{figure}

\begin{figure}[H]
    \centering
    \includegraphics[width=\linewidth]{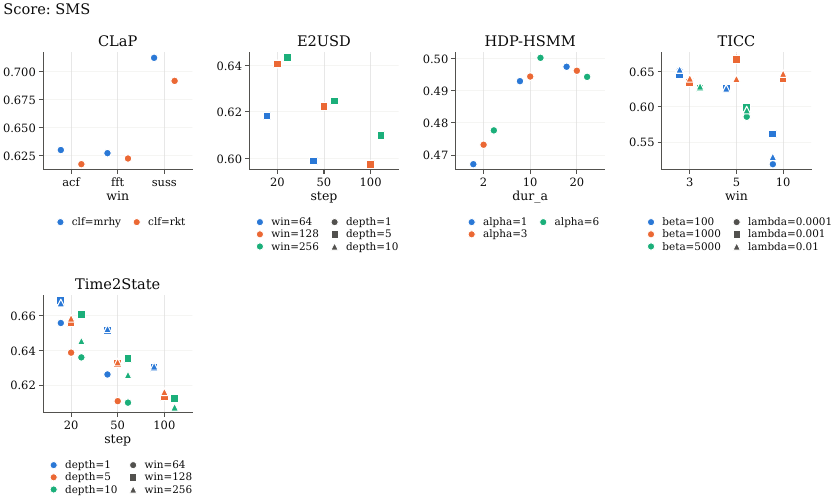}
    \caption{\textbf{Per-algorithm hyperparameter influence on SD performance, Guided mode.} Same conventions as Figure~\ref{fig:hyperparams_sd_ng}.}
    \label{fig:hyperparams_sd_g}
\end{figure}

To complement the aggregate tuning gap visible in the strip plots of RQ1 and the per-(algorithm, dataset) tuned/median/worst-case spreads of Appendix~\ref{app:main_pool_full_grid}, we report a per-algorithm, per-hyperparameter view of the score landscape. For every (algorithm, supervision mode) pair and each task (CPD scored under BiC, SD scored under SMS), Figures~\ref{fig:hyperparams_cpd_ng}-\ref{fig:hyperparams_sd_g} plot the marginal score of each completed grid combination as a function of its hyperparameters. The most influential parameter, defined as the one whose marginal-mean score has the largest cross-value spread, is used as the primary x-axis; up to two additional parameters are encoded by colour and marker shape (legend below each panel). Combinations that did not complete on every (dataset, series) pair are nonetheless retained as long as their completion fraction exceeds $65\%$, with missing series imputed at score $0$ so that incomplete configurations are visible but penalised. The five extended-scope algorithms (BOCD, EAgglo, TS-CP$^2$, tGLAD, Hidalgo) are excluded from these figures because their grids are either unrealised or exercised on too narrow a subset of datasets to support a meaningful marginal analysis, as are AutoPlait (a single configuration) and AMOC (a single change point, whose only parameter is a minimum segment length); PELT, which cannot use $n_{CP}$, appears in the Non-Guided figure only.

\paragraph{Reading the figures.} Three readings are useful. (i)~\emph{Vertical spread along the x-axis} measures how strongly the primary parameter shapes the score; flat panels (e.g.\ HDP-HSMM, IGTS) indicate a method that is robust to its grid, whereas steep trends (e.g.\ the penalty of BinSeg, BottomUp, PELT and Window, or \texttt{window\_size} for iCID, Time2State and E2USD) point to a knob that practitioners must tune. (ii)~\emph{Colour scatter at fixed x} captures the residual influence of the secondary parameter; tight colour clusters (e.g.\ ClaSP and CLaP, whose window-selection method barely matters) indicate robustness, while wide vertical bands (e.g.\ FLUSS, whose exclusion factor and window matter on top of its threshold) reveal a meaningful interaction. (iii)~\emph{Comparison NG~$\to$~G} (Figure~\ref{fig:hyperparams_cpd_ng} vs.\ Figure~\ref{fig:hyperparams_cpd_g}) shows how knowing $n_{CP}/n_S$ flattens or sharpens the response surface: most algorithms exhibit a clear upward shift accompanied by reduced grid sensitivity, suggesting that the supervision signal absorbs much of the per-series tuning headroom.

\paragraph{Take-aways.} The figures confirm and refine the recommendations of RQ3. (a)~For iCID, Time2State and E2USD, and for TIRE in Guided mode, the primary knob is the window size, with a monotone response; without $K$, TIRE and Prophet respond first to their detection threshold, best at the highest value of the grid; ClaSP responds mostly to its distance and CLaP to its window-selection method, within a narrow range: a single coarse sweep is therefore sufficient. The exception is ClaSP's complexity-invariant distance, which exceeds the time budget on PUMP and USC-HAD and drops in Guided mode, where its missing series are imputed at score $0$. (b)~Penalty-based methods (BinSeg, BottomUp, PELT, Window) respond mainly to the penalty, peaking near $8\times$ the BIC penalty ($1\times$ for Window); the cost function matters at low penalties, where the RBF cost degrades least, and in Guided mode, where RBF is best for DynP, BinSeg and BottomUp. (c)~Without $K$, ChangeFinder is split by its multivariate strategy: the \emph{ensembling} variant is markedly more sensitive to its smoothing window than the \emph{$\ell_2$} variant. (d)~One method stands out as inherently sensitive: TICC (jointly responsive to both $\beta$ and $\lambda$), confirming that it requires dataset-specific calibration to perform competitively. HDP-HSMM, and Prophet with $K$, are by contrast nearly flat over their grids, the duration-prior strength being HDP-HSMM's only visible knob.

\section{Sensitivity to Input Features} \label{app:feature_sensitivity}

We complement the runtime and accuracy analyses of RQ3 with a per-algorithm sensitivity study. For every (algorithm, supervision mode) pair we compute the Spearman rank correlation between the algorithm's \emph{tuned score} and four per-series structural features: length ($n$), number of dimensions ($d$), number of change points ($n_{CP}$) and number of states ($n_S$). Tuned scores are computed as in Section~\ref{sec:exp_strategy} (one configuration per dataset, applied uniformly to all series of that dataset).

\begin{figure}[h]
    \centering
    \includegraphics[width=\textwidth]{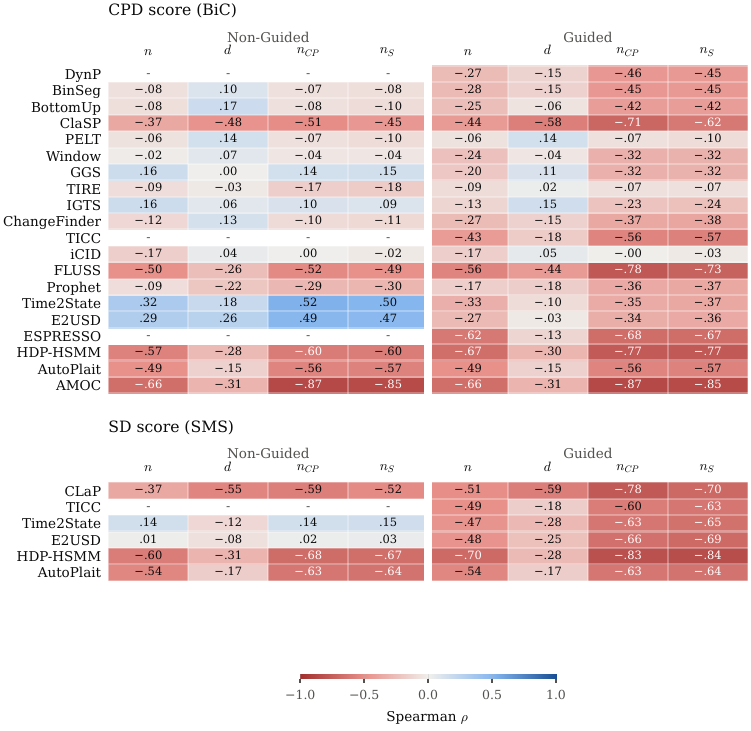}
    \caption{\textbf{Sensitivity of the tuned score to input series features (Spearman $\rho$).} Top: CPD score (BiC); bottom: SD score (SMS). One row per algorithm, the four features in each supervision mode as columns. Red: the score decreases as the feature grows ($\rho < 0$); blue: it increases; grey: no monotonic dependence.}
    \label{fig:feature_sensitivity}
\end{figure}

Figure~\ref{fig:feature_sensitivity} reports the resulting heatmaps for CPD (top) and SD (bottom). Each algorithm is one row, with the four features in the Non-Guided and the Guided mode as columns (``-'' marks a mode the algorithm does not run in); algorithms are ordered by mean tuned score over the modes they run in (descending). Two patterns stand out: in Non-Guided mode, the penalised detectors are nearly insensitive to the four features, ClaSP, CLaP, FLUSS, HDP-HSMM and AMOC degrade on series with many events ($\rho$ from $-0.5$ to $-0.9$), and Time2State and E2USD improve on them ($\rho$ up to $0.5$); in Guided mode, almost every method degrades as the number of events grows, the series with many events remaining hard even when their number is known.

\section{Performance on Long Series} \label{app:pamap2_case_study}

\begin{figure}[t]
    \centering
    \includegraphics[width=\linewidth]{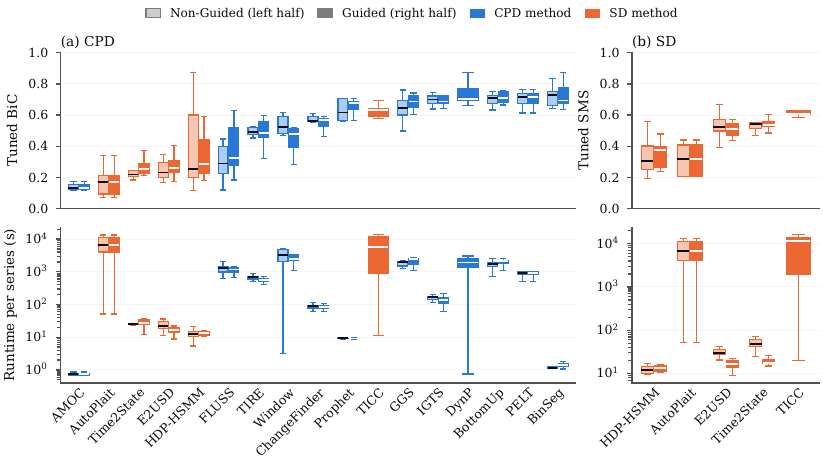}
    \caption{\textbf{Analysis on the PAMAP2 dataset (up to $447$k samples per series)}, for the 17 methods that process its 9 series. Top: tuned score of each series, (a) BiC for CPD, (b) SMS for SD; bottom: runtime of the tuned configuration on each series. Same conventions as Figure~\ref{fig:runtime_split_cpd_app}; methods sorted by median tuned score.}
    \label{fig:pamap2_analysis}
\end{figure}

This section provides additional details on the PAMAP2 case study discussed in RQ3. The PAMAP2 dataset contains 9 long multivariate series of $8$k to $447$k samples. Only 17 out of the 26 algorithms process all 9 series within the computational budget. Figure~\ref{fig:pamap2_analysis} presents the score distributions and runtime variations for the methods evaluated on this dataset, for both CPD and SD tasks under both supervision modes. Note, however, that long series can in practice be subsampled to recover the desired behaviour at scale, as noted in~\cite{Wang2023Time2State}. Runtimes follow the segment cost of the tuned configuration: $\ell_2$ for BinSeg (about $1$\,s per series), RBF for PELT and BottomUp (about $10^3$\,s), which beats their $\ell_2$ configuration ($90$-$100$\,s) by $0.01$-$0.02$ BiC.

\paragraph{Window.} On PAMAP2, Window scores higher in the Non-Guided mode (tuned BiC $0.54$) than in the Guided one ($0.47$). In Guided mode, Window must stop at the true number of peaks of its discrepancy curve, and only a third of them fall within the $F_1$ margin of a true change point. Its best Non-Guided configuration places about three times as many change points as the ground truth, which recovers twice as many true change points and raises the covering in both directions.

\section{Extending \texttt{TSB-SEG}} \label{app:extending}

\texttt{TSB-SEG} is meant to grow. New methods and datasets enter through the two open-source components that produced every number of this paper: the \texttt{tsseg} library (detectors and metrics) and the benchmark pipeline (configurations, runs and analysis). The \texttt{TSB-SEG} repository is frozen on this paper and pins the version of \texttt{tsseg} that produced it; \texttt{tsseg} keeps evolving, and new methods enter \texttt{tsseg} first.

\paragraph{Adding an algorithm.} Candidates include the recent methods collected in community reading lists~\cite{IssuesLzz19980125Awesometimeseriessegmentationpapers}. A method is contributed as a pull request to \texttt{tsseg} that provides: (i)~a detector class following the common API (\texttt{fit}/\texttt{predict} on an $n \times d$ array, returning either the sorted change-point indices or one state label per sample) and declaring its capabilities (task, univariate and/or multivariate input, whether it can take $K$); (ii)~a declarative schema of its parameters, with types and admissible ranges checked at fit time; (iii)~its default parameters (those recommended by the original authors, or a documented choice when they give none) and a search grid of $2$-$4$ values for its $2$-$4$ most influential parameters, flagging the parameters that $K$ replaces in Guided mode; (iv)~a provenance note (original paper, source code, licence). The generic test suite of \texttt{tsseg} checks the API contract of every registered detector. The benchmark then runs the default configuration and the grid on every dataset in both supervision modes, under the same budget as all other methods (Appendix~\ref{app:compute}); the method joins the main pool if it completes (nearly) every series within this budget, and the extended-scope pool otherwise.

\paragraph{Adding a dataset.} A dataset is contributed with a loader returning, for each series, the observations $T \in \mathbb{R}^{n \times d}$ and a dense state sequence $Q$ (one label per sample, the change points being derived from label changes), together with its provenance, licence and annotation protocol (Appendix~\ref{app:datasets}). Datasets annotated with boundaries only can enter by giving each segment its own state, but they then inform CPD only.

\paragraph{Reporting.} Every run is logged per series with all the metrics of Section~\ref{sec:metrics}, so that a new method or dataset enters every table and figure of this paper by re-running the analysis scripts; rankings and statistical tests are then recomputed over the extended collection.

\section{Per-(algorithm, dataset) Breakdown for the Main Pool}
\label{app:main_pool_full_grid}

For completeness, Tables~\ref{tab:main_pool_cpd}-\ref{tab:main_pool_sd} report the per-(algorithm, dataset) decomposition of the main-pool 21 algorithms into the four configurations of interest (\emph{Default}, \emph{Worst-case}, \emph{Median}, \emph{Tuned}) in both Non-Guided and Guided modes. Each cell averages the series that the method completed, without imputation (e.g.\ TICC in Guided mode on PUMP, $40$ of $120$ series), whereas the means, ranks and figures over the $592$ series count the missing ones as $0$ (Appendix~\ref{app:compute}). The worst-case/median/tuned triple quantifies the \emph{tuning gap} per cell: a small spread indicates a stable algorithm whose grid is largely interchangeable, whereas a wide spread signals a method whose realistic deployment requires a dataset-specific selection.

\begingroup
\setlength{\tabcolsep}{3pt}\renewcommand{\arraystretch}{0.95}\small
\begin{longtable}{l l c c c c c c c c c c}
\caption{\textbf{Main-pool algorithms on CPD: per-(algorithm, dataset) default / worst-case / median / tuned BiC, in both Non-Guided (NG) and Guided (G) modes.} The SD methods (last block) are scored on the change points induced by their labels, except CLaP, which labels the segments of ClaSP and is not scored separately. $\tilde{t}$ is the median wall-clock of the tuned configuration in seconds. ``-'' = (mode, dataset) pair not evaluated for that algorithm (e.g.\ intrinsically Guided methods have no NG entries).} \label{tab:main_pool_cpd} \\
\toprule
\multicolumn{2}{c}{} & \multicolumn{5}{c}{\textbf{Non-Guided}} & \multicolumn{5}{c}{\textbf{Guided}} \\ \cmidrule(lr){3-7} \cmidrule(lr){8-12}
 \textbf{Algo} & \textbf{Dataset}  & Def. & Worst & Med. & Tuned & $\tilde{t}$ [s]  & Def. & Worst & Med. & Tuned & $\tilde{t}$ [s] \\
\midrule
\endfirsthead
\multicolumn{12}{l}{\emph{(continued from previous page)}} \\
\toprule
\multicolumn{2}{c}{} & \multicolumn{5}{c}{\textbf{Non-Guided}} & \multicolumn{5}{c}{\textbf{Guided}} \\ \cmidrule(lr){3-7} \cmidrule(lr){8-12}
 \textbf{Algo} & \textbf{Dataset}  & Def. & Worst & Med. & Tuned & $\tilde{t}$ [s]  & Def. & Worst & Med. & Tuned & $\tilde{t}$ [s] \\
\midrule
\endhead
\midrule \multicolumn{12}{r}{\emph{(continued on next page)}} \\
\endfoot
\bottomrule
\endlastfoot
\multirow{8}{*}{\textbf{AMOC}} & ActRecTut & 0.24 & 0.24 & 0.24 & 0.24 & 0.2 & 0.24 & 0.24 & 0.24 & 0.24 & 0.2 \\*
 & HAS & 0.61 & 0.61 & 0.61 & 0.61 & 0.2 & 0.61 & 0.61 & 0.61 & 0.61 & 0.2 \\*
 & MoCap & 0.30 & 0.30 & 0.30 & 0.30 & 0.2 & 0.30 & 0.30 & 0.30 & 0.30 & 0.2 \\*
 & PUMP & 0.07 & 0.07 & 0.07 & 0.07 & 0.2 & 0.07 & 0.07 & 0.07 & 0.07 & 0.2 \\*
 & SKAB & 0.71 & 0.71 & 0.72 & 0.72 & 0.3 & 0.71 & 0.71 & 0.72 & 0.72 & 0.3 \\*
 & TSSB & 0.42 & 0.42 & 0.42 & 0.42 & 0.3 & 0.42 & 0.42 & 0.42 & 0.42 & 0.3 \\*
 & USC-HAD & 0.24 & 0.24 & 0.24 & 0.24 & 0.3 & 0.24 & 0.24 & 0.24 & 0.24 & 0.3 \\*
 & UTSA & 0.64 & 0.64 & 0.64 & 0.64 & 0.3 & 0.64 & 0.64 & 0.64 & 0.64 & 0.3 \\
\midrule
\multirow{8}{*}{\textbf{BinSeg}} & ActRecTut & 0.32 & 0.31 & 0.41 & 0.77 & 42.8 & 0.43 & 0.43 & 0.60 & 0.73 & 30.3 \\*
 & HAS & 0.66 & 0.30 & 0.53 & 0.73 & 1.1 & 0.78 & 0.78 & 0.80 & 0.82 & 1.1 \\*
 & MoCap & 0.66 & 0.14 & 0.37 & 0.72 & 1.0 & 0.66 & 0.66 & 0.71 & 0.82 & 1.7 \\*
 & PUMP & 0.54 & 0.31 & 0.57 & 0.62 & 22.6 & 0.62 & 0.60 & 0.62 & 0.62 & 19.2 \\*
 & SKAB & 0.77 & 0.43 & 0.56 & 0.78 & 0.2 & 0.76 & 0.76 & 0.78 & 0.80 & 0.2 \\*
 & TSSB & 0.42 & 0.33 & 0.40 & 0.48 & 0.6 & 0.42 & 0.42 & 0.57 & 0.70 & 1.1 \\*
 & USC-HAD & 0.81 & 0.55 & 0.78 & 0.87 & 52.7 & 0.86 & 0.82 & 0.85 & 0.87 & 53.2 \\*
 & UTSA & 0.55 & 0.34 & 0.49 & 0.63 & 1.8 & 0.66 & 0.66 & 0.75 & 0.80 & 2.7 \\
\midrule
\multirow{8}{*}{\textbf{BottomUp}} & ActRecTut & 0.46 & 0.31 & 0.65 & 0.78 & 21.7 & 0.63 & 0.61 & 0.69 & 0.74 & 8.7 \\*
 & HAS & 0.61 & 0.23 & 0.49 & 0.72 & 1.6 & 0.78 & 0.78 & 0.81 & 0.83 & 1.7 \\*
 & MoCap & 0.62 & 0.13 & 0.33 & 0.75 & 1.7 & 0.58 & 0.58 & 0.66 & 0.80 & 1.5 \\*
 & PUMP & 0.56 & 0.38 & 0.60 & 0.63 & 8.6 & 0.63 & 0.63 & 0.63 & 0.63 & 4.1 \\*
 & SKAB & 0.77 & 0.40 & 0.55 & 0.79 & 0.2 & 0.78 & 0.78 & 0.80 & 0.82 & 0.2 \\*
 & TSSB & 0.40 & 0.15 & 0.38 & 0.47 & 1.5 & 0.55 & 0.55 & 0.58 & 0.65 & 1.8 \\*
 & USC-HAD & 0.78 & 0.37 & 0.66 & 0.83 & 33.4 & 0.80 & 0.78 & 0.82 & 0.87 & 44.7 \\*
 & UTSA & 0.44 & 0.06 & 0.35 & 0.64 & 11.3 & 0.71 & 0.70 & 0.73 & 0.78 & 9.8 \\
\midrule
\multirow{8}{*}{\textbf{ChangeFinder}} & ActRecTut & 0.49 & 0.10 & 0.42 & 0.58 & 84.5 & 0.42 & 0.30 & 0.45 & 0.56 & 78.2 \\*
 & HAS & 0.58 & 0.19 & 0.47 & 0.68 & 10.5 & 0.68 & 0.62 & 0.67 & 0.71 & 10.0 \\*
 & MoCap & 0.36 & 0.16 & 0.33 & 0.63 & 7.3 & 0.66 & 0.33 & 0.48 & 0.66 & 1.9 \\*
 & PUMP & 0.35 & 0.04 & 0.41 & 0.62 & 66.2 & 0.56 & 0.50 & 0.56 & 0.58 & 7.0 \\*
 & SKAB & 0.52 & 0.33 & 0.48 & 0.67 & 2.8 & 0.64 & 0.40 & 0.52 & 0.69 & 0.4 \\*
 & TSSB & 0.49 & 0.41 & 0.45 & 0.50 & 1.8 & 0.62 & 0.60 & 0.63 & 0.66 & 1.5 \\*
 & USC-HAD & 0.59 & 0.18 & 0.50 & 0.66 & 55.1 & 0.63 & 0.48 & 0.61 & 0.64 & 64.1 \\*
 & UTSA & 0.51 & 0.26 & 0.50 & 0.65 & 4.4 & 0.68 & 0.61 & 0.65 & 0.71 & 2.9 \\
\midrule
\multirow{8}{*}{\textbf{ClaSP}} & ActRecTut & 0.44 & 0.20 & 0.36 & 0.42 & 510 & 0.31 & 0.30 & 0.35 & 0.41 & 7.8k \\*
 & HAS & 0.74 & 0.58 & 0.63 & 0.74 & 10.6 & 0.80 & 0.70 & 0.75 & 0.80 & 6.8 \\*
 & MoCap & 0.75 & 0.69 & 0.75 & 0.79 & 13.8 & 0.81 & 0.76 & 0.80 & 0.83 & 14.6 \\*
 & PUMP & 0.20 & 0.19 & 0.29 & 0.42 & 607 & 0.42 & 0.32 & 0.39 & 0.45 & 603 \\*
 & SKAB & 0.47 & 0.46 & 0.54 & 0.63 & 1.3 & 0.69 & 0.60 & 0.66 & 0.71 & 0.6 \\*
 & TSSB & 0.85 & 0.83 & 0.85 & 0.87 & 2.5 & 0.90 & 0.88 & 0.90 & 0.91 & 1.0 \\*
 & USC-HAD & 0.74 & 0.61 & 0.72 & 0.81 & 448 & 0.74 & 0.70 & 0.76 & 0.82 & 469 \\*
 & UTSA & 0.73 & 0.52 & 0.64 & 0.73 & 13.8 & 0.95 & 0.89 & 0.93 & 0.95 & 7.1 \\
\midrule
\multirow{8}{*}{\textbf{DynP}} & ActRecTut & - & - & - & - & - & 0.56 & 0.56 & 0.67 & 0.74 & 34.0 \\*
 & HAS & - & - & - & - & - & 0.78 & 0.78 & 0.81 & 0.83 & 0.8 \\*
 & MoCap & - & - & - & - & - & 0.68 & 0.68 & 0.74 & 0.84 & 1.1 \\*
 & PUMP & - & - & - & - & - & 0.62 & 0.61 & 0.62 & 0.63 & 21.4 \\*
 & SKAB & - & - & - & - & - & 0.78 & 0.78 & 0.79 & 0.81 & 0.3 \\*
 & TSSB & - & - & - & - & - & 0.49 & 0.49 & 0.59 & 0.72 & 0.5 \\*
 & USC-HAD & - & - & - & - & - & 0.85 & 0.83 & 0.85 & 0.88 & 31.3 \\*
 & UTSA & - & - & - & - & - & 0.68 & 0.68 & 0.76 & 0.81 & 1.9 \\
\midrule
\multirow{8}{*}{\textbf{ESPRESSO}} & ActRecTut & - & - & - & - & - & 0.49 & 0.48 & 0.49 & 0.51 & 5.1k \\*
 & HAS & - & - & - & - & - & 0.78 & 0.78 & 0.78 & 0.78 & 51.8 \\*
 & MoCap & - & - & - & - & - & 0.73 & 0.71 & 0.74 & 0.75 & 103 \\*
 & PUMP & - & - & - & - & - & 0.51 & 0.49 & 0.50 & 0.52 & 2.1k \\*
 & SKAB & - & - & - & - & - & 0.83 & 0.80 & 0.83 & 0.83 & 6.4 \\*
 & TSSB & - & - & - & - & - & 0.48 & 0.48 & 0.49 & 0.52 & 4.8 \\*
 & USC-HAD & - & - & - & - & - & 0.80 & 0.79 & 0.79 & 0.80 & 3.0k \\*
 & UTSA & - & - & - & - & - & 0.70 & 0.67 & 0.69 & 0.73 & 32.0 \\
\midrule
\multirow{8}{*}{\textbf{FLUSS}} & ActRecTut & 0.12 & 0.10 & 0.16 & 0.46 & 38.7 & 0.38 & 0.23 & 0.32 & 0.44 & 40.1 \\*
 & HAS & 0.48 & 0.33 & 0.48 & 0.57 & 12.4 & 0.65 & 0.60 & 0.66 & 0.71 & 12.0 \\*
 & MoCap & 0.16 & 0.16 & 0.16 & 0.54 & 1.9 & 0.42 & 0.30 & 0.40 & 0.63 & 1.6 \\*
 & PUMP & 0.08 & 0.04 & 0.07 & 0.46 & 24.5 & 0.22 & 0.10 & 0.19 & 0.41 & 4.0 \\*
 & SKAB & 0.43 & 0.26 & 0.46 & 0.71 & 10.8 & 0.62 & 0.55 & 0.60 & 0.69 & 11.1 \\*
 & TSSB & 0.51 & 0.39 & 0.51 & 0.69 & 1.7 & 0.71 & 0.69 & 0.75 & 0.80 & 1.9 \\*
 & USC-HAD & 0.24 & 0.10 & 0.24 & 0.41 & 24.8 & 0.29 & 0.25 & 0.34 & 0.52 & 21.4 \\*
 & UTSA & 0.54 & 0.20 & 0.44 & 0.63 & 2.2 & 0.84 & 0.82 & 0.86 & 0.91 & 2.1 \\
\midrule
\multirow{8}{*}{\textbf{GGS}} & ActRecTut & 0.57 & 0.46 & 0.57 & 0.78 & 165 & 0.60 & 0.60 & 0.60 & 0.67 & 270 \\*
 & HAS & 0.49 & 0.35 & 0.50 & 0.62 & 8.4 & 0.82 & 0.82 & 0.82 & 0.83 & 3.6 \\*
 & MoCap & 0.85 & 0.62 & 0.75 & 0.85 & 20.4 & 0.92 & 0.91 & 0.92 & 0.93 & 12.7 \\*
 & PUMP & 0.33 & 0.20 & 0.33 & 0.51 & 164 & 0.65 & 0.56 & 0.64 & 0.65 & 352 \\*
 & SKAB & 0.39 & 0.30 & 0.42 & 0.57 & 1.9 & 0.76 & 0.76 & 0.76 & 0.76 & 1.1 \\*
 & TSSB & 0.44 & 0.41 & 0.44 & 0.49 & 3.1 & 0.48 & 0.47 & 0.48 & 0.51 & 1.9 \\*
 & USC-HAD & 0.70 & 0.55 & 0.69 & 0.81 & 137 & 0.72 & 0.67 & 0.72 & 0.79 & 87.6 \\*
 & UTSA & 0.54 & 0.46 & 0.54 & 0.61 & 11.1 & 0.73 & 0.73 & 0.73 & 0.74 & 1.9 \\
\midrule
\multirow{8}{*}{\textbf{IGTS}} & ActRecTut & 0.38 & 0.32 & 0.38 & 0.46 & 29.2 & 0.46 & 0.42 & 0.45 & 0.46 & 75.3 \\*
 & HAS & 0.53 & 0.37 & 0.51 & 0.63 & 0.4 & 0.78 & 0.78 & 0.78 & 0.78 & 0.8 \\*
 & MoCap & 0.62 & 0.59 & 0.61 & 0.64 & 0.8 & 0.64 & 0.64 & 0.64 & 0.64 & 1.4 \\*
 & PUMP & 0.31 & 0.18 & 0.31 & 0.54 & 26.2 & 0.63 & 0.63 & 0.63 & 0.63 & 158 \\*
 & SKAB & 0.42 & 0.19 & 0.41 & 0.60 & 0.4 & 0.81 & 0.81 & 0.81 & 0.83 & 0.6 \\*
 & TSSB & 0.39 & 0.35 & 0.38 & 0.41 & 0.5 & 0.42 & 0.42 & 0.42 & 0.42 & 0.8 \\*
 & USC-HAD & 0.83 & 0.58 & 0.81 & 0.83 & 17.6 & 0.83 & 0.82 & 0.83 & 0.83 & 5.1 \\*
 & UTSA & 0.58 & 0.51 & 0.58 & 0.63 & 0.6 & 0.66 & 0.65 & 0.65 & 0.66 & 0.5 \\
\midrule
\multirow{8}{*}{\textbf{PELT}} & ActRecTut & 0.40 & 0.31 & 0.53 & 0.77 & 7.5 & 0.40 & 0.31 & 0.53 & 0.77 & 7.5 \\*
 & HAS & 0.64 & 0.24 & 0.51 & 0.73 & 0.6 & 0.64 & 0.24 & 0.51 & 0.73 & 0.6 \\*
 & MoCap & 0.61 & 0.13 & 0.34 & 0.77 & 0.6 & 0.61 & 0.13 & 0.34 & 0.77 & 0.6 \\*
 & PUMP & 0.55 & 0.35 & 0.59 & 0.62 & 0.6 & 0.55 & 0.35 & 0.59 & 0.62 & 0.6 \\*
 & SKAB & 0.77 & 0.41 & 0.55 & 0.79 & 0.2 & 0.77 & 0.41 & 0.55 & 0.79 & 0.2 \\*
 & TSSB & 0.40 & 0.21 & 0.39 & 0.47 & 0.4 & 0.40 & 0.21 & 0.39 & 0.47 & 0.4 \\*
 & USC-HAD & 0.81 & 0.50 & 0.71 & 0.86 & 9.1 & 0.81 & 0.50 & 0.71 & 0.86 & 9.1 \\*
 & UTSA & 0.50 & 0.14 & 0.39 & 0.63 & 1.7 & 0.50 & 0.14 & 0.39 & 0.63 & 1.7 \\
\midrule
\multirow{8}{*}{\textbf{Prophet}} & ActRecTut & 0.70 & 0.52 & 0.62 & 0.70 & 9.6 & 0.40 & 0.40 & 0.40 & 0.41 & 71.1 \\*
 & HAS & 0.48 & 0.28 & 0.44 & 0.58 & 48.0 & 0.65 & 0.65 & 0.66 & 0.67 & 46.2 \\*
 & MoCap & 0.51 & 0.34 & 0.46 & 0.55 & 8.6 & 0.51 & 0.40 & 0.46 & 0.54 & 7.3 \\*
 & PUMP & 0.25 & 0.20 & 0.28 & 0.43 & 8.3 & 0.52 & 0.52 & 0.53 & 0.53 & 74.6 \\*
 & SKAB & 0.47 & 0.10 & 0.32 & 0.67 & 39.7 & 0.61 & 0.61 & 0.67 & 0.76 & 39.3 \\*
 & TSSB & 0.46 & 0.19 & 0.30 & 0.49 & 5.5 & 0.59 & 0.58 & 0.58 & 0.59 & 4.5 \\*
 & USC-HAD & 0.69 & 0.48 & 0.63 & 0.69 & 8.5 & 0.70 & 0.63 & 0.67 & 0.70 & 8.3 \\*
 & UTSA & 0.50 & 0.15 & 0.31 & 0.59 & 7.6 & 0.65 & 0.65 & 0.65 & 0.65 & 6.6 \\
\midrule
\multirow{8}{*}{\textbf{TIRE}} & ActRecTut & 0.36 & 0.29 & 0.40 & 0.82 & 182 & 0.75 & 0.53 & 0.65 & 0.75 & 50.6 \\*
 & HAS & 0.56 & 0.13 & 0.44 & 0.65 & 44.3 & 0.63 & 0.55 & 0.61 & 0.68 & 41.1 \\*
 & MoCap & 0.50 & 0.15 & 0.43 & 0.59 & 11.1 & 0.52 & 0.42 & 0.51 & 0.62 & 51.8 \\*
 & PUMP & 0.21 & 0.12 & 0.25 & 0.57 & 55.0 & 0.67 & 0.53 & 0.61 & 0.67 & 49.2 \\*
 & SKAB & 0.64 & 0.17 & 0.58 & 0.73 & 4.9 & 0.70 & 0.63 & 0.71 & 0.79 & 9.2 \\*
 & TSSB & 0.41 & 0.05 & 0.27 & 0.58 & 6.5 & 0.60 & 0.52 & 0.55 & 0.65 & 16.7 \\*
 & USC-HAD & 0.46 & 0.28 & 0.37 & 0.66 & 68.1 & 0.61 & 0.31 & 0.53 & 0.67 & 71.4 \\*
 & UTSA & 0.37 & 0.02 & 0.24 & 0.62 & 93.2 & 0.68 & 0.68 & 0.73 & 0.75 & 71.5 \\
\midrule
\multirow{8}{*}{\textbf{Window}} & ActRecTut & 0.47 & 0.23 & 0.46 & 0.72 & 24.2 & 0.73 & 0.66 & 0.70 & 0.75 & 31.3 \\*
 & HAS & 0.40 & 0.22 & 0.46 & 0.67 & 1.7 & 0.69 & 0.66 & 0.71 & 0.75 & 1.2 \\*
 & MoCap & 0.41 & 0.24 & 0.40 & 0.54 & 2.4 & 0.42 & 0.40 & 0.45 & 0.49 & 1.2 \\*
 & PUMP & 0.61 & 0.22 & 0.54 & 0.64 & 25.2 & 0.63 & 0.58 & 0.63 & 0.65 & 0.7 \\*
 & SKAB & 0.41 & 0.18 & 0.56 & 0.73 & 0.2 & 0.68 & 0.64 & 0.71 & 0.76 & 0.2 \\*
 & TSSB & 0.42 & 0.41 & 0.43 & 0.53 & 0.2 & 0.63 & 0.57 & 0.63 & 0.70 & 1.8 \\*
 & USC-HAD & 0.42 & 0.26 & 0.56 & 0.73 & 36.7 & 0.44 & 0.30 & 0.58 & 0.71 & 69.3 \\*
 & UTSA & 0.59 & 0.36 & 0.56 & 0.69 & 3.9 & 0.71 & 0.62 & 0.67 & 0.71 & 4.5 \\
\midrule
\multirow{8}{*}{\textbf{iCID}} & ActRecTut & 0.65 & 0.32 & 0.57 & 0.71 & 742 & 0.68 & 0.32 & 0.56 & 0.75 & 697 \\*
 & HAS & 0.49 & 0.35 & 0.50 & 0.63 & 23.3 & 0.48 & 0.35 & 0.50 & 0.63 & 20.8 \\*
 & MoCap & 0.36 & 0.29 & 0.43 & 0.63 & 44.9 & 0.36 & 0.28 & 0.43 & 0.64 & 46.3 \\*
 & PUMP & 0.46 & 0.37 & 0.47 & 0.58 & 581 & 0.46 & 0.37 & 0.47 & 0.58 & 635 \\*
 & SKAB & 0.72 & 0.49 & 0.72 & 0.76 & 2.5 & 0.71 & 0.49 & 0.72 & 0.77 & 2.1 \\*
 & TSSB & 0.39 & 0.26 & 0.40 & 0.57 & 13.4 & 0.39 & 0.26 & 0.40 & 0.56 & 13.1 \\*
 & USC-HAD & 0.55 & 0.41 & 0.56 & 0.68 & 1.1k & 0.55 & 0.41 & 0.56 & 0.68 & 1.0k \\*
 & UTSA & 0.20 & 0.10 & 0.21 & 0.43 & 103 & 0.21 & 0.10 & 0.20 & 0.43 & 99.8 \\
\midrule
\multicolumn{12}{l}{\emph{SD methods, scored on the change points induced by their labels}} \\*
\midrule
\multirow{8}{*}{\textbf{AutoPlait}} & ActRecTut & 0.22 & 0.22 & 0.22 & 0.22 & 47.4 & 0.22 & 0.22 & 0.22 & 0.22 & 47.4 \\*
 & HAS & 0.65 & 0.65 & 0.65 & 0.65 & 10.4 & 0.65 & 0.65 & 0.65 & 0.65 & 10.4 \\*
 & MoCap & 0.88 & 0.88 & 0.88 & 0.88 & 25.0 & 0.88 & 0.88 & 0.88 & 0.88 & 25.0 \\*
 & PUMP & 0.27 & 0.27 & 0.27 & 0.27 & 316 & 0.27 & 0.27 & 0.27 & 0.27 & 316 \\*
 & SKAB & 0.71 & 0.71 & 0.71 & 0.71 & 1.5 & 0.71 & 0.71 & 0.71 & 0.71 & 1.5 \\*
 & TSSB & 0.42 & 0.42 & 0.42 & 0.42 & 0.7 & 0.42 & 0.42 & 0.42 & 0.42 & 0.7 \\*
 & USC-HAD & 0.43 & 0.43 & 0.43 & 0.43 & 155 & 0.43 & 0.43 & 0.43 & 0.43 & 155 \\*
 & UTSA & 0.61 & 0.61 & 0.61 & 0.61 & 2.2 & 0.61 & 0.61 & 0.61 & 0.61 & 2.2 \\
\midrule
\multirow{8}{*}{\textbf{E2USD}} & ActRecTut & 0.60 & 0.51 & 0.61 & 0.67 & 9.0 & 0.66 & 0.57 & 0.66 & 0.69 & 15.2 \\*
 & HAS & 0.48 & 0.33 & 0.39 & 0.48 & 12.7 & 0.72 & 0.61 & 0.67 & 0.72 & 12.7 \\*
 & MoCap & 0.76 & 0.29 & 0.59 & 0.76 & 14.3 & 0.71 & 0.50 & 0.65 & 0.73 & 10.2 \\*
 & PUMP & 0.50 & 0.48 & 0.54 & 0.59 & 18.1 & 0.50 & 0.48 & 0.54 & 0.60 & 12.2 \\*
 & SKAB & 0.47 & 0.30 & 0.41 & 0.47 & 15.6 & 0.64 & 0.64 & 0.67 & 0.70 & 15.7 \\*
 & TSSB & 0.36 & 0.21 & 0.29 & 0.37 & 16.5 & 0.54 & 0.41 & 0.50 & 0.58 & 15.3 \\*
 & USC-HAD & 0.69 & 0.45 & 0.59 & 0.69 & 14.8 & 0.68 & 0.46 & 0.59 & 0.68 & 15.1 \\*
 & UTSA & 0.28 & 0.12 & 0.21 & 0.28 & 14.1 & 0.61 & 0.34 & 0.48 & 0.64 & 12.7 \\
\midrule
\multirow{8}{*}{\textbf{HDP-HSMM}} & ActRecTut & 0.33 & 0.27 & 0.44 & 0.59 & 12.1 & 0.43 & 0.27 & 0.33 & 0.53 & 14.9 \\*
 & HAS & 0.64 & 0.60 & 0.65 & 0.67 & 7.4 & 0.73 & 0.67 & 0.73 & 0.74 & 4.0 \\*
 & MoCap & 0.56 & 0.51 & 0.59 & 0.71 & 6.3 & 0.63 & 0.51 & 0.60 & 0.68 & 7.9 \\*
 & PUMP & 0.12 & 0.11 & 0.11 & 0.13 & 12.7 & 0.12 & 0.11 & 0.11 & 0.13 & 25.9 \\*
 & SKAB & 0.70 & 0.63 & 0.69 & 0.75 & 6.2 & 0.78 & 0.67 & 0.73 & 0.78 & 4.1 \\*
 & TSSB & 0.57 & 0.47 & 0.55 & 0.60 & 15.1 & 0.61 & 0.54 & 0.60 & 0.62 & 4.4 \\*
 & USC-HAD & 0.37 & 0.35 & 0.36 & 0.42 & 11.0 & 0.37 & 0.36 & 0.36 & 0.41 & 9.9 \\*
 & UTSA & 0.61 & 0.51 & 0.63 & 0.69 & 11.9 & 0.64 & 0.52 & 0.64 & 0.68 & 5.2 \\
\midrule
\multirow{8}{*}{\textbf{TICC}} & ActRecTut & - & - & - & - & - & 0.72 & 0.40 & 0.69 & 0.75 & 93.0 \\*
 & HAS & - & - & - & - & - & 0.79 & 0.52 & 0.72 & 0.80 & 30.3 \\*
 & MoCap & - & - & - & - & - & 0.75 & 0.50 & 0.69 & 0.78 & 218 \\*
 & PUMP & - & - & - & - & - & 0.64 & 0.63 & 0.63 & 0.63 & 2.0k \\*
 & SKAB & - & - & - & - & - & 0.81 & 0.79 & 0.82 & 0.83 & 22.1 \\*
 & TSSB & - & - & - & - & - & 0.61 & 0.42 & 0.59 & 0.62 & 77.3 \\*
 & USC-HAD & - & - & - & - & - & 0.66 & 0.32 & 0.35 & 0.70 & 900 \\*
 & UTSA & - & - & - & - & - & 0.67 & 0.26 & 0.61 & 0.72 & 85.0 \\
\midrule
\multirow{8}{*}{\textbf{Time2State}} & ActRecTut & 0.65 & 0.54 & 0.60 & 0.66 & 50.4 & 0.65 & 0.57 & 0.63 & 0.67 & 4.7 \\*
 & HAS & 0.45 & 0.34 & 0.40 & 0.46 & 14.7 & 0.69 & 0.62 & 0.66 & 0.70 & 6.8 \\*
 & MoCap & 0.81 & 0.28 & 0.66 & 0.81 & 34.8 & 0.76 & 0.40 & 0.70 & 0.81 & 9.6 \\*
 & PUMP & 0.48 & 0.45 & 0.53 & 0.59 & 17.6 & 0.49 & 0.45 & 0.54 & 0.61 & 8.6 \\*
 & SKAB & 0.46 & 0.38 & 0.43 & 0.49 & 17.2 & 0.63 & 0.63 & 0.74 & 0.82 & 8.1 \\*
 & TSSB & 0.38 & 0.12 & 0.27 & 0.40 & 12.3 & 0.59 & 0.32 & 0.50 & 0.64 & 9.4 \\*
 & USC-HAD & 0.66 & 0.41 & 0.61 & 0.67 & 16.7 & 0.68 & 0.44 & 0.63 & 0.69 & 15.2 \\*
 & UTSA & 0.26 & 0.10 & 0.18 & 0.31 & 14.2 & 0.62 & 0.26 & 0.49 & 0.63 & 26.6 \\
\end{longtable}
\endgroup

\begingroup
\setlength{\tabcolsep}{3pt}\renewcommand{\arraystretch}{0.95}\small
\begin{longtable}{l l c c c c c c c c c c}
\caption{\textbf{Main-pool SD algorithms: per-(algorithm, dataset) default / worst-case / median / tuned SMS, in both Non-Guided (NG) and Guided (G) modes.} Conventions identical to Table~\ref{tab:main_pool_cpd}.} \label{tab:main_pool_sd} \\
\toprule
\multicolumn{2}{c}{} & \multicolumn{5}{c}{\textbf{Non-Guided}} & \multicolumn{5}{c}{\textbf{Guided}} \\ \cmidrule(lr){3-7} \cmidrule(lr){8-12}
 \textbf{Algo} & \textbf{Dataset}  & Def. & Worst & Med. & Tuned & $\tilde{t}$ [s]  & Def. & Worst & Med. & Tuned & $\tilde{t}$ [s] \\
\midrule
\endfirsthead
\multicolumn{12}{l}{\emph{(continued from previous page)}} \\
\toprule
\multicolumn{2}{c}{} & \multicolumn{5}{c}{\textbf{Non-Guided}} & \multicolumn{5}{c}{\textbf{Guided}} \\ \cmidrule(lr){3-7} \cmidrule(lr){8-12}
 \textbf{Algo} & \textbf{Dataset}  & Def. & Worst & Med. & Tuned & $\tilde{t}$ [s]  & Def. & Worst & Med. & Tuned & $\tilde{t}$ [s] \\
\midrule
\endhead
\midrule \multicolumn{12}{r}{\emph{(continued on next page)}} \\
\endfoot
\bottomrule
\endlastfoot
\multirow{8}{*}{\textbf{AutoPlait}} & ActRecTut & 0.31 & 0.31 & 0.31 & 0.31 & 47.4 & 0.31 & 0.31 & 0.31 & 0.31 & 47.4 \\*
 & HAS & 0.65 & 0.65 & 0.65 & 0.65 & 10.4 & 0.65 & 0.65 & 0.65 & 0.65 & 10.4 \\*
 & MoCap & 0.86 & 0.86 & 0.86 & 0.86 & 25.0 & 0.86 & 0.86 & 0.86 & 0.86 & 25.0 \\*
 & PUMP & 0.14 & 0.14 & 0.14 & 0.14 & 316 & 0.14 & 0.14 & 0.14 & 0.14 & 316 \\*
 & SKAB & 0.73 & 0.73 & 0.73 & 0.73 & 1.5 & 0.73 & 0.73 & 0.73 & 0.73 & 1.5 \\*
 & TSSB & 0.39 & 0.39 & 0.39 & 0.39 & 0.7 & 0.39 & 0.39 & 0.39 & 0.39 & 0.7 \\*
 & USC-HAD & 0.31 & 0.31 & 0.31 & 0.31 & 155 & 0.31 & 0.31 & 0.31 & 0.31 & 155 \\*
 & UTSA & 0.62 & 0.62 & 0.62 & 0.62 & 2.2 & 0.62 & 0.62 & 0.62 & 0.62 & 2.2 \\
\midrule
\multirow{8}{*}{\textbf{CLaP}} & ActRecTut & 0.40 & 0.53 & 0.57 & 0.59 & 623 & 0.54 & 0.50 & 0.53 & 0.57 & 647 \\*
 & HAS & 0.75 & 0.69 & 0.73 & 0.75 & 15.1 & 0.81 & 0.75 & 0.78 & 0.82 & 13.4 \\*
 & MoCap & 0.74 & 0.70 & 0.77 & 0.78 & 41.0 & 0.74 & 0.71 & 0.76 & 0.83 & 25.9 \\*
 & PUMP & 0.03 & -0.16 & -0.04 & 0.04 & 385 & 0.19 & -0.02 & 0.06 & 0.26 & 550 \\*
 & SKAB & 0.62 & 0.62 & 0.62 & 0.62 & 2.2 & 0.74 & 0.64 & 0.68 & 0.74 & 2.5 \\*
 & TSSB & 0.86 & 0.83 & 0.87 & 0.86 & 9.4 & 0.90 & 0.88 & 0.90 & 0.90 & 5.7 \\*
 & USC-HAD & 0.81 & 0.81 & 0.86 & 0.88 & 500 & 0.76 & 0.76 & 0.78 & 0.78 & 498 \\*
 & UTSA & 0.82 & 0.76 & 0.82 & 0.82 & 27.4 & 0.97 & 0.91 & 0.96 & 0.97 & 18.4 \\
\midrule
\multirow{8}{*}{\textbf{E2USD}} & ActRecTut & 0.68 & 0.60 & 0.71 & 0.78 & 12.5 & 0.78 & 0.77 & 0.79 & 0.84 & 15.2 \\*
 & HAS & 0.53 & 0.43 & 0.48 & 0.53 & 12.7 & 0.75 & 0.72 & 0.74 & 0.75 & 12.7 \\*
 & MoCap & 0.77 & 0.55 & 0.69 & 0.77 & 14.3 & 0.74 & 0.65 & 0.71 & 0.75 & 10.2 \\*
 & PUMP & 0.30 & 0.27 & 0.33 & 0.37 & 18.1 & 0.32 & 0.28 & 0.35 & 0.40 & 12.2 \\*
 & SKAB & 0.48 & 0.35 & 0.40 & 0.48 & 15.6 & 0.72 & 0.71 & 0.74 & 0.75 & 16.0 \\*
 & TSSB & 0.43 & 0.37 & 0.40 & 0.43 & 18.1 & 0.57 & 0.56 & 0.59 & 0.61 & 14.2 \\*
 & USC-HAD & 0.70 & 0.57 & 0.64 & 0.70 & 16.4 & 0.64 & 0.53 & 0.59 & 0.66 & 14.9 \\*
 & UTSA & 0.44 & 0.35 & 0.41 & 0.46 & 13.3 & 0.69 & 0.63 & 0.68 & 0.72 & 12.7 \\
\midrule
\multirow{8}{*}{\textbf{HDP-HSMM}} & ActRecTut & 0.41 & 0.34 & 0.48 & 0.63 & 6.2 & 0.54 & 0.41 & 0.47 & 0.54 & 15.1 \\*
 & HAS & 0.65 & 0.64 & 0.66 & 0.68 & 11.3 & 0.73 & 0.69 & 0.72 & 0.73 & 4.0 \\*
 & MoCap & 0.58 & 0.53 & 0.62 & 0.69 & 6.3 & 0.61 & 0.54 & 0.60 & 0.68 & 10.6 \\*
 & PUMP & -0.05 & -0.07 & -0.05 & -0.04 & 12.7 & -0.04 & -0.05 & -0.04 & -0.04 & 25.2 \\*
 & SKAB & 0.67 & 0.63 & 0.66 & 0.70 & 6.2 & 0.81 & 0.71 & 0.77 & 0.81 & 4.1 \\*
 & TSSB & 0.56 & 0.48 & 0.53 & 0.59 & 16.7 & 0.57 & 0.50 & 0.55 & 0.57 & 4.8 \\*
 & USC-HAD & 0.31 & 0.26 & 0.31 & 0.33 & 23.2 & 0.30 & 0.26 & 0.30 & 0.31 & 9.0 \\*
 & UTSA & 0.63 & 0.63 & 0.67 & 0.72 & 11.4 & 0.66 & 0.59 & 0.65 & 0.72 & 5.5 \\
\midrule
\multirow{8}{*}{\textbf{TICC}} & ActRecTut & - & - & - & - & - & 0.82 & 0.70 & 0.80 & 0.86 & 1.0k \\*
 & HAS & - & - & - & - & - & 0.79 & 0.70 & 0.77 & 0.80 & 29.3 \\*
 & MoCap & - & - & - & - & - & 0.81 & 0.64 & 0.73 & 0.80 & 218 \\*
 & PUMP & - & - & - & - & - & 0.56 & 0.54 & 0.54 & 0.54 & 1.8k \\*
 & SKAB & - & - & - & - & - & 0.83 & 0.82 & 0.85 & 0.85 & 22.1 \\*
 & TSSB & - & - & - & - & - & 0.44 & 0.44 & 0.46 & 0.55 & 22.4 \\*
 & USC-HAD & - & - & - & - & - & 0.62 & 0.48 & 0.52 & 0.64 & 716 \\*
 & UTSA & - & - & - & - & - & 0.61 & 0.56 & 0.61 & 0.65 & 12.1 \\
\midrule
\multirow{8}{*}{\textbf{Time2State}} & ActRecTut & 0.69 & 0.65 & 0.72 & 0.78 & 34.7 & 0.72 & 0.71 & 0.75 & 0.80 & 12.7 \\*
 & HAS & 0.47 & 0.41 & 0.45 & 0.51 & 14.3 & 0.73 & 0.73 & 0.74 & 0.75 & 6.8 \\*
 & MoCap & 0.84 & 0.56 & 0.75 & 0.84 & 15.7 & 0.79 & 0.59 & 0.78 & 0.85 & 9.6 \\*
 & PUMP & 0.29 & 0.25 & 0.34 & 0.41 & 17.6 & 0.32 & 0.26 & 0.37 & 0.47 & 8.6 \\*
 & SKAB & 0.44 & 0.32 & 0.42 & 0.49 & 12.5 & 0.67 & 0.67 & 0.80 & 0.86 & 8.1 \\*
 & TSSB & 0.46 & 0.26 & 0.39 & 0.47 & 14.4 & 0.61 & 0.50 & 0.60 & 0.68 & 10.2 \\*
 & USC-HAD & 0.71 & 0.61 & 0.68 & 0.72 & 17.2 & 0.69 & 0.60 & 0.66 & 0.70 & 14.7 \\*
 & UTSA & 0.42 & 0.29 & 0.39 & 0.48 & 13.2 & 0.75 & 0.59 & 0.70 & 0.76 & 26.6 \\
\end{longtable}
\endgroup

\section{Cross-metric Consistency Check}
\label{app:secondary_metrics}

The per-(algorithm, dataset) tables of Appendices~\ref{app:padding_strategy} and~\ref{app:main_pool_full_grid} use the primary metric of each task (BiC for CPD, SMS for SD). For transparency, we re-emit the same tables under the alternative metrics that the segmentation literature most commonly reports: \emph{Gaussian} $F_1$ for CPD (point-process matching with Gaussian tolerance), and \emph{ARI} / \emph{WARI} for SD (label-permutation invariant clustering metrics). Each table is self-contained: its \emph{Tuned} column is the configuration maximising the table's own metric, not a re-evaluation of the BiC/SMS-best configuration. The leaders do not depend on the metric. In CPD, the top five methods of each mode are the same under BiC, Covering, $F_1$ and Gaussian $F_1$: ClaSP and the penalised detectors without $K$; DynP, BinSeg, BottomUp, GGS and ClaSP with $K$, within $0.02$ of each other under the $F_1$-type metrics, which favour ClaSP and GGS. In SD, CLaP leads in both modes under every metric, with Time2State second in Guided mode. Lower ranks can swap (e.g.\ HDP-HSMM and E2USD under WARI): the conclusions of the main paper are not artefacts of the primary metric, but fine-grained orderings are metric-dependent.

\paragraph{Main pool.}
\begingroup
\setlength{\tabcolsep}{3pt}\renewcommand{\arraystretch}{0.95}\small
\begin{longtable}{l l c c c c c c c c c c}
\caption{\textbf{Main-pool algorithms on CPD: per-(algorithm, dataset) default / worst-case / median / tuned Gaussian $F_1$, in both Non-Guided (NG) and Guided (G) modes.} Same conventions as Table~\ref{tab:main_pool_cpd}, with BiC replaced by Gaussian $F_1$. The tuned configuration in this table is selected on Gaussian $F_1$ itself (not on BiC), so each table is self-contained.} \label{tab:main_pool_cpd_gf1} \\
\toprule
\multicolumn{2}{c}{} & \multicolumn{5}{c}{\textbf{Non-Guided}} & \multicolumn{5}{c}{\textbf{Guided}} \\ \cmidrule(lr){3-7} \cmidrule(lr){8-12}
 \textbf{Algo} & \textbf{Dataset}  & Def. & Worst & Med. & Tuned & $\tilde{t}$ [s]  & Def. & Worst & Med. & Tuned & $\tilde{t}$ [s] \\
\midrule
\endfirsthead
\multicolumn{12}{l}{\emph{(continued from previous page)}} \\
\toprule
\multicolumn{2}{c}{} & \multicolumn{5}{c}{\textbf{Non-Guided}} & \multicolumn{5}{c}{\textbf{Guided}} \\ \cmidrule(lr){3-7} \cmidrule(lr){8-12}
 \textbf{Algo} & \textbf{Dataset}  & Def. & Worst & Med. & Tuned & $\tilde{t}$ [s]  & Def. & Worst & Med. & Tuned & $\tilde{t}$ [s] \\
\midrule
\endhead
\midrule \multicolumn{12}{r}{\emph{(continued on next page)}} \\
\endfoot
\bottomrule
\endlastfoot
\multirow{8}{*}{\textbf{AMOC}} & ActRecTut & 0.05 & 0.05 & 0.05 & 0.05 & 0.2 & 0.05 & 0.05 & 0.05 & 0.05 & 0.2 \\*
 & HAS & 0.19 & 0.19 & 0.19 & 0.19 & 0.2 & 0.19 & 0.19 & 0.19 & 0.19 & 0.2 \\*
 & MoCap & 0.17 & 0.17 & 0.17 & 0.17 & 0.2 & 0.17 & 0.17 & 0.17 & 0.17 & 0.2 \\*
 & PUMP & 0.03 & 0.03 & 0.03 & 0.03 & 0.2 & 0.03 & 0.03 & 0.03 & 0.03 & 0.2 \\*
 & SKAB & 0.12 & 0.12 & 0.12 & 0.12 & 0.3 & 0.12 & 0.12 & 0.12 & 0.12 & 0.3 \\*
 & TSSB & 0.00 & 0.00 & 0.00 & 0.00 & 0.3 & 0.00 & 0.00 & 0.00 & 0.00 & 0.3 \\*
 & USC-HAD & 0.17 & 0.17 & 0.17 & 0.17 & 0.3 & 0.17 & 0.17 & 0.17 & 0.17 & 0.3 \\*
 & UTSA & 0.09 & 0.09 & 0.09 & 0.09 & 0.3 & 0.09 & 0.09 & 0.09 & 0.09 & 0.3 \\
\midrule
\multirow{8}{*}{\textbf{BinSeg}} & ActRecTut & 0.13 & 0.09 & 0.26 & 0.65 & 32.0 & 0.39 & 0.39 & 0.54 & 0.62 & 30.3 \\*
 & HAS & 0.33 & 0.12 & 0.25 & 0.42 & 1.1 & 0.46 & 0.46 & 0.50 & 0.54 & 1.1 \\*
 & MoCap & 0.47 & 0.10 & 0.22 & 0.61 & 2.0 & 0.51 & 0.51 & 0.57 & 0.69 & 1.7 \\*
 & PUMP & 0.47 & 0.24 & 0.58 & 0.68 & 22.6 & 0.67 & 0.66 & 0.67 & 0.68 & 19.2 \\*
 & SKAB & 0.18 & 0.00 & 0.11 & 0.20 & 0.2 & 0.18 & 0.17 & 0.17 & 0.20 & 0.3 \\*
 & TSSB & 0.07 & 0.02 & 0.06 & 0.14 & 0.6 & 0.08 & 0.08 & 0.20 & 0.35 & 1.1 \\*
 & USC-HAD & 0.79 & 0.16 & 0.65 & 0.82 & 43.7 & 0.80 & 0.79 & 0.82 & 0.87 & 53.2 \\*
 & UTSA & 0.10 & 0.02 & 0.02 & 0.24 & 3.3 & 0.10 & 0.10 & 0.17 & 0.34 & 2.7 \\
\midrule
\multirow{8}{*}{\textbf{BottomUp}} & ActRecTut & 0.32 & 0.09 & 0.43 & 0.63 & 7.0 & 0.54 & 0.54 & 0.58 & 0.65 & 8.7 \\*
 & HAS & 0.30 & 0.09 & 0.22 & 0.44 & 1.6 & 0.45 & 0.45 & 0.50 & 0.55 & 1.7 \\*
 & MoCap & 0.35 & 0.10 & 0.22 & 0.57 & 1.7 & 0.41 & 0.41 & 0.49 & 0.65 & 1.5 \\*
 & PUMP & 0.50 & 0.28 & 0.61 & 0.69 & 3.5 & 0.69 & 0.68 & 0.69 & 0.69 & 8.2 \\*
 & SKAB & 0.22 & 0.00 & 0.12 & 0.23 & 0.2 & 0.22 & 0.17 & 0.21 & 0.22 & 0.3 \\*
 & TSSB & 0.06 & 0.03 & 0.06 & 0.12 & 1.5 & 0.13 & 0.13 & 0.15 & 0.23 & 1.8 \\*
 & USC-HAD & 0.65 & 0.03 & 0.52 & 0.81 & 40.8 & 0.68 & 0.68 & 0.76 & 0.85 & 44.7 \\*
 & UTSA & 0.07 & 0.01 & 0.01 & 0.19 & 2.1 & 0.12 & 0.12 & 0.12 & 0.32 & 9.8 \\
\midrule
\multirow{8}{*}{\textbf{ChangeFinder}} & ActRecTut & 0.27 & 0.00 & 0.32 & 0.47 & 80.5 & 0.44 & 0.05 & 0.43 & 0.48 & 74.1 \\*
 & HAS & 0.25 & 0.12 & 0.18 & 0.38 & 10.5 & 0.36 & 0.28 & 0.35 & 0.43 & 10.0 \\*
 & MoCap & 0.20 & 0.00 & 0.15 & 0.47 & 7.3 & 0.36 & 0.13 & 0.24 & 0.37 & 1.9 \\*
 & PUMP & 0.32 & 0.00 & 0.37 & 0.59 & 65.0 & 0.60 & 0.53 & 0.60 & 0.65 & 7.5 \\*
 & SKAB & 0.05 & 0.00 & 0.02 & 0.21 & 2.8 & 0.10 & 0.01 & 0.04 & 0.13 & 0.5 \\*
 & TSSB & 0.15 & 0.08 & 0.08 & 0.19 & 1.4 & 0.24 & 0.14 & 0.23 & 0.26 & 1.3 \\*
 & USC-HAD & 0.51 & 0.09 & 0.45 & 0.57 & 55.1 & 0.53 & 0.41 & 0.51 & 0.54 & 11.1 \\*
 & UTSA & 0.18 & 0.00 & 0.10 & 0.22 & 3.5 & 0.21 & 0.05 & 0.20 & 0.23 & 3.9 \\
\midrule
\multirow{8}{*}{\textbf{ClaSP}} & ActRecTut & 0.19 & 0.08 & 0.15 & 0.21 & 653 & 0.38 & 0.28 & 0.30 & 0.38 & 601 \\*
 & HAS & 0.45 & 0.28 & 0.32 & 0.45 & 10.6 & 0.50 & 0.37 & 0.42 & 0.50 & 6.8 \\*
 & MoCap & 0.56 & 0.55 & 0.58 & 0.62 & 14.5 & 0.58 & 0.55 & 0.58 & 0.60 & 47.4 \\*
 & PUMP & 0.18 & 0.13 & 0.26 & 0.57 & 607 & 0.54 & 0.28 & 0.42 & 0.59 & 603 \\*
 & SKAB & 0.00 & 0.00 & 0.05 & 0.14 & 1.8 & 0.15 & 0.08 & 0.09 & 0.16 & 0.6 \\*
 & TSSB & 0.63 & 0.61 & 0.63 & 0.68 & 2.5 & 0.69 & 0.67 & 0.72 & 0.73 & 1.0 \\*
 & USC-HAD & 0.68 & 0.49 & 0.61 & 0.80 & 448 & 0.74 & 0.67 & 0.74 & 0.81 & 469 \\*
 & UTSA & 0.44 & 0.28 & 0.36 & 0.47 & 13.8 & 0.69 & 0.57 & 0.64 & 0.69 & 7.1 \\
\midrule
\multirow{8}{*}{\textbf{DynP}} & ActRecTut & - & - & - & - & - & 0.48 & 0.48 & 0.55 & 0.61 & 21.3 \\*
 & HAS & - & - & - & - & - & 0.46 & 0.46 & 0.51 & 0.55 & 0.8 \\*
 & MoCap & - & - & - & - & - & 0.52 & 0.52 & 0.58 & 0.73 & 1.1 \\*
 & PUMP & - & - & - & - & - & 0.68 & 0.67 & 0.68 & 0.69 & 28.8 \\*
 & SKAB & - & - & - & - & - & 0.17 & 0.17 & 0.18 & 0.21 & 0.3 \\*
 & TSSB & - & - & - & - & - & 0.10 & 0.10 & 0.19 & 0.33 & 0.5 \\*
 & USC-HAD & - & - & - & - & - & 0.77 & 0.77 & 0.81 & 0.87 & 31.3 \\*
 & UTSA & - & - & - & - & - & 0.13 & 0.12 & 0.16 & 0.37 & 1.9 \\
\midrule
\multirow{8}{*}{\textbf{ESPRESSO}} & ActRecTut & - & - & - & - & - & 0.50 & 0.50 & 0.52 & 0.53 & 5.1k \\*
 & HAS & - & - & - & - & - & 0.47 & 0.47 & 0.47 & 0.47 & 50.9 \\*
 & MoCap & - & - & - & - & - & 0.55 & 0.52 & 0.56 & 0.60 & 106 \\*
 & PUMP & - & - & - & - & - & 0.64 & 0.61 & 0.63 & 0.65 & 2.1k \\*
 & SKAB & - & - & - & - & - & 0.24 & 0.19 & 0.23 & 0.25 & 6.1 \\*
 & TSSB & - & - & - & - & - & 0.12 & 0.11 & 0.11 & 0.13 & 5.7 \\*
 & USC-HAD & - & - & - & - & - & 0.74 & 0.72 & 0.73 & 0.74 & 3.0k \\*
 & UTSA & - & - & - & - & - & 0.13 & 0.08 & 0.11 & 0.13 & 36.4 \\
\midrule
\multirow{8}{*}{\textbf{FLUSS}} & ActRecTut & 0.04 & 0.00 & 0.05 & 0.40 & 43.2 & 0.22 & 0.07 & 0.16 & 0.46 & 10.6 \\*
 & HAS & 0.12 & 0.07 & 0.13 & 0.24 & 13.0 & 0.29 & 0.24 & 0.28 & 0.36 & 12.0 \\*
 & MoCap & 0.00 & 0.00 & 0.00 & 0.31 & 6.2 & 0.23 & 0.15 & 0.22 & 0.31 & 10.5 \\*
 & PUMP & 0.02 & 0.00 & 0.02 & 0.54 & 24.5 & 0.15 & 0.08 & 0.19 & 0.49 & 4.3 \\*
 & SKAB & 0.10 & 0.00 & 0.00 & 0.15 & 10.8 & 0.10 & 0.02 & 0.02 & 0.17 & 11.1 \\*
 & TSSB & 0.21 & 0.14 & 0.21 & 0.35 & 1.6 & 0.47 & 0.40 & 0.42 & 0.51 & 1.9 \\*
 & USC-HAD & 0.06 & 0.02 & 0.10 & 0.31 & 4.6 & 0.22 & 0.09 & 0.26 & 0.49 & 21.4 \\*
 & UTSA & 0.05 & 0.02 & 0.06 & 0.27 & 2.2 & 0.58 & 0.55 & 0.55 & 0.63 & 2.1 \\
\midrule
\multirow{8}{*}{\textbf{GGS}} & ActRecTut & 0.32 & 0.19 & 0.32 & 0.54 & 165 & 0.50 & 0.49 & 0.50 & 0.59 & 270 \\*
 & HAS & 0.30 & 0.20 & 0.29 & 0.37 & 8.4 & 0.58 & 0.58 & 0.58 & 0.59 & 3.6 \\*
 & MoCap & 0.70 & 0.46 & 0.60 & 0.72 & 18.1 & 0.79 & 0.78 & 0.79 & 0.81 & 12.7 \\*
 & PUMP & 0.25 & 0.14 & 0.26 & 0.43 & 164 & 0.73 & 0.61 & 0.72 & 0.75 & 350 \\*
 & SKAB & 0.16 & 0.11 & 0.17 & 0.22 & 1.9 & 0.27 & 0.24 & 0.27 & 0.27 & 1.0 \\*
 & TSSB & 0.03 & 0.02 & 0.03 & 0.07 & 3.1 & 0.13 & 0.12 & 0.13 & 0.15 & 1.9 \\*
 & USC-HAD & 0.69 & 0.58 & 0.62 & 0.78 & 66.3 & 0.70 & 0.65 & 0.70 & 0.79 & 87.6 \\*
 & UTSA & 0.09 & 0.06 & 0.09 & 0.14 & 11.1 & 0.24 & 0.24 & 0.24 & 0.24 & 1.9 \\
\midrule
\multirow{8}{*}{\textbf{IGTS}} & ActRecTut & 0.17 & 0.14 & 0.17 & 0.35 & 193 & 0.39 & 0.38 & 0.39 & 0.39 & 75.3 \\*
 & HAS & 0.24 & 0.15 & 0.22 & 0.29 & 1.5 & 0.47 & 0.47 & 0.47 & 0.47 & 0.8 \\*
 & MoCap & 0.45 & 0.33 & 0.44 & 0.49 & 0.4 & 0.50 & 0.50 & 0.50 & 0.50 & 1.4 \\*
 & PUMP & 0.24 & 0.15 & 0.24 & 0.48 & 26.2 & 0.69 & 0.69 & 0.69 & 0.69 & 62.2 \\*
 & SKAB & 0.15 & 0.10 & 0.14 & 0.20 & 0.4 & 0.22 & 0.22 & 0.22 & 0.24 & 0.6 \\*
 & TSSB & 0.00 & 0.00 & 0.00 & 0.01 & 0.5 & 0.08 & 0.08 & 0.08 & 0.08 & 0.5 \\*
 & USC-HAD & 0.78 & 0.48 & 0.59 & 0.79 & 3.9 & 0.76 & 0.75 & 0.76 & 0.77 & 5.1 \\*
 & UTSA & 0.08 & 0.04 & 0.07 & 0.11 & 1.3 & 0.10 & 0.10 & 0.10 & 0.10 & 1.0 \\
\midrule
\multirow{8}{*}{\textbf{PELT}} & ActRecTut & 0.24 & 0.09 & 0.33 & 0.67 & 1.1 & 0.24 & 0.09 & 0.33 & 0.67 & 1.1 \\*
 & HAS & 0.32 & 0.10 & 0.24 & 0.43 & 0.6 & 0.32 & 0.10 & 0.24 & 0.43 & 0.6 \\*
 & MoCap & 0.37 & 0.11 & 0.22 & 0.65 & 0.6 & 0.37 & 0.11 & 0.22 & 0.65 & 0.6 \\*
 & PUMP & 0.49 & 0.26 & 0.60 & 0.69 & 0.6 & 0.49 & 0.26 & 0.60 & 0.69 & 0.6 \\*
 & SKAB & 0.16 & 0.00 & 0.12 & 0.21 & 0.1 & 0.16 & 0.00 & 0.12 & 0.21 & 0.1 \\*
 & TSSB & 0.05 & 0.03 & 0.06 & 0.14 & 0.4 & 0.05 & 0.03 & 0.06 & 0.14 & 0.4 \\*
 & USC-HAD & 0.77 & 0.03 & 0.59 & 0.82 & 12.0 & 0.77 & 0.03 & 0.59 & 0.82 & 12.0 \\*
 & UTSA & 0.08 & 0.01 & 0.01 & 0.22 & 1.7 & 0.08 & 0.01 & 0.01 & 0.22 & 1.7 \\
\midrule
\multirow{8}{*}{\textbf{Prophet}} & ActRecTut & 0.39 & 0.27 & 0.39 & 0.45 & 71.6 & 0.42 & 0.42 & 0.42 & 0.43 & 71.1 \\*
 & HAS & 0.16 & 0.15 & 0.16 & 0.20 & 42.1 & 0.28 & 0.28 & 0.29 & 0.30 & 45.0 \\*
 & MoCap & 0.27 & 0.05 & 0.24 & 0.31 & 8.6 & 0.22 & 0.10 & 0.16 & 0.26 & 7.3 \\*
 & PUMP & 0.23 & 0.18 & 0.30 & 0.43 & 75.0 & 0.54 & 0.54 & 0.54 & 0.54 & 70.1 \\*
 & SKAB & 0.12 & 0.07 & 0.08 & 0.14 & 39.7 & 0.12 & 0.12 & 0.14 & 0.22 & 39.3 \\*
 & TSSB & 0.13 & 0.11 & 0.11 & 0.13 & 2.9 & 0.20 & 0.20 & 0.20 & 0.20 & 1.8 \\*
 & USC-HAD & 0.63 & 0.42 & 0.54 & 0.63 & 8.5 & 0.65 & 0.50 & 0.59 & 0.65 & 8.8 \\*
 & UTSA & 0.07 & 0.06 & 0.06 & 0.09 & 6.5 & 0.09 & 0.09 & 0.09 & 0.09 & 6.6 \\
\midrule
\multirow{8}{*}{\textbf{TIRE}} & ActRecTut & 0.15 & 0.09 & 0.20 & 0.68 & 184 & 0.60 & 0.45 & 0.56 & 0.70 & 48.6 \\*
 & HAS & 0.24 & 0.05 & 0.16 & 0.31 & 22.1 & 0.26 & 0.14 & 0.22 & 0.34 & 41.1 \\*
 & MoCap & 0.27 & 0.12 & 0.22 & 0.35 & 11.1 & 0.25 & 0.13 & 0.25 & 0.36 & 12.6 \\*
 & PUMP & 0.21 & 0.07 & 0.25 & 0.57 & 54.3 & 0.73 & 0.59 & 0.67 & 0.73 & 49.2 \\*
 & SKAB & 0.17 & 0.07 & 0.10 & 0.23 & 2.5 & 0.17 & 0.08 & 0.12 & 0.23 & 8.8 \\*
 & TSSB & 0.17 & 0.04 & 0.09 & 0.25 & 27.6 & 0.16 & 0.09 & 0.09 & 0.26 & 38.6 \\*
 & USC-HAD & 0.39 & 0.04 & 0.26 & 0.59 & 178 & 0.50 & 0.22 & 0.41 & 0.65 & 156 \\*
 & UTSA & 0.09 & 0.01 & 0.03 & 0.14 & 46.7 & 0.08 & 0.05 & 0.08 & 0.17 & 40.5 \\
\midrule
\multirow{8}{*}{\textbf{Window}} & ActRecTut & 0.20 & 0.05 & 0.24 & 0.66 & 60.5 & 0.63 & 0.56 & 0.62 & 0.70 & 31.3 \\*
 & HAS & 0.18 & 0.10 & 0.20 & 0.34 & 1.2 & 0.33 & 0.30 & 0.36 & 0.44 & 1.2 \\*
 & MoCap & 0.15 & 0.08 & 0.18 & 0.30 & 0.2 & 0.12 & 0.12 & 0.16 & 0.23 & 0.3 \\*
 & PUMP & 0.65 & 0.23 & 0.58 & 0.68 & 52.6 & 0.66 & 0.63 & 0.66 & 0.69 & 0.5 \\*
 & SKAB & 0.11 & 0.08 & 0.13 & 0.18 & 0.2 & 0.20 & 0.16 & 0.17 & 0.20 & 0.2 \\*
 & TSSB & 0.05 & 0.02 & 0.07 & 0.21 & 0.9 & 0.21 & 0.15 & 0.16 & 0.32 & 1.8 \\*
 & USC-HAD & 0.28 & 0.13 & 0.44 & 0.67 & 64.1 & 0.30 & 0.18 & 0.46 & 0.67 & 69.3 \\*
 & UTSA & 0.04 & 0.01 & 0.02 & 0.13 & 0.2 & 0.15 & 0.03 & 0.02 & 0.15 & 0.4 \\
\midrule
\multirow{8}{*}{\textbf{iCID}} & ActRecTut & 0.47 & 0.15 & 0.48 & 0.67 & 708 & 0.49 & 0.14 & 0.49 & 0.64 & 704 \\*
 & HAS & 0.22 & 0.13 & 0.20 & 0.30 & 20.7 & 0.21 & 0.13 & 0.20 & 0.31 & 21.9 \\*
 & MoCap & 0.25 & 0.16 & 0.25 & 0.47 & 44.9 & 0.24 & 0.16 & 0.24 & 0.48 & 46.3 \\*
 & PUMP & 0.47 & 0.35 & 0.47 & 0.59 & 574 & 0.47 & 0.35 & 0.47 & 0.59 & 624 \\*
 & SKAB & 0.06 & 0.04 & 0.05 & 0.17 & 1.9 & 0.05 & 0.04 & 0.05 & 0.16 & 2.6 \\*
 & TSSB & 0.12 & 0.09 & 0.11 & 0.17 & 13.6 & 0.11 & 0.09 & 0.11 & 0.16 & 12.2 \\*
 & USC-HAD & 0.35 & 0.13 & 0.36 & 0.61 & 1.1k & 0.35 & 0.12 & 0.36 & 0.61 & 1.0k \\*
 & UTSA & 0.05 & 0.03 & 0.05 & 0.12 & 96.6 & 0.05 & 0.03 & 0.05 & 0.13 & 98.9 \\
\midrule
\multicolumn{12}{l}{\emph{SD methods, scored on the change points induced by their labels}} \\*
\midrule
\multirow{8}{*}{\textbf{AutoPlait}} & ActRecTut & 0.05 & 0.05 & 0.05 & 0.05 & 47.4 & 0.05 & 0.05 & 0.05 & 0.05 & 47.4 \\*
 & HAS & 0.33 & 0.33 & 0.33 & 0.33 & 10.4 & 0.33 & 0.33 & 0.33 & 0.33 & 10.4 \\*
 & MoCap & 0.75 & 0.75 & 0.75 & 0.75 & 25.0 & 0.75 & 0.75 & 0.75 & 0.75 & 25.0 \\*
 & PUMP & 0.19 & 0.19 & 0.19 & 0.19 & 316 & 0.19 & 0.19 & 0.19 & 0.19 & 316 \\*
 & SKAB & 0.18 & 0.18 & 0.18 & 0.18 & 1.5 & 0.18 & 0.18 & 0.18 & 0.18 & 1.5 \\*
 & TSSB & 0.08 & 0.08 & 0.08 & 0.08 & 0.7 & 0.08 & 0.08 & 0.08 & 0.08 & 0.7 \\*
 & USC-HAD & 0.40 & 0.40 & 0.40 & 0.40 & 155 & 0.40 & 0.40 & 0.40 & 0.40 & 155 \\*
 & UTSA & 0.17 & 0.17 & 0.17 & 0.17 & 2.2 & 0.17 & 0.17 & 0.17 & 0.17 & 2.2 \\
\midrule
\multirow{8}{*}{\textbf{E2USD}} & ActRecTut & 0.53 & 0.49 & 0.54 & 0.58 & 10.0 & 0.53 & 0.52 & 0.54 & 0.59 & 10.0 \\*
 & HAS & 0.17 & 0.10 & 0.14 & 0.18 & 12.1 & 0.33 & 0.31 & 0.32 & 0.36 & 13.4 \\*
 & MoCap & 0.49 & 0.17 & 0.31 & 0.49 & 14.3 & 0.47 & 0.29 & 0.38 & 0.47 & 12.3 \\*
 & PUMP & 0.60 & 0.58 & 0.63 & 0.66 & 17.8 & 0.61 & 0.59 & 0.63 & 0.67 & 12.8 \\*
 & SKAB & 0.04 & 0.02 & 0.05 & 0.14 & 15.1 & 0.03 & 0.01 & 0.03 & 0.12 & 16.0 \\*
 & TSSB & 0.08 & 0.08 & 0.09 & 0.13 & 17.6 & 0.18 & 0.17 & 0.18 & 0.23 & 14.4 \\*
 & USC-HAD & 0.44 & 0.10 & 0.27 & 0.44 & 14.8 & 0.45 & 0.11 & 0.28 & 0.45 & 15.1 \\*
 & UTSA & 0.06 & 0.02 & 0.05 & 0.07 & 17.7 & 0.15 & 0.06 & 0.09 & 0.16 & 12.7 \\
\midrule
\multirow{8}{*}{\textbf{HDP-HSMM}} & ActRecTut & 0.10 & 0.07 & 0.23 & 0.40 & 6.2 & 0.20 & 0.07 & 0.12 & 0.33 & 14.9 \\*
 & HAS & 0.30 & 0.27 & 0.30 & 0.32 & 18.7 & 0.43 & 0.37 & 0.42 & 0.44 & 3.9 \\*
 & MoCap & 0.45 & 0.43 & 0.47 & 0.56 & 6.3 & 0.53 & 0.39 & 0.49 & 0.57 & 11.0 \\*
 & PUMP & 0.08 & 0.07 & 0.08 & 0.10 & 12.7 & 0.08 & 0.08 & 0.08 & 0.09 & 25.2 \\*
 & SKAB & 0.15 & 0.09 & 0.12 & 0.20 & 5.7 & 0.18 & 0.10 & 0.15 & 0.18 & 4.8 \\*
 & TSSB & 0.08 & 0.06 & 0.04 & 0.16 & 15.1 & 0.17 & 0.15 & 0.13 & 0.19 & 4.8 \\*
 & USC-HAD & 0.33 & 0.32 & 0.32 & 0.38 & 11.0 & 0.33 & 0.32 & 0.32 & 0.37 & 9.9 \\*
 & UTSA & 0.19 & 0.13 & 0.17 & 0.25 & 19.7 & 0.19 & 0.11 & 0.17 & 0.22 & 5.5 \\
\midrule
\multirow{8}{*}{\textbf{TICC}} & ActRecTut & - & - & - & - & - & 0.48 & 0.12 & 0.48 & 0.57 & 134 \\*
 & HAS & - & - & - & - & - & 0.54 & 0.28 & 0.45 & 0.53 & 39.1 \\*
 & MoCap & - & - & - & - & - & 0.58 & 0.27 & 0.49 & 0.60 & 218 \\*
 & PUMP & - & - & - & - & - & 0.68 & 0.69 & 0.69 & 0.69 & 1.8k \\*
 & SKAB & - & - & - & - & - & 0.27 & 0.25 & 0.28 & 0.31 & 66.3 \\*
 & TSSB & - & - & - & - & - & 0.15 & 0.14 & 0.14 & 0.24 & 22.4 \\*
 & USC-HAD & - & - & - & - & - & 0.51 & 0.02 & 0.03 & 0.42 & 900 \\*
 & UTSA & - & - & - & - & - & 0.12 & 0.05 & 0.07 & 0.28 & 12.4 \\
\midrule
\multirow{8}{*}{\textbf{Time2State}} & ActRecTut & 0.63 & 0.44 & 0.54 & 0.63 & 75.1 & 0.58 & 0.47 & 0.57 & 0.64 & 26.8 \\*
 & HAS & 0.16 & 0.11 & 0.13 & 0.18 & 15.9 & 0.30 & 0.30 & 0.31 & 0.35 & 6.7 \\*
 & MoCap & 0.53 & 0.18 & 0.36 & 0.54 & 15.7 & 0.50 & 0.23 & 0.43 & 0.54 & 9.6 \\*
 & PUMP & 0.55 & 0.53 & 0.61 & 0.66 & 20.7 & 0.56 & 0.53 & 0.62 & 0.67 & 8.6 \\*
 & SKAB & 0.04 & 0.01 & 0.05 & 0.14 & 12.6 & 0.05 & 0.01 & 0.03 & 0.26 & 8.0 \\*
 & TSSB & 0.11 & 0.06 & 0.09 & 0.12 & 12.8 & 0.18 & 0.17 & 0.17 & 0.24 & 9.7 \\*
 & USC-HAD & 0.41 & 0.07 & 0.30 & 0.43 & 15.4 & 0.46 & 0.08 & 0.34 & 0.47 & 15.2 \\*
 & UTSA & 0.06 & 0.02 & 0.04 & 0.08 & 14.2 & 0.13 & 0.05 & 0.08 & 0.14 & 11.9 \\
\end{longtable}
\endgroup

\begingroup
\setlength{\tabcolsep}{3pt}\renewcommand{\arraystretch}{0.95}\small
\begin{longtable}{l l c c c c c c c c c c}
\caption{\textbf{Main-pool SD algorithms: per-(algorithm, dataset) default / worst-case / median / tuned ARI, in both Non-Guided (NG) and Guided (G) modes.} Same conventions as Table~\ref{tab:main_pool_sd}, with SMS replaced by ARI. The tuned configuration in this table is selected on ARI itself (not on SMS), so each table is self-contained.} \label{tab:main_pool_sd_ari} \\
\toprule
\multicolumn{2}{c}{} & \multicolumn{5}{c}{\textbf{Non-Guided}} & \multicolumn{5}{c}{\textbf{Guided}} \\ \cmidrule(lr){3-7} \cmidrule(lr){8-12}
 \textbf{Algo} & \textbf{Dataset}  & Def. & Worst & Med. & Tuned & $\tilde{t}$ [s]  & Def. & Worst & Med. & Tuned & $\tilde{t}$ [s] \\
\midrule
\endfirsthead
\multicolumn{12}{l}{\emph{(continued from previous page)}} \\
\toprule
\multicolumn{2}{c}{} & \multicolumn{5}{c}{\textbf{Non-Guided}} & \multicolumn{5}{c}{\textbf{Guided}} \\ \cmidrule(lr){3-7} \cmidrule(lr){8-12}
 \textbf{Algo} & \textbf{Dataset}  & Def. & Worst & Med. & Tuned & $\tilde{t}$ [s]  & Def. & Worst & Med. & Tuned & $\tilde{t}$ [s] \\
\midrule
\endhead
\midrule \multicolumn{12}{r}{\emph{(continued on next page)}} \\
\endfoot
\bottomrule
\endlastfoot
\multirow{8}{*}{\textbf{AutoPlait}} & ActRecTut & 0.19 & 0.19 & 0.19 & 0.19 & 47.4 & 0.19 & 0.19 & 0.19 & 0.19 & 47.4 \\*
 & HAS & 0.47 & 0.47 & 0.47 & 0.47 & 10.4 & 0.47 & 0.47 & 0.47 & 0.47 & 10.4 \\*
 & MoCap & 0.80 & 0.80 & 0.80 & 0.80 & 25.0 & 0.80 & 0.80 & 0.80 & 0.80 & 25.0 \\*
 & PUMP & 0.24 & 0.24 & 0.24 & 0.24 & 316 & 0.24 & 0.24 & 0.24 & 0.24 & 316 \\*
 & SKAB & 0.38 & 0.38 & 0.38 & 0.38 & 1.5 & 0.38 & 0.38 & 0.38 & 0.38 & 1.5 \\*
 & TSSB & 0.08 & 0.08 & 0.08 & 0.08 & 0.7 & 0.08 & 0.08 & 0.08 & 0.08 & 0.7 \\*
 & USC-HAD & 0.29 & 0.29 & 0.29 & 0.29 & 155 & 0.29 & 0.29 & 0.29 & 0.29 & 155 \\*
 & UTSA & 0.22 & 0.22 & 0.22 & 0.22 & 2.2 & 0.22 & 0.22 & 0.22 & 0.22 & 2.2 \\
\midrule
\multirow{8}{*}{\textbf{CLaP}} & ActRecTut & 0.21 & 0.35 & 0.42 & 0.43 & 623 & 0.38 & 0.32 & 0.34 & 0.44 & 504 \\*
 & HAS & 0.59 & 0.47 & 0.54 & 0.60 & 15.1 & 0.66 & 0.56 & 0.61 & 0.69 & 13.4 \\*
 & MoCap & 0.59 & 0.56 & 0.67 & 0.67 & 25.5 & 0.62 & 0.57 & 0.65 & 0.71 & 25.9 \\*
 & PUMP & 0.12 & 0.02 & 0.10 & 0.13 & 366 & 0.14 & 0.06 & 0.13 & 0.24 & 550 \\*
 & SKAB & 0.01 & 0.00 & 0.00 & 0.01 & 2.2 & 0.38 & 0.06 & 0.17 & 0.38 & 4.7 \\*
 & TSSB & 0.73 & 0.71 & 0.77 & 0.75 & 9.4 & 0.81 & 0.80 & 0.83 & 0.82 & 5.7 \\*
 & USC-HAD & 0.81 & 0.81 & 0.87 & 0.89 & 500 & 0.71 & 0.71 & 0.75 & 0.76 & 490 \\*
 & UTSA & 0.75 & 0.66 & 0.74 & 0.75 & 27.5 & 0.91 & 0.82 & 0.90 & 0.91 & 18.4 \\
\midrule
\multirow{8}{*}{\textbf{E2USD}} & ActRecTut & 0.62 & 0.58 & 0.67 & 0.76 & 12.5 & 0.68 & 0.68 & 0.72 & 0.80 & 15.2 \\*
 & HAS & 0.40 & 0.34 & 0.35 & 0.40 & 12.7 & 0.57 & 0.54 & 0.56 & 0.58 & 13.0 \\*
 & MoCap & 0.71 & 0.44 & 0.58 & 0.72 & 12.3 & 0.61 & 0.48 & 0.57 & 0.62 & 11.7 \\*
 & PUMP & 0.27 & 0.24 & 0.29 & 0.35 & 17.8 & 0.28 & 0.24 & 0.31 & 0.37 & 12.2 \\*
 & SKAB & 0.23 & 0.17 & 0.21 & 0.27 & 15.1 & 0.28 & 0.25 & 0.31 & 0.37 & 16.0 \\*
 & TSSB & 0.20 & 0.19 & 0.22 & 0.26 & 17.4 & 0.27 & 0.26 & 0.28 & 0.32 & 14.2 \\*
 & USC-HAD & 0.66 & 0.50 & 0.59 & 0.67 & 16.4 & 0.60 & 0.46 & 0.53 & 0.62 & 14.9 \\*
 & UTSA & 0.30 & 0.22 & 0.27 & 0.31 & 17.7 & 0.33 & 0.22 & 0.30 & 0.39 & 12.7 \\
\midrule
\multirow{8}{*}{\textbf{HDP-HSMM}} & ActRecTut & 0.34 & 0.16 & 0.39 & 0.50 & 6.2 & 0.44 & 0.14 & 0.38 & 0.44 & 15.1 \\*
 & HAS & 0.45 & 0.42 & 0.48 & 0.50 & 19.9 & 0.55 & 0.48 & 0.53 & 0.55 & 4.0 \\*
 & MoCap & 0.47 & 0.41 & 0.51 & 0.61 & 6.3 & 0.52 & 0.37 & 0.47 & 0.58 & 10.6 \\*
 & PUMP & 0.11 & 0.08 & 0.11 & 0.11 & 13.2 & 0.11 & 0.10 & 0.11 & 0.11 & 24.8 \\*
 & SKAB & 0.39 & 0.34 & 0.38 & 0.46 & 6.2 & 0.48 & 0.27 & 0.36 & 0.48 & 4.1 \\*
 & TSSB & 0.28 & 0.16 & 0.21 & 0.32 & 16.7 & 0.28 & 0.20 & 0.23 & 0.29 & 4.8 \\*
 & USC-HAD & 0.34 & 0.30 & 0.35 & 0.37 & 23.2 & 0.33 & 0.30 & 0.34 & 0.35 & 9.0 \\*
 & UTSA & 0.40 & 0.38 & 0.43 & 0.51 & 11.4 & 0.28 & 0.17 & 0.25 & 0.38 & 5.5 \\
\midrule
\multirow{8}{*}{\textbf{TICC}} & ActRecTut & - & - & - & - & - & 0.80 & 0.70 & 0.78 & 0.83 & 1.0k \\*
 & HAS & - & - & - & - & - & 0.64 & 0.52 & 0.60 & 0.64 & 38.8 \\*
 & MoCap & - & - & - & - & - & 0.74 & 0.53 & 0.61 & 0.73 & 218 \\*
 & PUMP & - & - & - & - & - & 0.50 & 0.49 & 0.50 & 0.50 & 2.0k \\*
 & SKAB & - & - & - & - & - & 0.52 & 0.49 & 0.56 & 0.57 & 22.1 \\*
 & TSSB & - & - & - & - & - & 0.08 & 0.08 & 0.10 & 0.27 & 22.4 \\*
 & USC-HAD & - & - & - & - & - & 0.58 & 0.45 & 0.50 & 0.61 & 716 \\*
 & UTSA & - & - & - & - & - & 0.14 & 0.12 & 0.14 & 0.25 & 12.4 \\
\midrule
\multirow{8}{*}{\textbf{Time2State}} & ActRecTut & 0.66 & 0.57 & 0.68 & 0.75 & 34.7 & 0.64 & 0.61 & 0.68 & 0.78 & 12.7 \\*
 & HAS & 0.37 & 0.33 & 0.35 & 0.38 & 14.7 & 0.54 & 0.52 & 0.54 & 0.57 & 7.0 \\*
 & MoCap & 0.79 & 0.47 & 0.70 & 0.81 & 31.8 & 0.71 & 0.46 & 0.66 & 0.75 & 9.6 \\*
 & PUMP & 0.31 & 0.27 & 0.34 & 0.41 & 17.6 & 0.32 & 0.27 & 0.36 & 0.45 & 8.6 \\*
 & SKAB & 0.22 & 0.21 & 0.25 & 0.33 & 12.5 & 0.21 & 0.21 & 0.43 & 0.61 & 8.1 \\*
 & TSSB & 0.27 & 0.12 & 0.23 & 0.31 & 14.5 & 0.33 & 0.20 & 0.30 & 0.43 & 10.2 \\*
 & USC-HAD & 0.70 & 0.57 & 0.65 & 0.71 & 17.4 & 0.66 & 0.54 & 0.63 & 0.68 & 16.2 \\*
 & UTSA & 0.30 & 0.18 & 0.25 & 0.33 & 13.2 & 0.41 & 0.20 & 0.33 & 0.42 & 26.6 \\
\end{longtable}
\endgroup

\begingroup
\setlength{\tabcolsep}{3pt}\renewcommand{\arraystretch}{0.95}\small
\begin{longtable}{l l c c c c c c c c c c}
\caption{\textbf{Main-pool SD algorithms: per-(algorithm, dataset) default / worst-case / median / tuned WARI, in both Non-Guided (NG) and Guided (G) modes.} Same conventions as Table~\ref{tab:main_pool_sd}, with SMS replaced by WARI. The tuned configuration in this table is selected on WARI itself (not on SMS), so each table is self-contained.} \label{tab:main_pool_sd_wari} \\
\toprule
\multicolumn{2}{c}{} & \multicolumn{5}{c}{\textbf{Non-Guided}} & \multicolumn{5}{c}{\textbf{Guided}} \\ \cmidrule(lr){3-7} \cmidrule(lr){8-12}
 \textbf{Algo} & \textbf{Dataset}  & Def. & Worst & Med. & Tuned & $\tilde{t}$ [s]  & Def. & Worst & Med. & Tuned & $\tilde{t}$ [s] \\
\midrule
\endfirsthead
\multicolumn{12}{l}{\emph{(continued from previous page)}} \\
\toprule
\multicolumn{2}{c}{} & \multicolumn{5}{c}{\textbf{Non-Guided}} & \multicolumn{5}{c}{\textbf{Guided}} \\ \cmidrule(lr){3-7} \cmidrule(lr){8-12}
 \textbf{Algo} & \textbf{Dataset}  & Def. & Worst & Med. & Tuned & $\tilde{t}$ [s]  & Def. & Worst & Med. & Tuned & $\tilde{t}$ [s] \\
\midrule
\endhead
\midrule \multicolumn{12}{r}{\emph{(continued on next page)}} \\
\endfoot
\bottomrule
\endlastfoot
\multirow{8}{*}{\textbf{AutoPlait}} & ActRecTut & 0.26 & 0.26 & 0.26 & 0.26 & 47.4 & 0.26 & 0.26 & 0.26 & 0.26 & 47.4 \\*
 & HAS & 0.51 & 0.51 & 0.51 & 0.51 & 10.4 & 0.51 & 0.51 & 0.51 & 0.51 & 10.4 \\*
 & MoCap & 0.89 & 0.89 & 0.89 & 0.89 & 25.0 & 0.89 & 0.89 & 0.89 & 0.89 & 25.0 \\*
 & PUMP & 0.36 & 0.36 & 0.36 & 0.36 & 316 & 0.36 & 0.36 & 0.36 & 0.36 & 316 \\*
 & SKAB & 0.57 & 0.57 & 0.57 & 0.57 & 1.5 & 0.57 & 0.57 & 0.57 & 0.57 & 1.5 \\*
 & TSSB & 0.08 & 0.08 & 0.08 & 0.08 & 0.7 & 0.08 & 0.08 & 0.08 & 0.08 & 0.7 \\*
 & USC-HAD & 0.41 & 0.41 & 0.41 & 0.41 & 155 & 0.41 & 0.41 & 0.41 & 0.41 & 155 \\*
 & UTSA & 0.21 & 0.21 & 0.21 & 0.21 & 2.2 & 0.21 & 0.21 & 0.21 & 0.21 & 2.2 \\
\midrule
\multirow{8}{*}{\textbf{CLaP}} & ActRecTut & 0.25 & 0.45 & 0.51 & 0.54 & 623 & 0.46 & 0.35 & 0.42 & 0.60 & 647 \\*
 & HAS & 0.64 & 0.52 & 0.60 & 0.64 & 15.1 & 0.74 & 0.63 & 0.70 & 0.76 & 13.4 \\*
 & MoCap & 0.70 & 0.66 & 0.79 & 0.79 & 25.5 & 0.74 & 0.68 & 0.79 & 0.82 & 25.9 \\*
 & PUMP & 0.18 & 0.03 & 0.14 & 0.20 & 385 & 0.25 & 0.08 & 0.20 & 0.40 & 550 \\*
 & SKAB & 0.02 & 0.00 & 0.00 & 0.02 & 2.2 & 0.54 & 0.10 & 0.28 & 0.54 & 4.7 \\*
 & TSSB & 0.77 & 0.74 & 0.81 & 0.79 & 8.3 & 0.86 & 0.85 & 0.87 & 0.86 & 5.7 \\*
 & USC-HAD & 0.89 & 0.89 & 0.94 & 0.95 & 500 & 0.82 & 0.82 & 0.85 & 0.85 & 490 \\*
 & UTSA & 0.79 & 0.68 & 0.80 & 0.79 & 27.4 & 0.95 & 0.85 & 0.94 & 0.95 & 18.4 \\
\midrule
\multirow{8}{*}{\textbf{E2USD}} & ActRecTut & 0.71 & 0.68 & 0.73 & 0.76 & 12.5 & 0.85 & 0.78 & 0.85 & 0.87 & 15.2 \\*
 & HAS & 0.44 & 0.35 & 0.38 & 0.44 & 12.7 & 0.64 & 0.57 & 0.61 & 0.64 & 13.3 \\*
 & MoCap & 0.84 & 0.47 & 0.66 & 0.85 & 12.3 & 0.70 & 0.51 & 0.64 & 0.73 & 10.2 \\*
 & PUMP & 0.32 & 0.28 & 0.33 & 0.37 & 18.1 & 0.33 & 0.28 & 0.34 & 0.40 & 12.2 \\*
 & SKAB & 0.38 & 0.24 & 0.32 & 0.38 & 15.6 & 0.45 & 0.44 & 0.49 & 0.53 & 16.0 \\*
 & TSSB & 0.25 & 0.22 & 0.27 & 0.32 & 17.4 & 0.31 & 0.30 & 0.33 & 0.38 & 15.3 \\*
 & USC-HAD & 0.76 & 0.60 & 0.69 & 0.76 & 14.8 & 0.71 & 0.55 & 0.63 & 0.72 & 14.9 \\*
 & UTSA & 0.32 & 0.23 & 0.29 & 0.33 & 17.7 & 0.39 & 0.25 & 0.36 & 0.45 & 12.7 \\
\midrule
\multirow{8}{*}{\textbf{HDP-HSMM}} & ActRecTut & 0.47 & 0.16 & 0.48 & 0.79 & 6.2 & 0.62 & 0.10 & 0.48 & 0.62 & 15.1 \\*
 & HAS & 0.50 & 0.47 & 0.53 & 0.55 & 19.9 & 0.62 & 0.53 & 0.59 & 0.62 & 4.0 \\*
 & MoCap & 0.55 & 0.48 & 0.60 & 0.68 & 6.3 & 0.63 & 0.44 & 0.55 & 0.67 & 10.6 \\*
 & PUMP & 0.15 & 0.12 & 0.15 & 0.16 & 13.2 & 0.15 & 0.14 & 0.15 & 0.16 & 24.8 \\*
 & SKAB & 0.57 & 0.48 & 0.59 & 0.65 & 6.2 & 0.62 & 0.33 & 0.51 & 0.62 & 4.1 \\*
 & TSSB & 0.36 & 0.20 & 0.27 & 0.40 & 16.7 & 0.33 & 0.22 & 0.27 & 0.35 & 4.8 \\*
 & USC-HAD & 0.47 & 0.40 & 0.48 & 0.50 & 23.2 & 0.45 & 0.40 & 0.46 & 0.47 & 9.2 \\*
 & UTSA & 0.45 & 0.43 & 0.47 & 0.57 & 11.4 & 0.31 & 0.15 & 0.27 & 0.42 & 5.5 \\
\midrule
\multirow{8}{*}{\textbf{TICC}} & ActRecTut & - & - & - & - & - & 0.82 & 0.66 & 0.79 & 0.93 & 1.0k \\*
 & HAS & - & - & - & - & - & 0.68 & 0.53 & 0.63 & 0.69 & 38.8 \\*
 & MoCap & - & - & - & - & - & 0.83 & 0.59 & 0.71 & 0.84 & 218 \\*
 & PUMP & - & - & - & - & - & 0.53 & 0.48 & 0.49 & 0.50 & 2.0k \\*
 & SKAB & - & - & - & - & - & 0.70 & 0.68 & 0.74 & 0.74 & 22.1 \\*
 & TSSB & - & - & - & - & - & 0.08 & 0.08 & 0.10 & 0.29 & 22.4 \\*
 & USC-HAD & - & - & - & - & - & 0.68 & 0.52 & 0.58 & 0.72 & 716 \\*
 & UTSA & - & - & - & - & - & 0.16 & 0.13 & 0.16 & 0.27 & 12.4 \\
\midrule
\multirow{8}{*}{\textbf{Time2State}} & ActRecTut & 0.73 & 0.64 & 0.72 & 0.78 & 34.7 & 0.74 & 0.73 & 0.77 & 0.89 & 19.0 \\*
 & HAS & 0.41 & 0.35 & 0.37 & 0.43 & 14.7 & 0.61 & 0.58 & 0.60 & 0.64 & 6.8 \\*
 & MoCap & 0.91 & 0.50 & 0.79 & 0.92 & 31.8 & 0.82 & 0.48 & 0.75 & 0.87 & 10.4 \\*
 & PUMP & 0.39 & 0.34 & 0.42 & 0.48 & 17.6 & 0.41 & 0.35 & 0.44 & 0.51 & 9.1 \\*
 & SKAB & 0.34 & 0.29 & 0.35 & 0.43 & 12.5 & 0.34 & 0.34 & 0.66 & 0.77 & 8.1 \\*
 & TSSB & 0.32 & 0.14 & 0.26 & 0.38 & 14.5 & 0.40 & 0.22 & 0.35 & 0.51 & 9.4 \\*
 & USC-HAD & 0.81 & 0.65 & 0.75 & 0.81 & 28.1 & 0.79 & 0.64 & 0.74 & 0.79 & 16.2 \\*
 & UTSA & 0.34 & 0.20 & 0.28 & 0.36 & 13.2 & 0.49 & 0.24 & 0.40 & 0.50 & 26.6 \\
\end{longtable}
\endgroup

\paragraph{Extended-scope pool.}
Additional metrics for the extended-scope algorithms are reported in the same spirit as Tables~\ref{tab:excluded_algos_cpd}-\ref{tab:excluded_algos_sd}: only for the (algorithm, dataset, mode) cells they actually completed, and with the tuned configuration defined as the one maximising the table's own metric.

\begin{table}[H]
    \centering
    \caption{\textbf{Per-dataset coverage and score (Gaussian $F_1$) of the five extended-scope CPD algorithms.} Tuned configurations are selected on Gaussian $F_1$.}
    \label{tab:excluded_algos_cpd_gf1}
    \resizebox{\textwidth}{!}{\begin{tabular}{l l c c c c c c c c c c c c c}
\toprule
 &  &  & \multicolumn{3}{c}{\textbf{Default, NG}} & \multicolumn{3}{c}{\textbf{Default, G}} & \multicolumn{3}{c}{\textbf{Tuned, NG}} & \multicolumn{3}{c}{\textbf{Tuned, G}} \\
\cmidrule(lr){4-6} \cmidrule(lr){7-9} \cmidrule(lr){10-12} \cmidrule(lr){13-15}
\textbf{Algorithm} & \textbf{Dataset} & \textbf{Series} & Compl. & gF1 & $\tilde{t}$ [s] & Compl. & gF1 & $\tilde{t}$ [s] & Compl. & gF1 & $\tilde{t}$ [s] & Compl. & gF1 & $\tilde{t}$ [s] \\
\midrule
\multirow{8}{*}{tGLAD} & actrectut & 2 & 2/2 & 0.05 & 36706 & 2/2 & 0.05 & 33986 & 2/2 & 0.36 & 9885 & 2/2 & 0.39 & 8928 \\
 & has & 250 & 206/250 & 0.11 & 288 & 171/250 & 0.10 & 343 & 6/250 & 0.17 & 35.2 & 171/250 & 0.14 & 323 \\
 & mocap & 9 & 9/9 & 0.05 & 7304 & 9/9 & 0.05 & 5256 & 9/9 & 0.18 & 1752 & 9/9 & 0.19 & 1735 \\
 & pump & 120 & 57/120 & 0.03 & 1367 & 55/120 & 0.02 & 1575 & 10/120 & 0.49 & 7741 & 12/120 & 0.50 & 6669 \\
 & skab & 34 & 34/34 & 0.06 & 1326 & 34/34 & 0.06 & 1301 & 34/34 & 0.09 & 309 & 34/34 & 0.11 & 317 \\
 & tssb & 75 & 75/75 & 0.06 & 191 & 75/75 & 0.06 & 240 & 75/75 & 0.08 & 147 & 75/75 & 0.08 & 19.8 \\
 & usc-had & 70 & 44/70 & 0.01 & 1796 & 42/70 & 0.01 & 1891 & 10/70 & 0.24 & 8153 & 7/70 & 0.25 & 12580 \\
 & utsa & 32 & 32/32 & 0.02 & 585 & 32/32 & 0.02 & 548 & 32/32 & 0.03 & 522 & 32/32 & 0.02 & 71.0 \\
\midrule
\multirow{8}{*}{TS-CP$^2$} & actrectut & 2 & 2/2 & 0.07 & 10924 & 2/2 & 0.20 & 10931 & - & - & - & - & - & - \\
 & has & 250 & 32/250 & 0.02 & 1727 & 33/250 & 0.16 & 1630 & - & - & - & - & - & - \\
 & mocap & 9 & 9/9 & 0.04 & 3151 & 9/9 & 0.11 & 2516 & - & - & - & - & - & - \\
 & pump & 120 & 10/120 & 0.14 & 8133 & 10/120 & 0.17 & 8144 & - & - & - & - & - & - \\
 & skab & 34 & 34/34 & 0.02 & 237 & 34/34 & 0.04 & 245 & - & - & - & - & - & - \\
 & tssb & 75 & 50/75 & 0.11 & 1222 & 48/75 & 0.23 & 1166 & - & - & - & - & - & - \\
 & usc-had & 70 & 6/70 & 0.04 & 14199 & 6/70 & 0.15 & 13447 & - & - & - & - & - & - \\
 & utsa & 32 & 20/32 & 0.01 & 3728 & 20/32 & 0.15 & 3636 & - & - & - & - & - & - \\
\midrule
\multirow{8}{*}{EAgglo} & actrectut & 2 & 0/2 & - & - & 0/2 & - & - & - & - & - & - & - & - \\
 & has & 250 & 8/250 & 0.26 & 33.8 & 8/250 & 0.26 & 33.4 & - & - & - & - & - & - \\
 & mocap & 9 & 9/9 & 0.23 & 108 & 9/9 & 0.23 & 105 & - & - & - & - & - & - \\
 & pump & 120 & 9/120 & 0.32 & 995 & 9/120 & 0.32 & 1036 & - & - & - & - & - & - \\
 & skab & 34 & 34/34 & 0.00 & 6.9 & 34/34 & 0.00 & 3.7 & - & - & - & - & - & - \\
 & tssb & 75 & 75/75 & 0.01 & 22.3 & 75/75 & 0.01 & 26.7 & - & - & - & - & - & - \\
 & usc-had & 70 & 0/70 & - & - & 0/70 & - & - & - & - & - & - & - & - \\
 & utsa & 32 & 28/32 & 0.01 & 320 & 28/32 & 0.01 & 278 & - & - & - & - & - & - \\
\midrule
\multirow{8}{*}{BOCD} & actrectut & 2 & 0/2 & - & - & 0/2 & - & - & - & - & - & - & - & - \\
 & has & 250 & 8/250 & 0.13 & 718 & 8/250 & 0.13 & 738 & - & - & - & - & - & - \\
 & mocap & 9 & 9/9 & 0.00 & 3213 & 9/9 & 0.00 & 3310 & - & - & - & - & - & - \\
 & pump & 120 & 1/120 & 0.00 & 56912 & 1/120 & 0.00 & 65789 & - & - & - & - & - & - \\
 & skab & 34 & 34/34 & 0.14 & 63.5 & 34/34 & 0.13 & 67.0 & - & - & - & - & - & - \\
 & tssb & 75 & 28/75 & 0.07 & 391 & 28/75 & 0.07 & 397 & - & - & - & - & - & - \\
 & usc-had & 70 & 0/70 & - & - & 0/70 & - & - & - & - & - & - & - & - \\
 & utsa & 32 & 5/32 & 0.00 & 1443 & 5/32 & 0.00 & 1494 & - & - & - & - & - & - \\
\midrule
\multirow{6}{*}{Hidalgo} & actrectut & 2 & - & - & - & 2/2 & 0.00 & 10846 & - & - & - & 2/2 & 0.22 & 16506 \\
 & has & 250 & - & - & - & 8/250 & 0.13 & 757 & - & - & - & 8/250 & 0.13 & 776 \\
 & mocap & 9 & - & - & - & 9/9 & 0.02 & 1855 & - & - & - & 9/9 & 0.03 & 2746 \\
 & pump & 120 & - & - & - & 2/120 & 0.02 & 41896 & - & - & - & 2/120 & 0.06 & 33118 \\
 & skab & 34 & - & - & - & 34/34 & 0.01 & 151 & - & - & - & 34/34 & 0.02 & 134 \\
 & usc-had & 70 & - & - & - & 3/70 & 0.00 & 24346 & - & - & - & 4/70 & 0.01 & 18350 \\
\bottomrule
\end{tabular}
}
\end{table}

\begin{table}[H]
    \centering
    \caption{\textbf{Per-dataset coverage and score (ARI) of the SD-side extended-scope algorithm (Hidalgo).} Tuned configurations are selected on ARI.}
    \label{tab:excluded_algos_sd_ari}
    \resizebox{\textwidth}{!}{\begin{tabular}{l l c c c c c c c c c c c c c}
\toprule
 &  &  & \multicolumn{3}{c}{\textbf{Default, NG}} & \multicolumn{3}{c}{\textbf{Default, G}} & \multicolumn{3}{c}{\textbf{Tuned, NG}} & \multicolumn{3}{c}{\textbf{Tuned, G}} \\
\cmidrule(lr){4-6} \cmidrule(lr){7-9} \cmidrule(lr){10-12} \cmidrule(lr){13-15}
\textbf{Algorithm} & \textbf{Dataset} & \textbf{Series} & Compl. & ARI & $\tilde{t}$ [s] & Compl. & ARI & $\tilde{t}$ [s] & Compl. & ARI & $\tilde{t}$ [s] & Compl. & ARI & $\tilde{t}$ [s] \\
\midrule
\multirow{6}{*}{Hidalgo} & actrectut & 2 & - & - & - & 2/2 & 0.07 & 10846 & - & - & - & 2/2 & 0.07 & 10487 \\
 & has & 250 & - & - & - & 8/250 & 0.15 & 757 & - & - & - & 8/250 & 0.15 & 731 \\
 & mocap & 9 & - & - & - & 9/9 & 0.05 & 1855 & - & - & - & 9/9 & 0.05 & 1716 \\
 & pump & 120 & - & - & - & 2/120 & 0.06 & 41896 & - & - & - & 1/120 & 0.06 & 72877 \\
 & skab & 34 & - & - & - & 34/34 & 0.13 & 151 & - & - & - & 34/34 & 0.14 & 221 \\
 & usc-had & 70 & - & - & - & 3/70 & 0.07 & 24346 & - & - & - & 3/70 & 0.07 & 23105 \\
\bottomrule
\end{tabular}
}
\end{table}

\begin{table}[H]
    \centering
    \caption{\textbf{Per-dataset coverage and score (WARI) of the SD-side extended-scope algorithm (Hidalgo).} Tuned configurations are selected on WARI.}
    \label{tab:excluded_algos_sd_wari}
    \resizebox{\textwidth}{!}{\begin{tabular}{l l c c c c c c c c c c c c c}
\toprule
 &  &  & \multicolumn{3}{c}{\textbf{Default, NG}} & \multicolumn{3}{c}{\textbf{Default, G}} & \multicolumn{3}{c}{\textbf{Tuned, NG}} & \multicolumn{3}{c}{\textbf{Tuned, G}} \\
\cmidrule(lr){4-6} \cmidrule(lr){7-9} \cmidrule(lr){10-12} \cmidrule(lr){13-15}
\textbf{Algorithm} & \textbf{Dataset} & \textbf{Series} & Compl. & WARI & $\tilde{t}$ [s] & Compl. & WARI & $\tilde{t}$ [s] & Compl. & WARI & $\tilde{t}$ [s] & Compl. & WARI & $\tilde{t}$ [s] \\
\midrule
\multirow{6}{*}{Hidalgo} & actrectut & 2 & - & - & - & 2/2 & 0.06 & 10846 & - & - & - & 2/2 & 0.06 & 10487 \\
 & has & 250 & - & - & - & 8/250 & 0.15 & 757 & - & - & - & 8/250 & 0.15 & 731 \\
 & mocap & 9 & - & - & - & 9/9 & 0.06 & 1855 & - & - & - & 9/9 & 0.06 & 1716 \\
 & pump & 120 & - & - & - & 2/120 & 0.05 & 41896 & - & - & - & 2/120 & 0.05 & 42438 \\
 & skab & 34 & - & - & - & 34/34 & 0.17 & 151 & - & - & - & 34/34 & 0.21 & 221 \\
 & usc-had & 70 & - & - & - & 3/70 & 0.09 & 24346 & - & - & - & 3/70 & 0.09 & 23105 \\
\bottomrule
\end{tabular}
}
\end{table}

\end{document}